\documentclass{beamer}%
\usepackage{amsmath}
\usepackage{amsfonts}
\usepackage{amssymb}
\usepackage{mathpazo}
\usepackage{hyperref}
\usepackage{multimedia}
\usepackage{graphicx}
\providecommand{\U}[1]{\protect\rule{.1in}{.1in}}
\newtheorem{remark}[theorem]{Remark}
\begin{document}

\title{Flooding edge or node weighted graphs}
\author{Fernand Meyer}
\institute{Centre de Morphologie Math\'{e}matique}
\date{22\ March 2013}
\maketitle

\section{Flooding node and edge weighted graphs}%

%TCIMACRO{\TeXButton{BeginFrame}{\begin{frame}}}%
%BeginExpansion
\begin{frame}%
%EndExpansion
%

%TCIMACRO{\QTR{frametitle}{Introduction}}%
%BeginExpansion
\frametitle{Introduction}%
%EndExpansion

Floodings are useful for:

\begin{itemize}
\item filtering images

\item suppressing regional minima and filling them by lakes -%
%TCIMACRO{\TEXTsymbol{>} }%
%BeginExpansion
$>$
%EndExpansion
regularization and control of the watershed segmentation
\end{itemize}

Combined with the dual operator, the razings, they permit to construct
powerful autodual filters (alternate sequential, flattenings and levelings)%

%TCIMACRO{\TeXButton{EndFrame}{\end{frame}}}%
%BeginExpansion
\end{frame}%
%EndExpansion%
%TCIMACRO{\TeXButton{BeginFrame}{\begin{frame}}}%
%BeginExpansion
\begin{frame}%
%EndExpansion
%

%TCIMACRO{\QTR{frametitle}{Outline of the talk}}%
%BeginExpansion
\frametitle{Outline of the talk}%
%EndExpansion

We flood two types of graphs:

- node weighted graphs with a ground level on the nodes

- edge weighted graph without ground level on the nodes

The flooding assigns a flooding to each node.

We characterize valid floodings on both types of graphs: lakes, regional
minima lakes.\ 

We show that flooding a node weighted graph is equivalent with flooding an
edge weighted graph with appropriate edge weights%

%TCIMACRO{\TeXButton{EndFrame}{\end{frame}}}%
%BeginExpansion
\end{frame}%
%EndExpansion
{}%

%TCIMACRO{\TeXButton{BeginFrame}{\begin{frame}}}%
%BeginExpansion
\begin{frame}%
%EndExpansion
%

%TCIMACRO{\QTR{frametitle}{Outline of the talk}}%
%BeginExpansion
\frametitle{Outline of the talk}%
%EndExpansion

We then introduce dominated floodings under a ceiling function.

We present two classes of algorithms :

\begin{itemize}
\item shortest path algorithms for the ultrametric flooding distance

\item direct construction of the flooding on the dendrogram of the closed
balls of the flooding distance
\end{itemize}

%

%TCIMACRO{\TeXButton{EndFrame}{\end{frame}}}%
%BeginExpansion
\end{frame}%
%EndExpansion
{}%

%TCIMACRO{\TeXButton{BeginFrame}{\begin{frame}}}%
%BeginExpansion
\begin{frame}%
%EndExpansion

\begin{center}
{\Large \alert{Reminders on graphs}}
\end{center}

%

%TCIMACRO{\TeXButton{EndFrame}{\end{frame}}}%
%BeginExpansion
\end{frame}%
%EndExpansion
%

%TCIMACRO{\TeXButton{BeginFrame}{\begin{frame}}}%
%BeginExpansion
\begin{frame}%
%EndExpansion
%

%TCIMACRO{\QTR{frametitle}{Graphs}}%
%BeginExpansion
\frametitle{Graphs}%
%EndExpansion

A \textit{non oriented graph} $G=\left[  N,E\right]  :$ $N$ = nodes ;
$E=,$edges ; an edge $u\in E=$ a pair of vertices

\textit{A chain} of length $n$ is a sequence of $n$ edges $L=\left\{
e_{1},e_{2,}\ldots,e_{n}\right\}  $, with successive edges having a common node.

A \textit{path }between two nodes $x$ and $y$ is a sequence of nodes
$(n_{1}=x,n_{2},...,n_{k}=y)$ with successive nodes linked by an edge.\ 

\textit{A cocycle} is the set of all edges with one extremity in a subset $Y$
and the other in the complementary set $\overline{Y}.$

The subgraph spanning a set $A\subset N$ is the graph $G_{A}=[A,E_{A}]$, where
$E_{A}$ are the edges linking two nodes of $A.$

The partial graph associated to the edges $E^{\prime}\subset E$ is $G^{\prime
}=[N,E^{\prime}].$%

%TCIMACRO{\TeXButton{EndFrame}{\end{frame}}}%
%BeginExpansion
\end{frame}%
%EndExpansion
{}%

%TCIMACRO{\TeXButton{BeginFrame}{\begin{frame}}}%
%BeginExpansion
\begin{frame}%
%EndExpansion
%

%TCIMACRO{\QTR{frametitle}{Weighted graphs}}%
%BeginExpansion
\frametitle{Weighted graphs}%
%EndExpansion

In a graph $G=\left[  N,E\right]  ,$ edges and nodes may be weighted :

\begin{itemize}
\item $e_{ij}$ is the weight of the edge $(i,j)$

\item $n_{i}$ the weight of the node $i.$ The weights take their value in a
completely ordered lattice $\mathcal{T}$.
\end{itemize}

%

%TCIMACRO{\TeXButton{EndFrame}{\end{frame}}}%
%BeginExpansion
\end{frame}%
%EndExpansion
{}%

%TCIMACRO{\TeXButton{BeginFrame}{\begin{frame}}}%
%BeginExpansion
\begin{frame}%
%EndExpansion
%

%TCIMACRO{\QTR{frametitle}{Flat zones and regional minima on node weighted
%graphs}}%
%BeginExpansion
\frametitle{Flat zones and regional minima on node weighted graphs}%
%EndExpansion

A subgraph $G^{\prime}$ of a node weighted graph $G$ is a flat zone, if any
two nodes of $G^{\prime}$ are connected by a path where all nodes have the
same altitude.

A subgraph $G^{\prime}$ of a graph $G$ is a regional minimum if $G^{\prime} $
is a flat zone and all neighboring nodes have a higher altitude%

%TCIMACRO{\TeXButton{EndFrame}{\end{frame}}}%
%BeginExpansion
\end{frame}%
%EndExpansion
{}%

%TCIMACRO{\TeXButton{BeginFrame}{\begin{frame}}}%
%BeginExpansion
\begin{frame}%
%EndExpansion

\begin{center}
{\Large \alert{Distances on a graph}}

\bigskip

\textbf{Case of edge weighed graphs}
\end{center}

%

%TCIMACRO{\TeXButton{EndFrame}{\end{frame}}}%
%BeginExpansion
\end{frame}%
%EndExpansion
%

%TCIMACRO{\TeXButton{BeginFrame}{\begin{frame}}}%
%BeginExpansion
\begin{frame}%
%EndExpansion
%

%TCIMACRO{\QTR{frametitle}{Constructing distances on an edge weighted graph.}}%
%BeginExpansion
\frametitle{Constructing distances on an edge weighted graph.}%
%EndExpansion

Distances on an edge weighted graph have chains as support :

1) Definition of the weight of a chain, as a measure derived from the edge
weights of the chain elements (example : sum, maximum, etc.)

2) Comparison of two chains by their weight.\ The chain with the smallest
weight is called the shortest.\ 

The distance $d(x,y)$ between two nodes $x$ and $y$ of a graph is $\infty$ if
there is no chain linking these two nodes and equal to the weight of the
shortest chain if such a chain exists.

Given three nodes $(x,y,z)$ the concatenation of the shortest chain $\pi_{xy}
$ between $x$ and $y$ and the shortest chain $\pi_{yz}$ between $y$ and $z$ is
a chain $\pi_{xz}$ between $x$ and $z$, whose weight is smaller or equal to
the weight of the shortest chain between $x$ and $z.$ To each distance
corresponds a particular triangular inequality : $d(x,z)\leq weight( $
$\pi_{xy}\rhd$ $\pi_{yz})$ where $\pi_{xy}\rhd$ $\pi_{yz}$ represents the
concatenation of both chains.%

%TCIMACRO{\TeXButton{EndFrame}{\end{frame}}}%
%BeginExpansion
\end{frame}%
%EndExpansion
%

%TCIMACRO{\TeXButton{BeginFrame}{\begin{frame}}}%
%BeginExpansion
\begin{frame}%
%EndExpansion
%

%TCIMACRO{\QTR{frametitle}{Distance on an edge weighted graph based a the
%length of the shortest chain}}%
%BeginExpansion
\frametitle{Distance on an edge weighted graph based a the length of the
shortest chain}%
%EndExpansion

\textbf{Length of a chain:} The length of a chain between two nodes $x$ and
$y$ is defined as the sum of the weights of its edges.

\textbf{Distance:} The distance $d(x,y)$ between two nodes $x$ and $y$ is the
minimal length of all chains between $x$ and $y$. If there is no chain between
them, the distance is equal to $\infty$.

\textbf{Triangular inequality :} For $(x,y,z):d(x,z)\leq d(x,y)+d(y,z)$%

%TCIMACRO{\TeXButton{EndFrame}{\end{frame}}}%
%BeginExpansion
\end{frame}%
%EndExpansion
%

%TCIMACRO{\TeXButton{BeginFrame}{\begin{frame}}}%
%BeginExpansion
\begin{frame}%
%EndExpansion
%

%TCIMACRO{\QTR{frametitle}{Distance on a graph based on the maximal edge weight
%along the chain}}%
%BeginExpansion
\frametitle{Distance on a graph based on the maximal edge weight along the
chain}%
%EndExpansion

The weights are assigned to the edges, and represent their altitudes.

\textbf{Altitude of a chain: }The altitude of a chain is equal to the highest
weight of the edges along the chain.

\textbf{Flooding distance between two nodes: }The flooding distance
$\operatorname*{fldist}(x,y)$ between nodes $x$ and $y$ is equal to the
minimal altitude of all chains between $x$ and $y$. During a flooding process,
in which a source is placed at location $x,$ the flood would proceed along
this chain of minimal highest altitude to reach the pixel $y$. If there is no
chain between them, the level distance is equal to $\infty$.

\textbf{Triangular inequality :} For $(x,y,z):d(x,z)\leq d(x,y)\vee d(y,z)$ :
ultrametric inequality%

%TCIMACRO{\TeXButton{EndFrame}{\end{frame}}}%
%BeginExpansion
\end{frame}%
%EndExpansion
%

%TCIMACRO{\TeXButton{BeginFrame}{\begin{frame}}}%
%BeginExpansion
\begin{frame}%
%EndExpansion
%

%TCIMACRO{\QTR{frametitle}{The flooding distance is an ultrametric distance}}%
%BeginExpansion
\frametitle{The flooding distance is an ultrametric distance}%
%EndExpansion

An ultrametric distance verifies

* reflexivity : $d(x,x)=0$

* symmetry: $d(x,y)=d(y,x)$

* ultrametric inequality: for all $x,y,z:d(x,y)\leq max\{d(x,z),d(z,y)$\} :
the lowest lake containing both $x$ and $y$ is lower or equal than the lowest
lake containing $x,$ $y$ and $z.$%

%TCIMACRO{\TeXButton{EndFrame}{\end{frame}}}%
%BeginExpansion
\end{frame}%
%EndExpansion
%

%TCIMACRO{\TeXButton{BeginFrame}{\begin{frame}}}%
%BeginExpansion
\begin{frame}%
%EndExpansion
%

%TCIMACRO{\QTR{frametitle}{Distances on a graph : sum and maximum of the edge
%weights}}%
%BeginExpansion
\frametitle{Distances on a graph : sum and maximum of the edge weights}%
%EndExpansion%
%TCIMACRO{\FRAME{ftbpF}{1.4885in}{1.0491in}{0pt}{}{}{wfal62-eps-converted-to.pdf}%
%width 1.4885in;  height 1.0491in;  depth 0pt;  original-width 2.3967in;
%original-height 1.6764in;  cropleft "0";  croptop "1";  cropright "1";
%cropbottom "0";  filename 'wfal62-eps-converted-to.pdf';file-properties "XNPEU";}}}%
%BeginExpansion
\begin{figure}
[ptb]
\begin{center}
\includegraphics[
height=1.0491in,
width=1.4885in
]%
{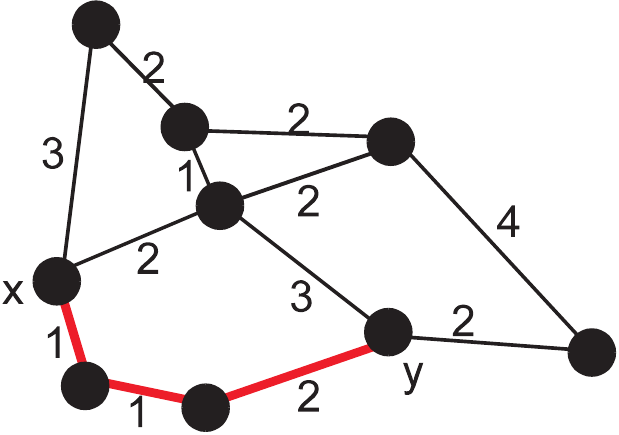}%
\end{center}
\end{figure}
%EndExpansion

The shortest chain (sum of weights of the edges) between $x$ and $y$ is a red
line and has a length of $4$.

The lowest chain (maximal weight of the edges) between $x$ and $y$ is a red
line and a maximal weight of $2$. A flooding between $x$ and $y$ would follow
this chain.%

%TCIMACRO{\TeXButton{EndFrame}{\end{frame}}}%
%BeginExpansion
\end{frame}%
%EndExpansion
%

%TCIMACRO{\TeXButton{BeginFrame}{\begin{frame}}}%
%BeginExpansion
\begin{frame}%
%EndExpansion

\begin{center}
{\Large \alert{Flooding a topographic surface or flooding a graph}}
\end{center}

%

%TCIMACRO{\TeXButton{EndFrame}{\end{frame}}}%
%BeginExpansion
\end{frame}%
%EndExpansion
%

%TCIMACRO{\TeXButton{BeginFrame}{\begin{frame}}}%
%BeginExpansion
\begin{frame}%
%EndExpansion
%

%TCIMACRO{\QTR{frametitle}{The region adjacency graph }}%
%BeginExpansion
\frametitle{The region adjacency graph }%
%EndExpansion

We will work with "neighborhood graphs" where the nodes are the catchment
basins and the edges connect neighboring bassins.\ The edges are weighted by a
dissimilarity measure between adjacent catchment basins; the simplest being
the altitude of the pass-point between two basins.%
%TCIMACRO{\FRAME{ftbpFU}{1.5118in}{1.8041in}{0pt}{\Qcb{The region adjacency
%graph of a topographic surface}}{\Qlb{neigh0}}{neighg0-eps-converted-to.pdf}%
%width 1.5118in;  height 1.8041in;  depth 0pt;  original-width 4.0083in;
%original-height 4.8003in;  cropleft "0";  croptop "1";  cropright "1";
%cropbottom "0";  filename 'neighg0-eps-converted-to.pdf';file-properties "XNPEU";}}}%
%BeginExpansion
\begin{figure}
[ptb]
\begin{center}
\includegraphics[
height=1.8041in,
width=1.5118in
]%
{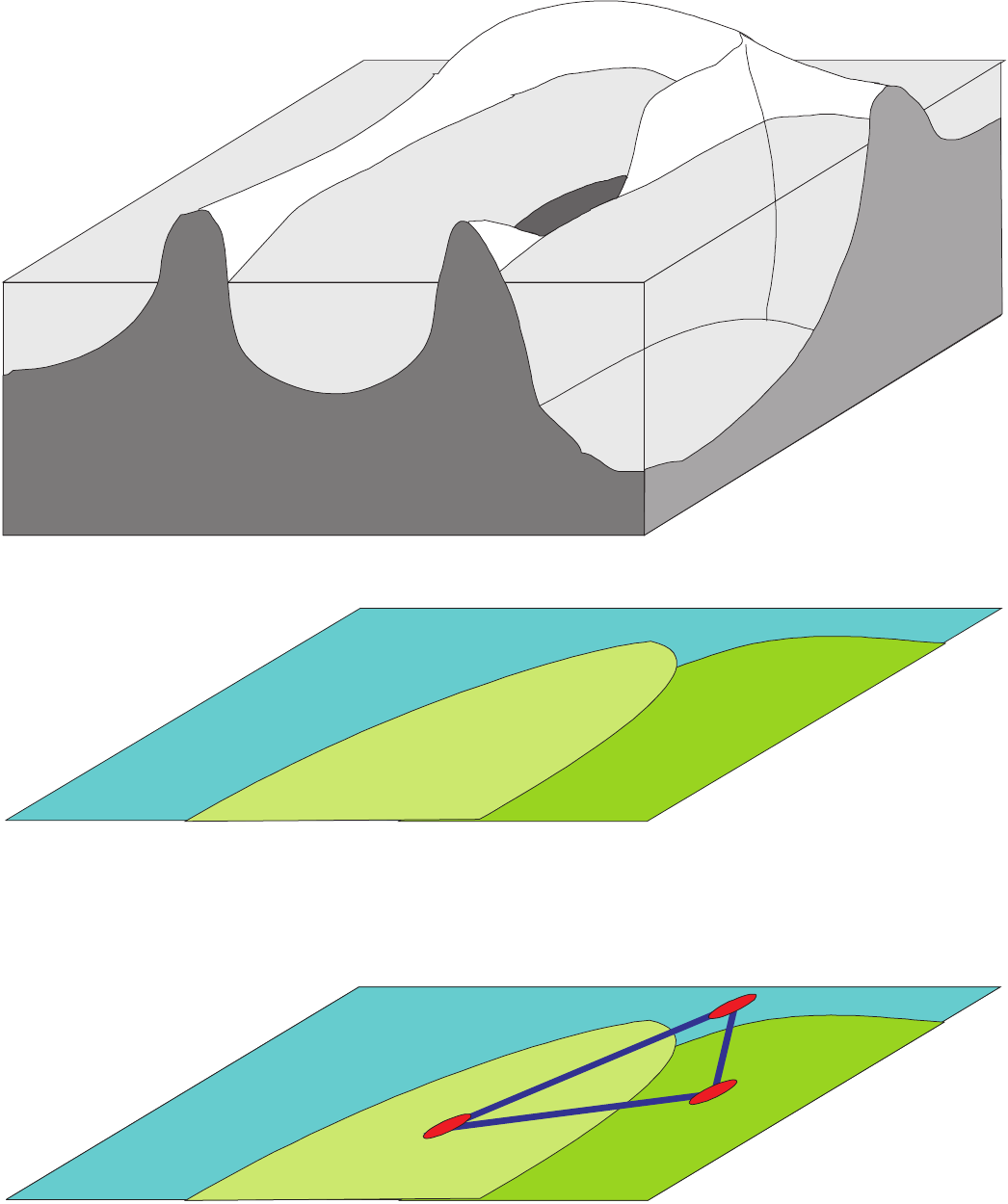}%
\caption{The region adjacency graph of a topographic surface}%
\label{neigh0}%
\end{center}
\end{figure}
%EndExpansion
%

%TCIMACRO{\TeXButton{EndFrame}{\end{frame}}}%
%BeginExpansion
\end{frame}%
%EndExpansion
%

%TCIMACRO{\TeXButton{BeginFrame}{\begin{frame}}}%
%BeginExpansion
\begin{frame}%
%EndExpansion
%

%TCIMACRO{\QTR{frametitle}{Representation of a flooded topographic surface as a
%node weighted graph. }}%
%BeginExpansion
\frametitle{Representation of a flooded topographic surface as a node weighted
graph. }%
%EndExpansion

An image may be considered as a topographic surface.\ The altitude of each
pixel corresponds to its gray level. An image may be modelled by a graph, the
nodes being the pixels and the edges connecting neighboring pixels. A first
weight distribution $f$ represents the ground level.

For a flooded surface, the nodes hold a second weight $\tau\geq f$ equal to
the flooding level.

The edges are not weighted.%

%TCIMACRO{\TeXButton{EndFrame}{\end{frame}}}%
%BeginExpansion
\end{frame}%
%EndExpansion
%

%TCIMACRO{\TeXButton{BeginFrame}{\begin{frame}}}%
%BeginExpansion
\begin{frame}%
%EndExpansion
%

%TCIMACRO{\QTR{frametitle}{Representation of a flooded RAG as an edge weighted
%graph}}%
%BeginExpansion
\frametitle{Representation of a flooded RAG as an edge weighted graph}%
%EndExpansion

A physical interpretation of a flooded RAG: the nodes are tanks with infinite
height and depth, their weight represent the height of the flooding in the
tank. If two nodes are connected by a weighted edge, the corresponding tanks
are linked by a pipe at an altitude of the weight.\ The pipes allow the water
to pass from tank to tank, according the laws of hydrostatics.We call such a
graph tank network (TN).\ The level in each tank is indicated in blue.\ %

%TCIMACRO{\FRAME{ftbpFU}{1.7783in}{1.3997in}{0pt}{\Qcb{Tank and pipe
%network:\newline- A and B form a regional minimum with $\tau_{A}=\tau
%_{B}=\lambda$ ; $e_{AB}\leq\lambda$ ; $e_{BC}>\lambda$\newline- B and C have
%unequal levels but are separated by a higher pipe.\newline- D and E form a
%full lake, reaching the level of its lowest exhaust pipe $e_{CD}$ \newline- E
%and F have the same level ; however they do not form a lake, as they are
%linked by a pipe which is higher}}{\Qlb{inond3}}{inond3-eps-converted-to.pdf}%
%width 1.7783in;  height 1.3997in;  depth 0pt;  original-width 3.1808in;
%original-height 2.4967in;  cropleft "0";  croptop "1";  cropright "1";
%cropbottom "0";  filename 'inond3-eps-converted-to.pdf';file-properties "XNPEU";}}}%
%BeginExpansion
\begin{figure}
[ptb]
\begin{center}
\includegraphics[
height=1.3997in,
width=1.7783in
]%
{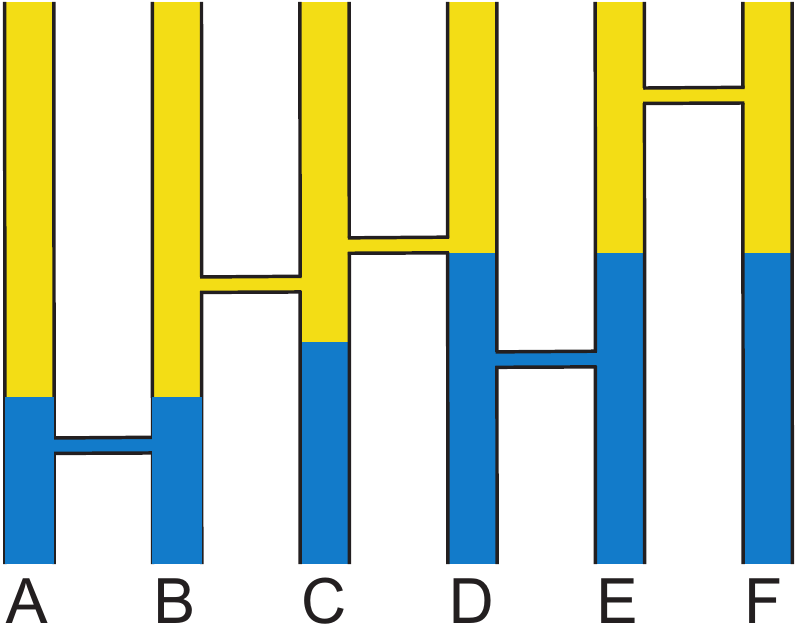}%
\caption{Tank and pipe network:\newline- A and B form a regional minimum with
$\tau_{A}=\tau_{B}=\lambda$ ; $e_{AB}\leq\lambda$ ; $e_{BC}>\lambda$\newline-
B and C have unequal levels but are separated by a higher pipe.\newline- D and
E form a full lake, reaching the level of its lowest exhaust pipe $e_{CD}$
\newline- E and F have the same level ; however they do not form a lake, as
they are linked by a pipe which is higher}%
\label{inond3}%
\end{center}
\end{figure}
%EndExpansion
%

%TCIMACRO{\TeXButton{EndFrame}{\end{frame}}}%
%BeginExpansion
\end{frame}%
%EndExpansion
%

%TCIMACRO{\TeXButton{BeginFrame}{\begin{frame}}}%
%BeginExpansion
\begin{frame}%
%EndExpansion
%

%TCIMACRO{\QTR{frametitle}{Flooding a topographic surface of its region
%adjacency graph}}%
%BeginExpansion
\frametitle{Flooding a topographic surface of its region adjacency graph}%
%EndExpansion
%

%TCIMACRO{\FRAME{ftbpFU}{3.6106in}{1.1665in}{0pt}{\Qcb{Flooding a topographic
%surface or flooding its region adjacency graph.\ }}{\Qlb{flood3}}%
%{flood3-eps-converted-to.pdf}{\special{ language "Scientific Word";  type "GRAPHIC";
%width 3.6106in;  height 1.1665in;  depth 0pt;  original-width 14.9771in;
%original-height 3.0676in;  cropleft "0";  croptop "1";  cropright "1";
%cropbottom "0";  filename 'flood3-eps-converted-to.pdf';file-properties "XNPEU";}}}%
%BeginExpansion
\begin{figure}
[ptb]
\begin{center}
\includegraphics[
height=1.1665in,
width=3.6106in
]%
{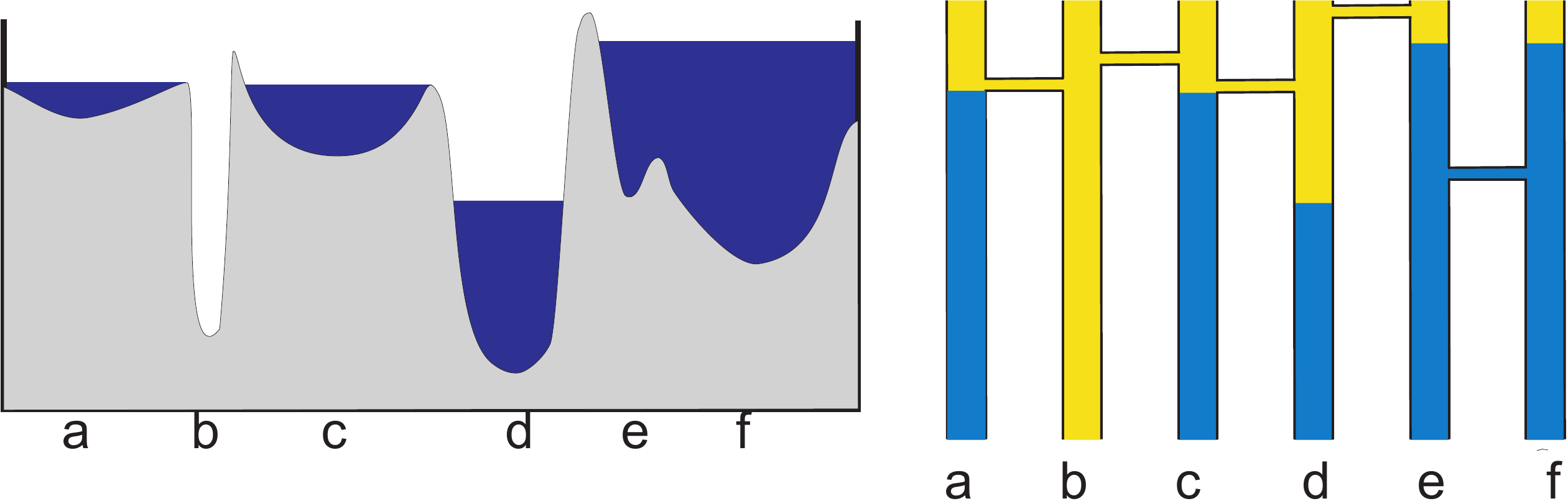}%
\caption{Flooding a topographic surface or flooding its region adjacency
graph.\ }%
\label{flood3}%
\end{center}
\end{figure}
%EndExpansion

The flooding of a topographic surface is perfectly defined if one knows the
flooding level in or above each catchment basin.\ The same flooding may be
represented on the region adjacency graph by assigning to each node the
flooding level in the corresponding basin.\ %

%TCIMACRO{\TeXButton{EndFrame}{\end{frame}}}%
%BeginExpansion
\end{frame}%
%EndExpansion
{}%

%TCIMACRO{\TeXButton{BeginFrame}{\begin{frame}}}%
%BeginExpansion
\begin{frame}%
%EndExpansion

\begin{center}
{\Large \alert{Modelling the laws of hydrostatics in node and edge weighted
tanks}}

\bigskip

As flooding a topographic surface and flooding its RAG represent the same
phenomenon, we have to find two models, one for node weighted graphs, the
other for edge weighted graphs, expressing this same phenomenon
\end{center}

%

%TCIMACRO{\TeXButton{EndFrame}{\end{frame}}}%
%BeginExpansion
\end{frame}%
%EndExpansion
%

%TCIMACRO{\TeXButton{BeginFrame}{\begin{frame}}}%
%BeginExpansion
\begin{frame}%
%EndExpansion

\begin{center}
{\Large \alert{Flooding a topographic surface or nodes weighted graph}}

\bigskip
\end{center}

%

%TCIMACRO{\TeXButton{EndFrame}{\end{frame}}}%
%BeginExpansion
\end{frame}%
%EndExpansion

\subsection{Flooding a topographic surface}%

%TCIMACRO{\TeXButton{BeginFrame}{\begin{frame}}}%
%BeginExpansion
\begin{frame}%
%EndExpansion
%

%TCIMACRO{\QTR{frametitle}{Definition of a flooding of a function}}%
%BeginExpansion
\frametitle{Definition of a flooding of a function}%
%EndExpansion

\begin{definition}
A function $g$ is a flooding of a function $f$ if and only if $g\geq f$ and
for any couple of neighboring pixels $(p,q)$ : $g_{p}>g_{q}\Rightarrow
g_{p}=f_{p}$
\end{definition}

Fig.\ \ref{wfal20}A presents a physically possible flooding.\ On the contrary
the flooding in fig. \ref{wfal20}B is impossible, as the lake containing the
pixel $p$ where $g_{p}>f_{p}$ is not limited by solid ground since
$g_{p}>g_{q}.\ $%

%TCIMACRO{\FRAME{ftbpFU}{3.188in}{0.8344in}{0pt}{\Qcb{A possible and an
%impossible flooding}}{\Qlb{wfal20}}{wfal20-eps-converted-to.pdf}%
%width 3.188in;  height 0.8344in;  depth 0pt;  original-width 3.0844in;
%original-height 0.7874in;  cropleft "0";  croptop "1";  cropright "1";
%cropbottom "0";  filename 'wfal20-eps-converted-to.pdf';file-properties "XNPEU";}}}%
%BeginExpansion
\begin{figure}
[ptb]
\begin{center}
\includegraphics[
height=0.8344in,
width=3.188in
]%
{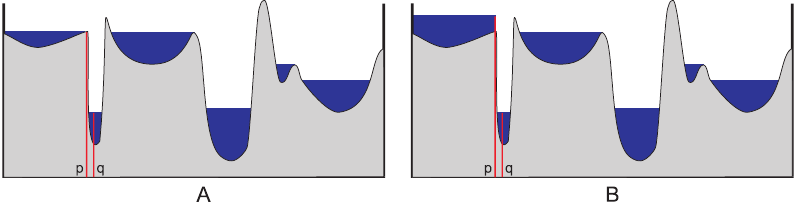}%
\caption{A possible and an impossible flooding}%
\label{wfal20}%
\end{center}
\end{figure}
%EndExpansion
%

%TCIMACRO{\TeXButton{EndFrame}{\end{frame}}}%
%BeginExpansion
\end{frame}%
%EndExpansion
{}

\subsection{Flooding a node weighted graph}%

%TCIMACRO{\TeXButton{BeginFrame}{\begin{frame}}}%
%BeginExpansion
\begin{frame}%
%EndExpansion
%

%TCIMACRO{\QTR{frametitle}{Flooding a node weighted graph}}%
%BeginExpansion
\frametitle{Flooding a node weighted graph}%
%EndExpansion

Images are particular node weighted graphs, the pixels being the
nodes.\ Neighboring pixels are linked by an unweighted edge.\ 

We now consider arbitrary node weighted graphs.\ The node weights $f_{i}$
indicate the ground level.\ The edges are not weighted.

Such a topographic graph is flooded if the nodes are assigned a second family
of weights indicating the level of the flooding at each node.\ %

%TCIMACRO{\TeXButton{EndFrame}{\end{frame}}}%
%BeginExpansion
\end{frame}%
%EndExpansion

\subsection{Flooding a node weighted graph}%

%TCIMACRO{\TeXButton{BeginFrame}{\begin{frame}}}%
%BeginExpansion
\begin{frame}%
%EndExpansion
%

%TCIMACRO{\QTR{frametitle}{Flooding a node weighted graph}}%
%BeginExpansion
\frametitle{Flooding a node weighted graph}%
%EndExpansion

A distribution $\tau$ of node weights will represent an effective flooding if
it verifies a number of conditions of equilibrium:

\begin{itemize}
\item A flooding being always above the ground level: $\tau_{i}\geq f_{i}.$

\item As there is nothing to prevent the water to flow from a higher to a
lower position, an inequal level of water at two neighboring nodes $p$ and $q
$ is impossible, except when the highest node is dry ; hence $\tau_{p}%
>\tau_{q}\Rightarrow\tau_{p}=f_{p}$ indicating that the highest level is dry,
without water.

\item Consequence 1: in a lake, the level of all nodes is the same.\ 

\item Consequence 2: floodings are connected operators : $f_{p}=f_{q}%
\Rightarrow\tau_{p}=\tau_{q}$
\end{itemize}

%

%TCIMACRO{\TeXButton{EndFrame}{\end{frame}}}%
%BeginExpansion
\end{frame}%
%EndExpansion
%

%TCIMACRO{\TeXButton{BeginFrame}{\begin{frame}}}%
%BeginExpansion
\begin{frame}%
%EndExpansion

\begin{center}
{\Large \alert{Flooding an edge weighted graph}}

\bigskip
\end{center}

%

%TCIMACRO{\TeXButton{EndFrame}{\end{frame}}}%
%BeginExpansion
\end{frame}%
%EndExpansion

\subsection{Flooding an edge weighted graph}%

%TCIMACRO{\TeXButton{BeginFrame}{\begin{frame}}}%
%BeginExpansion
\begin{frame}%
%EndExpansion
%

%TCIMACRO{\QTR{frametitle}{Flooding an edge weighted graph}}%
%BeginExpansion
\frametitle{Flooding an edge weighted graph}%
%EndExpansion

$G=[E,N]$ : a node and edge weighted graph, $E$ = edges, $N$ = nodes.\ 

The edges are weighted: the weight $e_{ij}$ of the edge $(i,j)$ represents the
altitude at which a flood coming from one extremity may reach the other
extremity of the edge.\ 

The nodes also are weighted; $\tau_{i}$ represents the altitude of the flood
at node $i.$%

%TCIMACRO{\TeXButton{EndFrame}{\end{frame}}}%
%BeginExpansion
\end{frame}%
%EndExpansion

\subsection{Flooding an edge weighted graph}%

%TCIMACRO{\TeXButton{BeginFrame}{\begin{frame}}}%
%BeginExpansion
\begin{frame}%
%EndExpansion
%

%TCIMACRO{\QTR{frametitle}{Flooding an edge weighted graph}}%
%BeginExpansion
\frametitle{Flooding an edge weighted graph}%
%EndExpansion
%

%TCIMACRO{\FRAME{ftbpFU}{1.7783in}{1.3997in}{0pt}{\Qcb{{}}}{}{inond3-eps-converted-to.pdf}%
%width 1.7783in;  height 1.3997in;  depth 0pt;  original-width 3.1808in;
%original-height 2.4967in;  cropleft "0";  croptop "1";  cropright "1";
%cropbottom "0";  filename 'inond3-eps-converted-to.pdf';file-properties "XNPEU";}}}%
%BeginExpansion
\begin{figure}
[ptb]
\begin{center}
\includegraphics[
height=1.3997in,
width=1.7783in
]%
{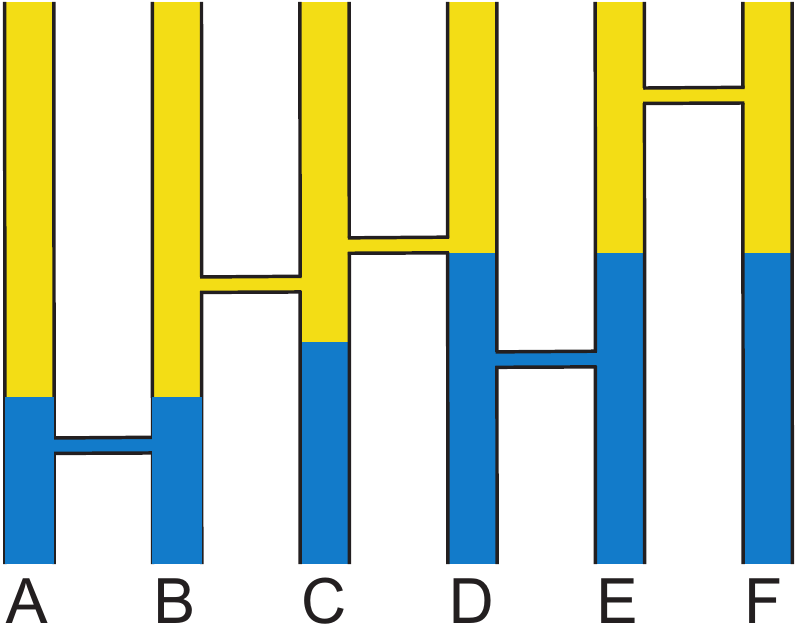}%
\caption{{}}%
\end{center}
\end{figure}
%EndExpansion

We consider the nodes as vertical tanks of infinite height and depth : there
is no ground level.\ 

The weight $\tau_{i}$ represents the level of water in the tank $i.\ $

Two neighboring tanks $i$ and $j$ are linked by a pipe at an altitude $e_{ij}
$ equal to the weight of the edge.\ 

We call such an edge weighted graph a tank network.%

%TCIMACRO{\TeXButton{EndFrame}{\end{frame}}}%
%BeginExpansion
\end{frame}%
%EndExpansion

\subsection{Flooding an edge weighted graph}%

%TCIMACRO{\TeXButton{BeginFrame}{\begin{frame}}}%
%BeginExpansion
\begin{frame}%
%EndExpansion
%

%TCIMACRO{\QTR{frametitle}{Flooding an edge weighted graph}}%
%BeginExpansion
\frametitle{Flooding an edge weighted graph}%
%EndExpansion
%

%TCIMACRO{\FRAME{ftbpFU}{1.7783in}{1.3997in}{0pt}{\Qcb{}}{\Qlb{inond3}%
%}{inond3-eps-converted-to.pdf}{\special{ language "Scientific Word";  type "GRAPHIC";
%width 1.7783in;  height 1.3997in;  depth 0pt;  original-width 3.1808in;
%original-height 2.4967in;  cropleft "0";  croptop "1";  cropright "1";
%cropbottom "0";  filename 'inond3-eps-converted-to.pdf';file-properties "XNPEU";}}}%
%BeginExpansion
\begin{figure}
[ptb]
\begin{center}
\includegraphics[
height=1.3997in,
width=1.7783in
]%
{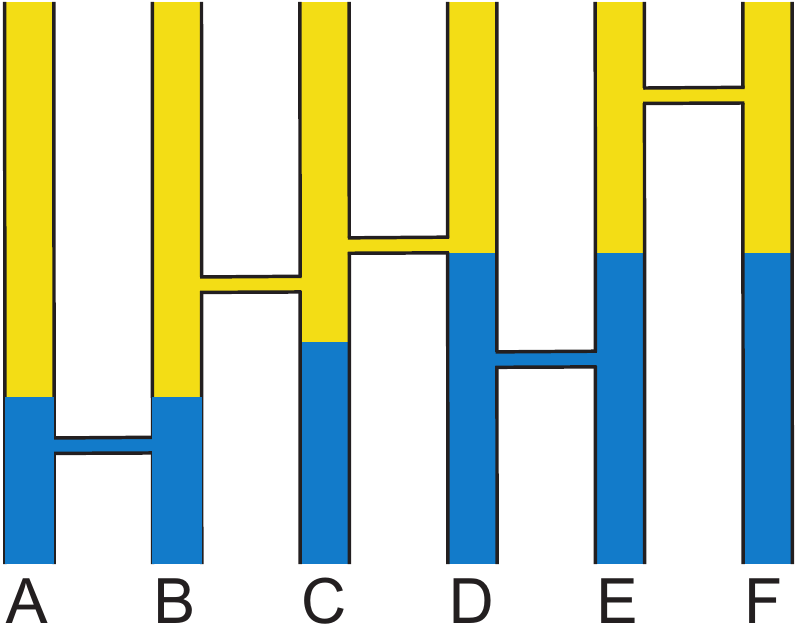}%
\label{inond3}%
\end{center}
\end{figure}
%EndExpansion

Laws of hydrostatics:

\begin{itemize}
\item if the level $\tau_{i}$ in the tank $i$ is higher than the pipe
$e_{ij},$ then the levels is the same in both tanks $i$ and $j:$ $\tau
_{i}=\tau_{j}.$

\item the level $\tau_{i}$ in the tank $i$ cannot be higher than the level
$\tau_{j},$ unless $e_{ij}\geq\tau_{i}$.\ 
\end{itemize}

%

%TCIMACRO{\TeXButton{Transition: Box Out}{\transboxout}}%
%BeginExpansion
\transboxout
%EndExpansion%
%TCIMACRO{\TeXButton{EndFrame}{\end{frame}}}%
%BeginExpansion
\end{frame}%
%EndExpansion

\subsection{Flooding an edge weighted graph}%

%TCIMACRO{\TeXButton{BeginFrame}{\begin{frame}}}%
%BeginExpansion
\begin{frame}%
%EndExpansion
%

%TCIMACRO{\QTR{frametitle}{Flooding an edge weighted graph}}%
%BeginExpansion
\frametitle{Flooding an edge weighted graph}%
%EndExpansion
%

%TCIMACRO{\FRAME{ftbpFU}{1.7783in}{1.3997in}{0pt}{\Qcb{Tank and pipe
%network:\newline- A and B form a regional minimum with $\tau_{A}=\tau
%_{B}=\lambda$ ; $e_{AB}\leq\lambda$ ; $e_{BC}>\lambda$\newline- B and C have
%unequal levels but are separated by a higher pipe.\newline- D and E form a
%full lake, reaching the level of its lowest exhaust pipe $e_{CD}$ \newline- E
%and F have the same level ; however they do not form a lake, as they are
%linked by a pipe which is higher}}{\Qlb{inond3}}{inond3-eps-converted-to.pdf}%
%width 1.7783in;  height 1.3997in;  depth 0pt;  original-width 3.1808in;
%original-height 2.4967in;  cropleft "0";  croptop "1";  cropright "1";
%cropbottom "0";  filename 'inond3-eps-converted-to.pdf';file-properties "XNPEU";}}}%
%BeginExpansion
\begin{figure}
[ptb]
\begin{center}
\includegraphics[
height=1.3997in,
width=1.7783in
]%
{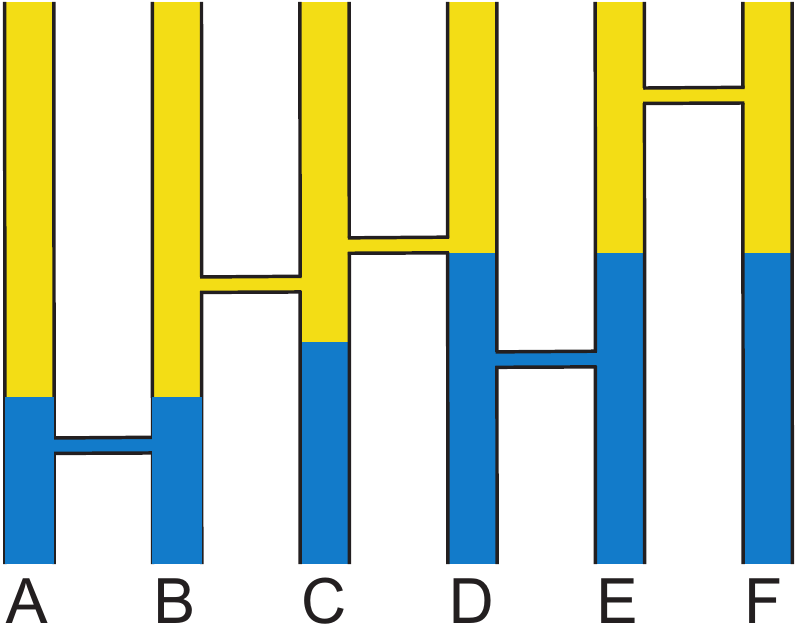}%
\caption{Tank and pipe network:\newline- A and B form a regional minimum with
$\tau_{A}=\tau_{B}=\lambda$ ; $e_{AB}\leq\lambda$ ; $e_{BC}>\lambda$\newline-
B and C have unequal levels but are separated by a higher pipe.\newline- D and
E form a full lake, reaching the level of its lowest exhaust pipe $e_{CD}$
\newline- E and F have the same level ; however they do not form a lake, as
they are linked by a pipe which is higher}%
\label{inond3}%
\end{center}
\end{figure}
%EndExpansion
%

%TCIMACRO{\TeXButton{Transition: Box Out}{\transboxout}}%
%BeginExpansion
\transboxout
%EndExpansion%
%TCIMACRO{\TeXButton{EndFrame}{\end{frame}}}%
%BeginExpansion
\end{frame}%
%EndExpansion

\subsection{Flooding an edge weighted graph}%

%TCIMACRO{\TeXButton{BeginFrame}{\begin{frame}}}%
%BeginExpansion
\begin{frame}%
%EndExpansion
%

%TCIMACRO{\QTR{frametitle}{Flooding an edge weighted graph}}%
%BeginExpansion
\frametitle{Flooding an edge weighted graph}%
%EndExpansion

\begin{definition}
The distribution $\tau$ of water in the pipes of the graph $[E,N]$ is a
flooding of this graph, i.e. is a stable distribution of fluid if it verifies
the criterion:\ \newline for any couple of neighboring nodes $(p,q)$ we have:
$(\tau_{p}>\tau_{q}\Rightarrow e_{pq}\geq\tau_{p})\qquad(criterion\ 1)$
\end{definition}

%

%TCIMACRO{\FRAME{ftbpFU}{3.0834in}{1.4197in}{0pt}{\Qcb{The water distribution
%marked OK are compatible with the laws of physics ; the others are not.}%
%}{\Qlb{inond4}}{inond4-eps-converted-to.pdf}{\special{ language "Scientific Word";
%type "GRAPHIC";  maintain-aspect-ratio TRUE;  display "USEDEF";
%valid_file "F";  width 3.0834in;  height 1.4197in;  depth 0pt;
%original-width 5.5564in;  original-height 2.5313in;  cropleft "0";
%filename 'inond4-eps-converted-to.pdf';file-properties "XNPEU";}}}%
%BeginExpansion
\begin{figure}
[ptb]
\begin{center}
\includegraphics[
height=1.4197in,
width=3.0834in
]%
{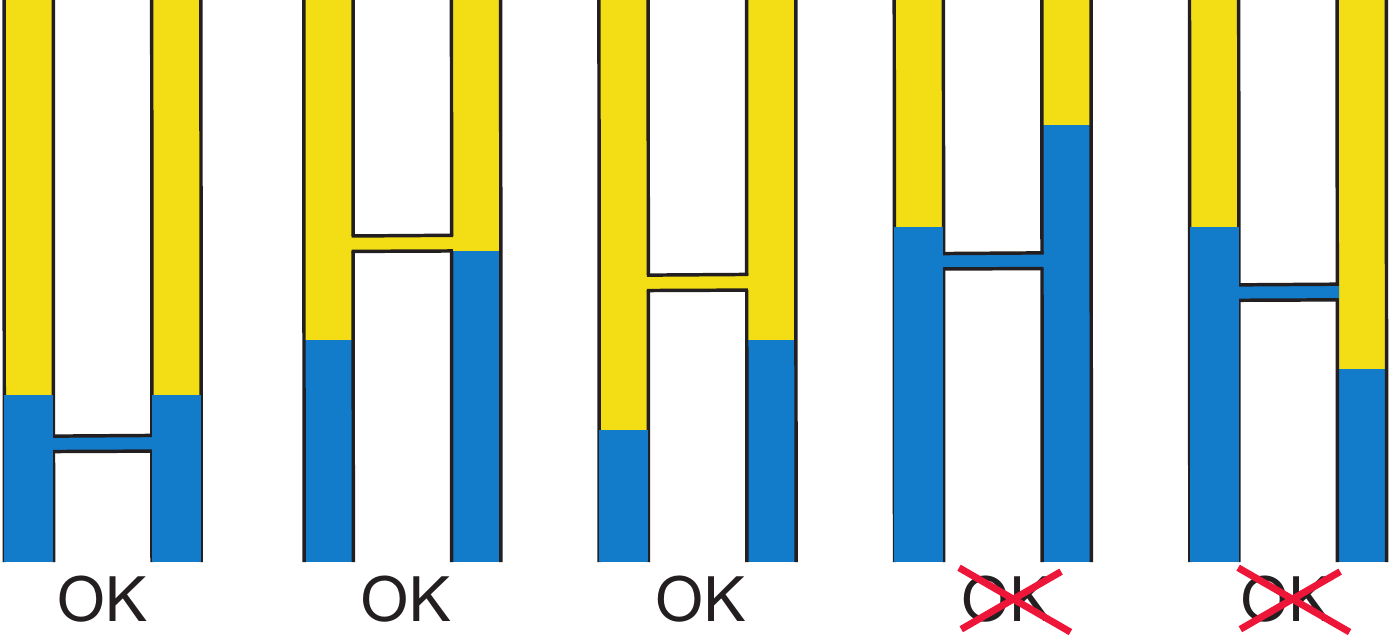}%
\caption{The water distribution marked OK are compatible with the laws of
physics ; the others are not.}%
\label{inond4}%
\end{center}
\end{figure}
%EndExpansion
%

%TCIMACRO{\TeXButton{Transition: Box Out}{\transboxout}}%
%BeginExpansion
\transboxout
%EndExpansion%
%TCIMACRO{\TeXButton{EndFrame}{\end{frame}}}%
%BeginExpansion
\end{frame}%
%EndExpansion

\subsection{Flooding an edge weighted graph : criteria}%

%TCIMACRO{\TeXButton{BeginFrame}{\begin{frame}}}%
%BeginExpansion
\begin{frame}%
%EndExpansion
%

%TCIMACRO{\QTR{frametitle}{Criteria}}%
%BeginExpansion
\frametitle{Criteria}%
%EndExpansion

$(\tau_{p}>\tau_{q}\Rightarrow e_{pq}\geq\tau_{p})\Leftrightarrow($not
$(\tau_{p}>\tau_{q})\ $or $e_{pq}\geq\tau_{p})\Leftrightarrow(\tau_{p}\leq
\tau_{q}\ $or $\tau_{p}\leq e_{pq})\Leftrightarrow$

$(\tau_{p}\leq\tau_{q}\vee e_{pq})\qquad(criterion\ 2)$

$\Leftrightarrow(\tau_{p}\leq\tau_{q}\vee%
%TCIMACRO{\tbigwedge \limits_{(p,q)\text{ neighbors}}}%
%BeginExpansion
{\textstyle\bigwedge\limits_{(p,q)\text{ neighbors}}}
%EndExpansion
\left(  \tau_{q}\vee e_{pq}\right)  \qquad(criterion\ 3)$

\begin{remark}
The criterion $(\tau_{p}>\tau_{q}\Rightarrow e_{pq}\geq\tau_{p})$ is
equivalent with $(e_{pq}<\tau_{p}\Rightarrow\tau_{p}\leq\tau_{q}).\ $Hence if
$e_{pq}<\tau_{p},$ we have $\tau_{p}\leq\tau_{q}$ ; so we also have
$e_{pq}<\tau_{q}$ implying $\tau_{q}\leq\tau_{p}$ ; finally $\tau_{p}=\tau
_{q}.$
\end{remark}

%

%TCIMACRO{\TeXButton{EndFrame}{\end{frame}}}%
%BeginExpansion
\end{frame}%
%EndExpansion
%

%TCIMACRO{\TeXButton{BeginFrame}{\begin{frame}}}%
%BeginExpansion
\begin{frame}%
%EndExpansion

\begin{center}
{\Large \alert{Flooding a topographic graph is the same as flooding an
associated edge weighted graph}}

\bigskip
\end{center}

%

%TCIMACRO{\TeXButton{EndFrame}{\end{frame}}}%
%BeginExpansion
\end{frame}%
%EndExpansion

\subsection{Flooding a TG is the same as flooding a TN}%

%TCIMACRO{\TeXButton{BeginFrame}{\begin{frame}}}%
%BeginExpansion
\begin{frame}%
%EndExpansion
%

%TCIMACRO{\QTR{frametitle}{Flooding a TG is the same as flooding a TN}}%
%BeginExpansion
\frametitle{Flooding a TG is the same as flooding a TN}%
%EndExpansion

$G_{n}=[E,N]$ : a topographic graph.\ Ground level = $f.\ $The edges are not weighted.\ 

The lowest level of flood covering two neighboring nodes $p$ and $q$ is equal
to $f_{p}\vee f_{q}$.\ 

Consider now a second graph $G_{e}$ with the same structure but with edge
weights $e_{pq}=f_{p}\vee f_{q}.$

Any flooding $\tau\geq f$ of $G_{e}$ verifies: for $(p,q)$ neighbors
$(\tau_{p}\leq\tau_{q}\vee e_{pq})\Leftrightarrow(\tau_{p}\leq\tau_{q}\vee
f_{q}\vee f_{p})\Leftrightarrow(\tau_{p}\leq\tau_{q}\vee f_{p})$ as $\tau
_{q}\geq f_{q}$

But this last criterion characterizes a flooding of $G_{n}.$

\begin{theorem}
There is an equivalence between the floodings $\tau\geq f$ of $G_{e}$ of
$G_{n}$ and the floodings $\tau\geq f$ of $G_{e}$, with edge weights
$e_{pq}=f_{p}\vee f_{q}$ .
\end{theorem}

%

%TCIMACRO{\TeXButton{EndFrame}{\end{frame}}}%
%BeginExpansion
\end{frame}%
%EndExpansion
%

%TCIMACRO{\TeXButton{BeginFrame}{\begin{frame}}}%
%BeginExpansion
\begin{frame}%
%EndExpansion

\begin{center}
{\Large \alert{Lakes of edge weighted graphs}}
\end{center}

%

%TCIMACRO{\TeXButton{EndFrame}{\end{frame}}}%
%BeginExpansion
\end{frame}%
%EndExpansion
%

%TCIMACRO{\TeXButton{BeginFrame}{\begin{frame}}}%
%BeginExpansion
\begin{frame}%
%EndExpansion
%

%TCIMACRO{\QTR{frametitle}{Lakes of edge weighted graphs }}%
%BeginExpansion
\frametitle{Lakes of edge weighted graphs }%
%EndExpansion

If a node $p$ has no neighboring node $q$ such that $\tau_{p}=\tau_{q},$ then
$p$ is an isolated node and isolated lake.\ 

Consider now two neighboring nodes $p$ and $q$ verifying $\tau_{p}=\tau
_{q}.\ $Adding a drop of water at the node $p$ has no impact on node $q,$ if
there exists no path linking $p$ and $q$ with edge weights $\leq\tau_{p}%
,\tau_{q}.$

We define a binary relation between neighboring pixels $p,q:p\thicksim
q\Leftrightarrow\tau_{p}=\tau_{q}$ and $e_{pq}\leq\tau_{p},\tau_{q}.\ $

\begin{lemma}
If we cut all edges which do not verify $p\thicksim q,$ we get a partial graph
$\widetilde{G}$ ; the connected components of $\widetilde{G}$ are the lakes of
the graph $G$ $.\ $
\end{lemma}

%

%TCIMACRO{\TeXButton{EndFrame}{\end{frame}}}%
%BeginExpansion
\end{frame}%
%EndExpansion
{}%

%TCIMACRO{\TeXButton{BeginFrame}{\begin{frame}}}%
%BeginExpansion
\begin{frame}%
%EndExpansion

\begin{center}
{\Large \alert{Lakes of node weighted graphs}}
\end{center}

%

%TCIMACRO{\TeXButton{EndFrame}{\end{frame}}}%
%BeginExpansion
\end{frame}%
%EndExpansion
%

%TCIMACRO{\TeXButton{BeginFrame}{\begin{frame}}}%
%BeginExpansion
\begin{frame}%
%EndExpansion
%

%TCIMACRO{\QTR{frametitle}{Lakes of node weighted graphs }}%
%BeginExpansion
\frametitle{Lakes of node weighted graphs }%
%EndExpansion

Consider now a topographic graph $G_{n}$ with a ground level $f$ and its
derived tank network $G_{e}$ with edge weights $e_{pq}=f_{p}\vee f_{q}$.\ Any
flooding of $G_{n}$ also is a flooding of $G_{e}.\ $Applying the definition of
lakes given above we distinguish two cases:

\begin{itemize}
\item $p$ is an isolated node : it has no neighboring node $q$ such that
$\tau_{p}=\tau_{q}$

\item $p$ is not isolated, and has at least one neighbor $q$ such that
$\tau_{p}=\tau_{q}.\ $As $\tau\geq f,$ we have $\tau_{p}=\tau_{q}=\tau_{p}%
\vee\tau_{q}\geq f_{p}\vee f_{q}=e_{pq}.\ $This shows that $\tau_{p}=\tau
_{q}\Rightarrow p\thicksim q.\ $This shows that the lakes of $G_{n}$ simply
are its flat zones.\ 
\end{itemize}

\begin{definition}
\textbf{\ }The lakes of a TG are its flat zones, that is maximal connected
components of nodes with the same altitude.
\end{definition}

%

%TCIMACRO{\TeXButton{EndFrame}{\end{frame}}}%
%BeginExpansion
\end{frame}%
%EndExpansion
{}%

%TCIMACRO{\TeXButton{BeginFrame}{\begin{frame}}}%
%BeginExpansion
\begin{frame}%
%EndExpansion
%

%TCIMACRO{\QTR{frametitle}{Lakes of node weighted graphs }}%
%BeginExpansion
\frametitle{Lakes of node weighted graphs }%
%EndExpansion

A lake on a topographic graph is dry if it has a uniform altitude at the
ground level.\ It is a wet lake, if it contains at least one pixel $p$ for
which $\tau_{p}>f_{p}$.\ The two following lemmas concern wet lakes.\ The
first is a reinterpretation of a lemma established for TN.\ 

\begin{lemma}
If two neighboring nodes $p$ and $q$ verify $\tau_{p}>e_{pq}=f_{p}\vee f_{q},
$ then $\tau_{p}=\tau_{q}.$
\end{lemma}

The second derives from criterion TG-1.

\begin{lemma}
If two neighboring nodes $p$ and $q$ verify $\tau_{p}>f_{p}$ and $\tau
_{q}>f_{q},$ then $\tau_{p}=\tau_{q}.$
\end{lemma}

\textbf{Proof: }$\left\{  \tau_{p}>\tau_{q}\Rightarrow\tau_{p}\leq
f_{p}\right\}  \Leftrightarrow\left\{  \tau_{p}>f_{p}\Rightarrow\tau_{p}%
\leq\tau_{q}\right\}  .\ $Applying the last implication to $\tau_{p}>f_{p}$
and $\tau_{q}>f_{q}$ yields $\tau_{p}\leq\tau_{q}$ and $\tau_{p}\geq\tau_{q},$
which together gives $\tau_{p}=\tau_{q}.$%

%TCIMACRO{\TeXButton{EndFrame}{\end{frame}}}%
%BeginExpansion
\end{frame}%
%EndExpansion
%

%TCIMACRO{\TeXButton{BeginFrame}{\begin{frame}}}%
%BeginExpansion
\begin{frame}%
%EndExpansion

\begin{center}
{\Large \alert{Regional minima lakes and full lakes}}
\end{center}

%

%TCIMACRO{\TeXButton{EndFrame}{\end{frame}}}%
%BeginExpansion
\end{frame}%
%EndExpansion
%

%TCIMACRO{\TeXButton{BeginFrame}{\begin{frame}}}%
%BeginExpansion
\begin{frame}%
%EndExpansion
%

%TCIMACRO{\QTR{frametitle}{Full lakes and regional minimum lakes in a tank
%network}}%
%BeginExpansion
\frametitle{Full lakes and regional minimum lakes in a tank network}%
%EndExpansion

What happens at the boundary of a lake $X$ in a tank network ?\ \ Consider 2
neighboring pixels $(p,q),$ $p$ being inside a lake of altitude $\lambda$ and
$q$ outside.\ These pixels do not verify $p\thicksim q:$ $e_{pq}>\tau_{p}$ and
$\tau_{p}\neq\tau_{q}:$

\begin{itemize}
\item if $e_{pq}>\tau_{p}$ and one has to climb for going from $p$ to $q.$

\item else $e_{pq}\leq\tau_{p}$ implying $\tau_{q}\leq\tau_{p}\vee e_{pq}%
=\tau_{p}.\ $As $\tau_{p}\neq\tau_{q}$ we have $\tau_{q}<\tau_{p,}$ which
implies $\tau_{p}\leq e_{pq}.\ $Thus $\tau_{p}=e_{pq}$ and $\tau_{q}<\tau
_{p,}$ indicating the $q$ is an exhaust node of the lake , and the lake $X$ is
a full lake.\ 
\end{itemize}

In other terms, in a lake without exhaust edges, all outgoing edges are higher
than the level of the lake. Such a lake is called regional minimum lake.\ A
lake with one or several exhautst edges is called full lake.\ Adding a drop of
water to a full lake provokes an overflow through the exhaust edges.

\begin{definition}
A regional minimum of a tank network is a lake with all outgoing edges, or
cocycle edges having a higher altitude.\ 
\end{definition}

%

%TCIMACRO{\TeXButton{EndFrame}{\end{frame}}}%
%BeginExpansion
\end{frame}%
%EndExpansion
%

%TCIMACRO{\TeXButton{BeginFrame}{\begin{frame}}}%
%BeginExpansion
\begin{frame}%
%EndExpansion
%

%TCIMACRO{\QTR{frametitle}{Full lakes and regional minimum lakes in a tank
%network}}%
%BeginExpansion
\frametitle{Full lakes and regional minimum lakes in a tank network}%
%EndExpansion

\begin{definition}
A lake of level $\lambda$ of the flooding of a tank network is a full lake, if
there exists an an exhaust edge from an inside node $p$ to an outside node $q$
verifying $\tau_{p}=e_{pq}=\lambda>\tau_{q}.$
\end{definition}

\begin{lemma}
Each regional minimum of the flooding of a tank network contains a regional
minimum of the tank network itself or is an isolated regional minimum.\ .
\end{lemma}

\textbf{Proof: }Either $X=\{p\}$ is an isolated regional minimum node $p$ with
all adjacent edges having a weight $>\lambda.\ $If $X$ contains inside edges,
and $(p,q)$ is the edge for which $e_{pq}$ is minimal, then the maximal
connected component containing $p$ with edge weights equal to $e_{pq} $ is a
regional minimum of the graph.%

%TCIMACRO{\TeXButton{EndFrame}{\end{frame}}}%
%BeginExpansion
\end{frame}%
%EndExpansion
%

%TCIMACRO{\TeXButton{BeginFrame}{\begin{frame}}}%
%BeginExpansion
\begin{frame}%
%EndExpansion
%

%TCIMACRO{\QTR{frametitle}{Full lakes and regional minimum lakes in topographic
%graph}}%
%BeginExpansion
\frametitle{Full lakes and regional minimum lakes in topographic graph}%
%EndExpansion

A couple of neighboring nodes belongs to the cocycle of a lake $X$, $p\in X$
and $q\notin X$ only if $\tau_{p}\neq\tau_{q}.$ Either $\tau_{p}>\tau_{q}%
.\ $Or $\tau_{p}>\tau_{q}$ implying $\tau_{p}=f_{p}$ and the lake is a full
lake having an exhaust node $p.\ $In the graph $G_{e}$ we have $e_{pq}%
=f_{p}\vee f_{q}=\tau_{p}$ as $f_{p}=\tau_{p}>\tau_{q}\geq f_{q}.$

We get the two following definitions.

\begin{definition}
\textbf{\ }A regional minimum is a lake for which the ground level of all
outside neighbors has a higher altitude.
\end{definition}

\begin{definition}
A lake of the flooding of a topographic surface is a full lake of altitude
$\lambda$ if there exist two neighboring nodes $p$ inside the lake and $q$
outside, such that $\tau_{p}<\tau_{q}=f_{q}.$
\end{definition}

Each regional minimum of the flooding of a topographic graph contains a
regional minimum of the topographic graph itself.%

%TCIMACRO{\TeXButton{EndFrame}{\end{frame}}}%
%BeginExpansion
\end{frame}%
%EndExpansion
%

%TCIMACRO{\TeXButton{BeginFrame}{\begin{frame}}}%
%BeginExpansion
\begin{frame}%
%EndExpansion

\begin{center}
{\Large \alert{Among all possible flooidngs, chosing one}}
\end{center}

%

%TCIMACRO{\TeXButton{EndFrame}{\end{frame}}}%
%BeginExpansion
\end{frame}%
%EndExpansion
%

%TCIMACRO{\TeXButton{BeginFrame}{\begin{frame}}}%
%BeginExpansion
\begin{frame}%
%EndExpansion
%

%TCIMACRO{\QTR{frametitle}{Specifying a particular flooding among all possible
%ones}}%
%BeginExpansion
\frametitle{Specifying a particular flooding among all possible ones}%
%EndExpansion

Many floodings of a topographic surface of of an edge weighted graph are
possible.\ In order to specify a particular flooding we have to add other
criteria. For instance the lowest flooding for which each lake is a full lake
or has a surface area higher than a given threshold specifies the so-called
area flooding. we are interested by the highest flooding under a ceiling
function $\omega.$%

%TCIMACRO{\FRAME{ftbpFU}{2.9689in}{1.5218in}{0pt}{\Qcb{Various flood
%distribution on the same topographic surface}}{\Qlb{flood5}}{flood5-eps-converted-to.pdf}%
%width 2.9689in;  height 1.5218in;  depth 0pt;  original-width 10.8916in;
%original-height 5.5317in;  cropleft "0";  croptop "1";  cropright "1";
%cropbottom "0";  filename 'flood5-eps-converted-to.pdf';file-properties "XNPEU";}}}%
%BeginExpansion
\begin{figure}
[ptb]
\begin{center}
\includegraphics[
height=1.5218in,
width=2.9689in
]%
{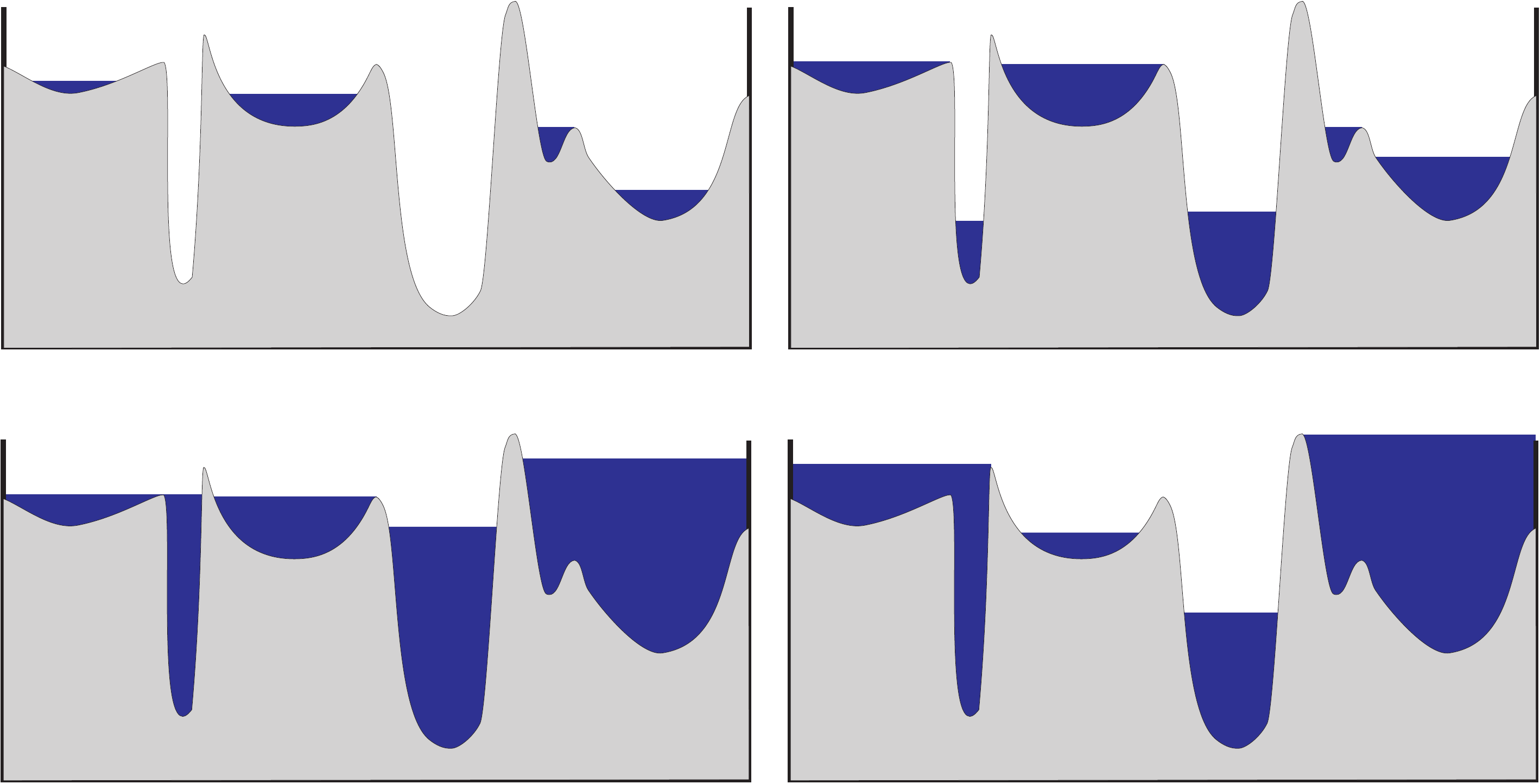}%
\caption{Various flood distribution on the same topographic surface}%
\label{flood5}%
\end{center}
\end{figure}
%EndExpansion
%

%TCIMACRO{\TeXButton{EndFrame}{\end{frame}}}%
%BeginExpansion
\end{frame}%
%EndExpansion
%

%TCIMACRO{\TeXButton{BeginFrame}{\begin{frame}}}%
%BeginExpansion
\begin{frame}%
%EndExpansion
%

%TCIMACRO{\QTR{frametitle}{The algebra of floodings}}%
%BeginExpansion
\frametitle{The algebra of floodings}%
%EndExpansion

\begin{lemma}
If $\tau$ and $\nu$ are two floodings of a node or edge weighted graph $G$,
then $\tau\vee\nu$ and $\tau\wedge\nu$ also are floodings of $G.$
\end{lemma}

Hence the family of floodings of the graph $G_{e}$ or $G_{n}$ below\ the
function $\omega$ is closed by supremum.\ This supremum is itself a flooding
and is below $\omega.\ $For this reason it is the highest flooding of the
graph $G_{e}$ or $G_{n}$ below $\omega.$%

%TCIMACRO{\TeXButton{EndFrame}{\end{frame}}}%
%BeginExpansion
\end{frame}%
%EndExpansion
%

%TCIMACRO{\TeXButton{BeginFrame}{\begin{frame}}}%
%BeginExpansion
\begin{frame}%
%EndExpansion
%

%TCIMACRO{\QTR{frametitle}{Illustration}}%
%BeginExpansion
\frametitle{Illustration}%
%EndExpansion
%

%TCIMACRO{\FRAME{ftbpFU}{3.3881in}{1.0718in}{0pt}{\Qcb{The highest flooding of
%a topographic surface below a ceiling functino (in red). The ceiling function
%on the left and on the right yield the same flooding, as they constrain the
%level regional minima lakes at identical levels ; all other lakes being full
%lakes.}}{\Qlb{flood6}}{flood6-eps-converted-to.pdf}{\special{ language "Scientific Word";
%type "GRAPHIC";  maintain-aspect-ratio TRUE;  display "USEDEF";
%valid_file "F";  width 3.3881in;  height 1.0718in;  depth 0pt;
%original-width 10.8393in;  original-height 3.3657in;  cropleft "0";
%filename 'flood6-eps-converted-to.pdf';file-properties "XNPEU";}}}%
%BeginExpansion
\begin{figure}
[ptb]
\begin{center}
\includegraphics[
height=1.0718in,
width=3.3881in
]%
{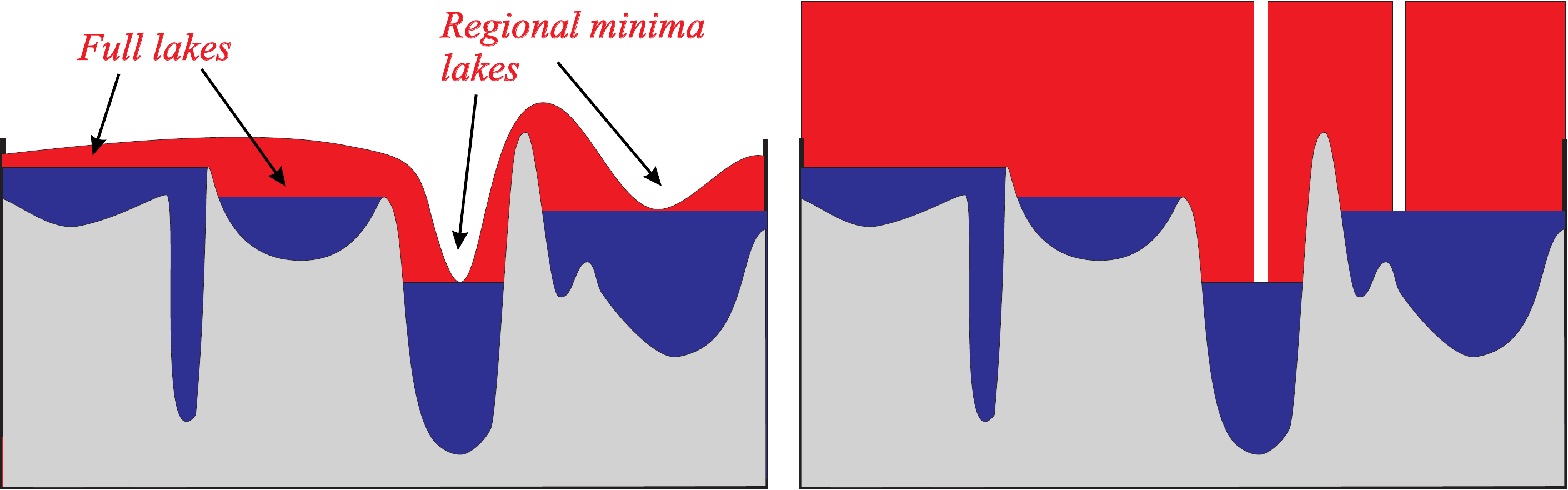}%
\caption{The highest flooding of a topographic surface below a ceiling
functino (in red). The ceiling function on the left and on the right yield the
same flooding, as they constrain the level regional minima lakes at identical
levels ; all other lakes being full lakes.}%
\label{flood6}%
\end{center}
\end{figure}
%EndExpansion
%

%TCIMACRO{\TeXButton{EndFrame}{\end{frame}}}%
%BeginExpansion
\end{frame}%
%EndExpansion
%

%TCIMACRO{\TeXButton{BeginFrame}{\begin{frame}}}%
%BeginExpansion
\begin{frame}%
%EndExpansion
%

%TCIMACRO{\QTR{frametitle}{Marker driven watershed segmentation}}%
%BeginExpansion
\frametitle{Marker driven watershed segmentation}%
%EndExpansion

The watershed of the gradient contains the contours of the image.\ The regions
to segment contain each a marker.\ A ceiling function equal to the gradient
image on the markers and equal to $\infty$ everywhere else is
constructed.\ The highest flooding under this ceiling function has regional
minima lakes at the position of the markers and full lakes everywhere
else.\ The watershed of the flooded surface gives the result.%

%TCIMACRO{\FRAME{ftbpFU}{3.3873in}{0.9348in}{0pt}{\Qcb{Left: a ceiling function
%with three minima above a topographic surface\newline Right: the highest
%flooding of the topographic surface below the ceiling function.\ It contains
%three regional minima lakes.\ The watershed partition of this function is
%indicated below, each region labeled with a distinct color and contains one
%minimum of the ceiling function.\ }}{\Qlb{flood8}}{flood8-eps-converted-to.pdf}%
%width 3.3873in;  height 0.9348in;  depth 0pt;  original-width 4.2515in;
%original-height 1.1474in;  cropleft "0";  croptop "1";  cropright "1";
%cropbottom "0";  filename 'flood8-eps-converted-to.pdf';file-properties "XNPEU";}}}%
%BeginExpansion
\begin{figure}
[ptb]
\begin{center}
\includegraphics[
height=0.9348in,
width=3.3873in
]%
{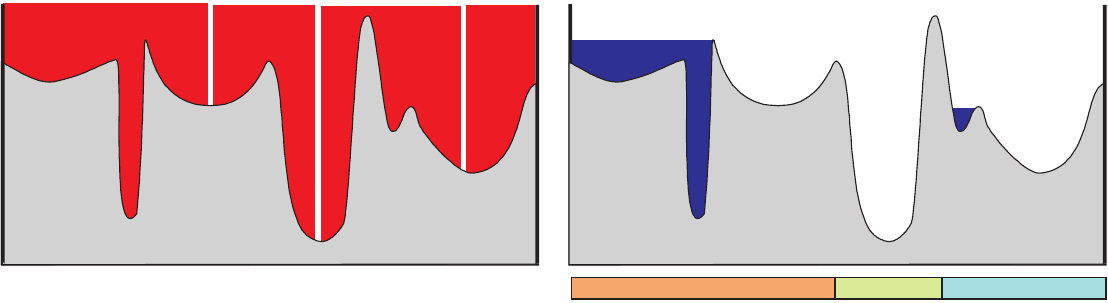}%
\caption{Left: a ceiling function with three minima above a topographic
surface\newline Right: the highest flooding of the topographic surface below
the ceiling function.\ It contains three regional minima lakes.\ The watershed
partition of this function is indicated below, each region labeled with a
distinct color and contains one minimum of the ceiling function.\ }%
\label{flood8}%
\end{center}
\end{figure}
%EndExpansion
%

%TCIMACRO{\TeXButton{EndFrame}{\end{frame}}}%
%BeginExpansion
\end{frame}%
%EndExpansion
%

%TCIMACRO{\TeXButton{BeginFrame}{\begin{frame}}}%
%BeginExpansion
\begin{frame}%
%EndExpansion

\begin{center}
{\Large \alert{The regional minimum lakes of dominated floodings}}
\end{center}

%

%TCIMACRO{\TeXButton{EndFrame}{\end{frame}}}%
%BeginExpansion
\end{frame}%
%EndExpansion
%

%TCIMACRO{\TeXButton{BeginFrame}{\begin{frame}}}%
%BeginExpansion
\begin{frame}%
%EndExpansion
%

%TCIMACRO{\QTR{frametitle}{Regional minima of dominated edge weighted graphs}}%
%BeginExpansion
\frametitle{Regional minima of dominated edge weighted graphs}%
%EndExpansion

Consider a flooding $\tau$ of a tank network and $X$, one of its lakes of
altitude $\lambda.\ $Each node $p$ of $X$ verifies $\tau_{p}=\lambda.\ $

If there exists a pair of nodes, $s\in X$ and $t\notin X,$ such that $\tau
_{t}<\lambda$ and $e_{st}=\lambda$ then $X$ is a full lake and the edge
$(s,t)$ an exhaust edge for $X.$ Such a lake cannot have a higher altitude
than its exhaust edges, and the constraining function plays no role in the
level of the lake.

On the contrary, if all edges of the cocycle of $X$ have an altitude above
$\lambda,$ then $X$ is a regional minimum of the flooding $\tau.\ $If the
level $\lambda$ of the lake cannot become higher, it is because it is
constrained by at least one node of the ceiling function $\omega$ at altitude
$\lambda.\ $This node cannot be any node, as stated now.

\begin{theorem}
Any regional minimum lake of the highest flooding of a graph $G_{e}$ with edge
weights $e,$ below a ceiling function $\omega$, contains a regional minimum
lake of the graph with edge weights $\delta_{en}\omega\vee e$ and node weights
$\omega.$
\end{theorem}

%

%TCIMACRO{\TeXButton{EndFrame}{\end{frame}}}%
%BeginExpansion
\end{frame}%
%EndExpansion
%

%TCIMACRO{\TeXButton{BeginFrame}{\begin{frame}}}%
%BeginExpansion
\begin{frame}%
%EndExpansion
%

%TCIMACRO{\QTR{frametitle}{Regional minima of dominated node weighted graphs}}%
%BeginExpansion
\frametitle{Regional minima of dominated node weighted graphs}%
%EndExpansion

\begin{theorem}
Any regional minimum lake of the highest flooding of a graph $G$ with node
weights $f,$ below a ceiling function $\omega$, contains a regional minimum
lake of the function $\omega.$
\end{theorem}

A regional minimum lake $X$ on a topographic graph has a uniform flooding
level $\lambda$ and all its neighboring nodes have a flooding level
$>\lambda.$ The level of $X$ could be higher, were it not constrained by the
ceiling function $\omega.\ $There exists a node $p\in X$ for which $\omega
_{p}=\tau_{p}=\lambda.\ $The connected component $Y$ of nodes for which
$\omega=\lambda$ contains $p$ and is included in $X,$ as for each outside
neighbor $q$ of $X,$ we have $\omega_{q}\geq\tau_{q}>\lambda.$ On $X$ we have
$\omega\geq\tau=\lambda,$ showing that $Y$ has no lower neighbor.\ Thus $Y$ is
a regional minimum of the ceiling function.\ %

%TCIMACRO{\TeXButton{EndFrame}{\end{frame}}}%
%BeginExpansion
\end{frame}%
%EndExpansion
%

%TCIMACRO{\TeXButton{BeginFrame}{\begin{frame}}}%
%BeginExpansion
\begin{frame}%
%EndExpansion
%

%TCIMACRO{\QTR{frametitle}{Algorithmic consequences}}%
%BeginExpansion
\frametitle{Algorithmic consequences}%
%EndExpansion

The highest flooding of $f$ under $\omega$ if made of lakes and of dry zones,
where the flooding equals the ground level.\ The lakes themselves are divided
between full lakes and regional minima lakes.\ The level of the full lakes is
solely determined by the altitude of the lowest pass point surounding the
lakes.\ The level of regional minima lakes is determined by the level of the
regional minima of the ceiling function. In fact, the blocking effect is the
same whatever the size of this regional minimum~; a single point is sufficient.

Replace the ceiling function $\omega$ by a function equal to $\omega$ on at
least one node of each regional minimum produces the same dominated flooding.\ %

%TCIMACRO{\TeXButton{EndFrame}{\end{frame}}}%
%BeginExpansion
\end{frame}%
%EndExpansion
%

%TCIMACRO{\TeXButton{BeginFrame}{\begin{frame}}}%
%BeginExpansion
\begin{frame}%
%EndExpansion

\begin{center}
{\Large \alert{Considering the flooding process itself}}
\end{center}

%

%TCIMACRO{\TeXButton{EndFrame}{\end{frame}}}%
%BeginExpansion
\end{frame}%
%EndExpansion
%

%TCIMACRO{\TeXButton{BeginFrame}{\begin{frame}}}%
%BeginExpansion
\begin{frame}%
%EndExpansion
%

%TCIMACRO{\QTR{frametitle}{Observing the progression of a flooding process on
%an edge weighted graph.}}%
%BeginExpansion
\frametitle{Observing the progression of a flooding process on an edge
weighted graph.}%
%EndExpansion

We place a source pouring water at a node $\Omega$ of an edge weighted graph
$G_{e}$ and flood the graph.\ We are interested by the level of the flood at
each other node of the graph when it reaches for the first time this node. If
$p$ is a node of the graph, the flood coming from $\Omega$ will reach $p$
following the easiest path: among all paths between $\Omega$ and $p$, it
follows the path for which the highest edge is the lowest. This value
constitutes precisely the ultrametric distance $d(\Omega,p)$ between $\Omega$
and $p.$%

%TCIMACRO{\TeXButton{EndFrame}{\end{frame}}}%
%BeginExpansion
\end{frame}%
%EndExpansion
%

%TCIMACRO{\TeXButton{BeginFrame}{\begin{frame}}}%
%BeginExpansion
\begin{frame}%
%EndExpansion
%

%TCIMACRO{\QTR{frametitle}{The result of a flooding process is a flooding}}%
%BeginExpansion
\frametitle{The result of a flooding process is a flooding}%
%EndExpansion

Consider the shortest path between $\Omega$ and $p$.\ The value $\tau_{p}$ is
the weight of the highest edge between $\Omega$ and $p.\ $The last node on
this shortest path is a node $q.\ $If the highest edge on the path is $(q,p),$
then $\tau_{p}=e_{pq}.$ If not $\tau_{p}=\tau_{q}\geq e_{pq}$ as the highest
edge between $\Omega$ and $p$ of the path $\pi$ lies between $\Omega$ and $q.$
In all cases we have $\tau_{p}=\tau_{q}\vee e_{pq}.$

Consider now any other neighboring node $s$ of $p.$ The path obtained by
concatenating the shortest path between $\Omega$ and $s$ and the edge $(s,p)$
is not necessarily the shortest path between $\Omega$ and $p,$ hence $\tau
_{p}\leq\tau_{s}\vee e_{ps}.\ $

\begin{theorem}
The shortest ultrametric distance of each node $p$ of an edge weighted graph
to a particular node $\Omega$ is a flooding of this graph.\ 
\end{theorem}

%

%TCIMACRO{\TeXButton{EndFrame}{\end{frame}}}%
%BeginExpansion
\end{frame}%
%EndExpansion
%

%TCIMACRO{\TeXButton{BeginFrame}{\begin{frame}}}%
%BeginExpansion
\begin{frame}%
%EndExpansion
%

%TCIMACRO{\QTR{frametitle}{The flooding process produces a dominated
%flooding}}%
%BeginExpansion
\frametitle{The flooding process produces a dominated flooding}%
%EndExpansion

Let $I\ $be the subset of nodes for which $d(\Omega,i)=\omega_{i}.\ $If the
geodesic path between $\Omega$ and a node $p$ passes through $i,$ then
$\tau_{p}=d(\Omega,p)=e_{\Omega\omega_{i}}\vee d(i,p)$ $=\omega_{i}\vee
d(i,p).\ $

For any node $q$ we have $\tau_{q}=%
%TCIMACRO{\tbigwedge \limits_{i\in I}}%
%BeginExpansion
{\textstyle\bigwedge\limits_{i\in I}}
%EndExpansion
\omega_{i}\vee d(i,q)$.\ 

This shows that $\tau$ is the highest possible flooding of $G_{e}$ on all
nodes $\omega_{i}$ and also on all other nodes.\ Suppressing the node $\Omega$
and all edges linking $\Omega$ with another node of $G_{e}$ produces a graph
$G^{\prime}$ for which $\tau$ is the highest flooding dominated by $\omega.$%

%TCIMACRO{\TeXButton{EndFrame}{\end{frame}}}%
%BeginExpansion
\end{frame}%
%EndExpansion
%

%TCIMACRO{\TeXButton{BeginFrame}{\begin{frame}}}%
%BeginExpansion
\begin{frame}%
%EndExpansion
%

%TCIMACRO{\QTR{frametitle}{Inversely, each dominated flooding is produced by a
%flooding.\ }}%
%BeginExpansion
\frametitle{Inversely, each dominated flooding is produced by a flooding.\ }%
%EndExpansion

Any dominated flooding verifies $\tau_{p}\leq%
%TCIMACRO{\tbigwedge \limits_{q\text{ neighbor of }p}}%
%BeginExpansion
{\textstyle\bigwedge\limits_{q\text{ neighbor of }p}}
%EndExpansion
\left(  \tau_{q}\vee e_{pq}\right)  $ and $\tau_{p}\leq\omega_{p}:$ $\tau
_{p}\leq\omega_{p}\wedge%
%TCIMACRO{\tbigwedge \limits_{q\text{ neighbor of }p}}%
%BeginExpansion
{\textstyle\bigwedge\limits_{q\text{ neighbor of }p}}
%EndExpansion
\left(  \tau_{q}\vee e_{pq}\right)  .$

The highest of them verifies $\tau_{p}=\omega_{p}\wedge%
%TCIMACRO{\tbigwedge \limits_{q\text{ neighbor of }p}}%
%BeginExpansion
{\textstyle\bigwedge\limits_{q\text{ neighbor of }p}}
%EndExpansion
\left(  \tau_{q}\vee e_{pq}\right)  $

Adding to the graph $G$ a dummy node $\Omega$ with a weight $\tau_{\Omega}=0$
linked by a dummy edge $(\Omega,p)$ with a weight $\omega_{p}$ produces a
graph $\widehat{G}.$ Rewritten as $\tau_{p}=(\tau_{\Omega}\vee e_{\Omega
p})\wedge%
%TCIMACRO{\tbigwedge \limits_{q\text{ neighbor of }p}}%
%BeginExpansion
{\textstyle\bigwedge\limits_{q\text{ neighbor of }p}}
%EndExpansion
\left(  \tau_{q}\vee e_{qp}\right)  $, this formula is the expression of the
algorithm of Berge for computing the shortest ultrametric distance of each
node to $\Omega$ in the augmented graph $\widehat{G}.$

The algorithm of Berge expresses that the shortest path between $\Omega$ and
$p$ is $e_{\Omega p}=\omega_{p}$ if the path is simply the edge $(\Omega,p)$
or it is equal to $\left(  \tau_{s}\vee e_{ps}\right)  $ if the path passes
through the neighbor $s$ of $p,$ and if $\left(  \tau_{q}\vee e_{qp}\right)  $
takes its smallest value for $q=s.$%

%TCIMACRO{\TeXButton{EndFrame}{\end{frame}}}%
%BeginExpansion
\end{frame}%
%EndExpansion
%

%TCIMACRO{\TeXButton{BeginFrame}{\begin{frame}}}%
%BeginExpansion
\begin{frame}%
%EndExpansion
%

%TCIMACRO{\QTR{frametitle}{Inversely, each dominated flooding is produced by a
%flooding.\ }}%
%BeginExpansion
\frametitle{Inversely, each dominated flooding is produced by a flooding.\ }%
%EndExpansion
%

%TCIMACRO{\FRAME{ftbpFU}{4.2781in}{1.8846in}{0pt}{\Qcb{Adding a dummy node
%linked to each node $x$ in $X$ by an edge weighted by the offset at $x.$}}%
%{}{trains4-eps-converted-to.pdf}{\special{ language "Scientific Word";  type "GRAPHIC";
%width 4.2781in;  height 1.8846in;  depth 0pt;  original-width 9.8837in;
%original-height 4.3188in;  cropleft "0";  croptop "1";  cropright "1";
%cropbottom "0";  filename 'trains4-eps-converted-to.pdf';file-properties "XNPEU";}}}%
%BeginExpansion
\begin{figure}
[ptb]
\begin{center}
\includegraphics[
height=1.8846in,
width=4.2781in
]%
{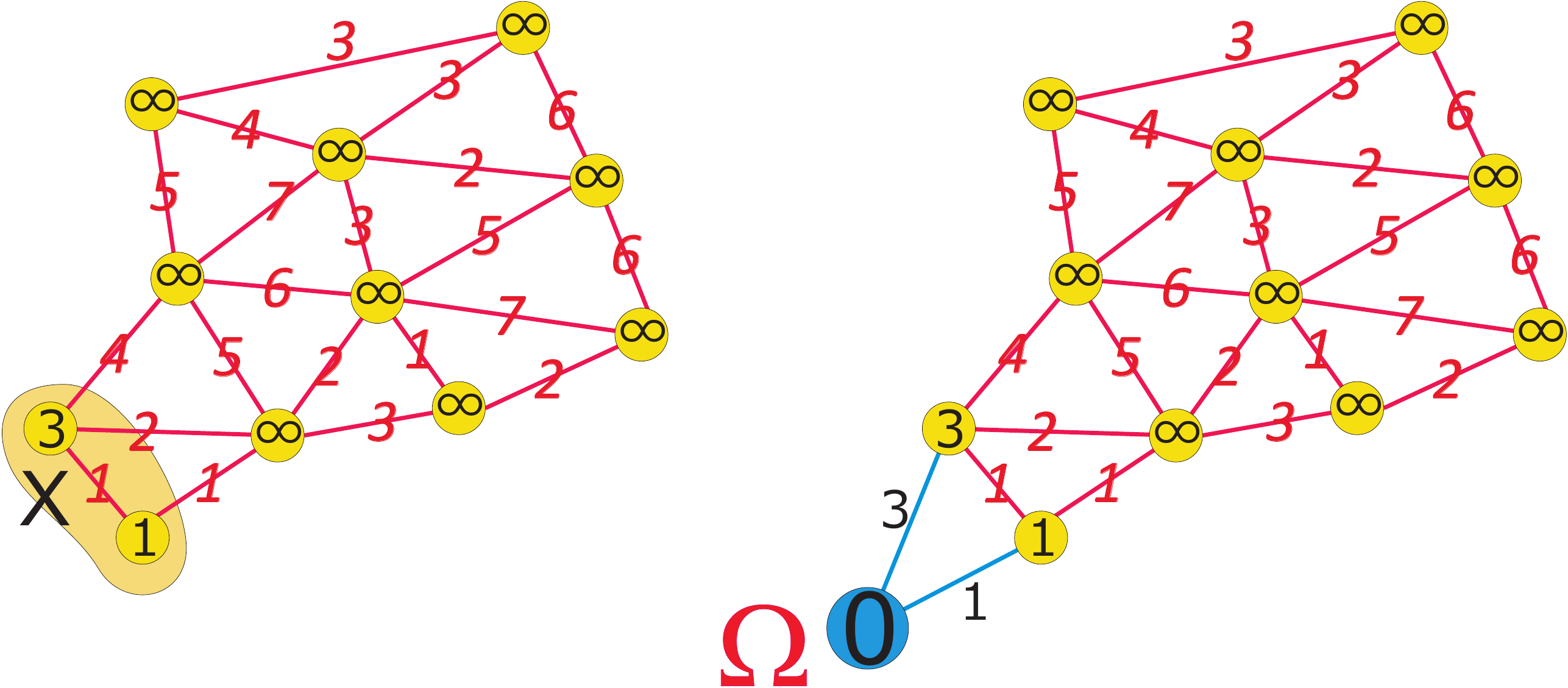}%
\caption{Adding a dummy node linked to each node $x$ in $X$ by an edge
weighted by the offset at $x.$}%
\end{center}
\end{figure}
%EndExpansion
%

%TCIMACRO{\TeXButton{EndFrame}{\end{frame}}}%
%BeginExpansion
\end{frame}%
%EndExpansion
%

%TCIMACRO{\TeXButton{BeginFrame}{\begin{frame}}}%
%BeginExpansion
\begin{frame}%
%EndExpansion
%

%TCIMACRO{\QTR{frametitle}{Inversely, each dominated flooding is produced by a
%flooding.\ }}%
%BeginExpansion
\frametitle{Inversely, each dominated flooding is produced by a flooding.\ }%
%EndExpansion

\begin{theorem}
The highest flooding of the graph $G$ below a function $\omega$ defined on the
nodes is the shortest distance of each node to $\Omega$ in augmented graph
$\widehat{G}$.\ 
\end{theorem}

If the flooding is not constrained on a node $p,$ i.e. $\omega_{p}=\infty,$
then it is not necessary to link the node $p$ with the dummy node $\Omega,$ as
the highest flooding will reach $p$ through one of its neighboring nodes.%

%TCIMACRO{\TeXButton{EndFrame}{\end{frame}}}%
%BeginExpansion
\end{frame}%
%EndExpansion
%

%TCIMACRO{\TeXButton{BeginFrame}{\begin{frame}}}%
%BeginExpansion
\begin{frame}%
%EndExpansion

\begin{center}
{\Large \alert{Pruning the graph and getting the same result}}
\end{center}

%

%TCIMACRO{\TeXButton{EndFrame}{\end{frame}}}%
%BeginExpansion
\end{frame}%
%EndExpansion
%

%TCIMACRO{\TeXButton{BeginFrame}{\begin{frame}}}%
%BeginExpansion
\begin{frame}%
%EndExpansion
%

%TCIMACRO{\QTR{frametitle}{Edge weighted graphs}}%
%BeginExpansion
\frametitle{Edge weighted graphs}%
%EndExpansion

Each dominated flooding results in a flood distribution verifying criterion
DF-2: $\tau_{q}=%
%TCIMACRO{\tbigwedge \limits_{i\in N}}%
%BeginExpansion
{\textstyle\bigwedge\limits_{i\in N}}
%EndExpansion
\omega_{i}\vee d(i,q)$.\ 

The ultrametric distance $d(i,q)$ is the weight of the highest edge in a path
$\pi$ of lowest sup-section between $i$ and $q.$ Consider now each edge
$(p,q)$ of the path $\pi.\ $If it belongs to $\pi$ we keep it. If not we may
replace it in $\pi$ by the unique path between $p$ and $q$ contained in $T,$
as all edges along this path have a weight $\leq e_{pq}$ ; these substitutions
produce paths with the same sup-section. In other words, the edges of the tree
$T$ are sufficient for computing the ultrametric distances of the graph%

%TCIMACRO{\TeXButton{EndFrame}{\end{frame}}}%
%BeginExpansion
\end{frame}%
%EndExpansion%
%TCIMACRO{\TeXButton{BeginFrame}{\begin{frame}}}%
%BeginExpansion
\begin{frame}%
%EndExpansion
%

%TCIMACRO{\QTR{frametitle}{The flooding always follows the union of minimum
%spanning trees}}%
%BeginExpansion
\frametitle{The flooding always follows the union of minimum spanning trees}%
%EndExpansion

Illustration :%

%TCIMACRO{\FRAME{ftbpF}{2.5645in}{1.9477in}{0pt}{}{}{mstfl1-eps-converted-to.pdf}%
%width 2.5645in;  height 1.9477in;  depth 0pt;  original-width 5.8979in;
%original-height 4.4632in;  cropleft "0";  croptop "1";  cropright "1";
%cropbottom "0";  filename 'mstfl1-eps-converted-to.pdf';file-properties "XNPEU";}}}%
%BeginExpansion
\begin{figure}
[ptb]
\begin{center}
\includegraphics[
height=1.9477in,
width=2.5645in
]%
{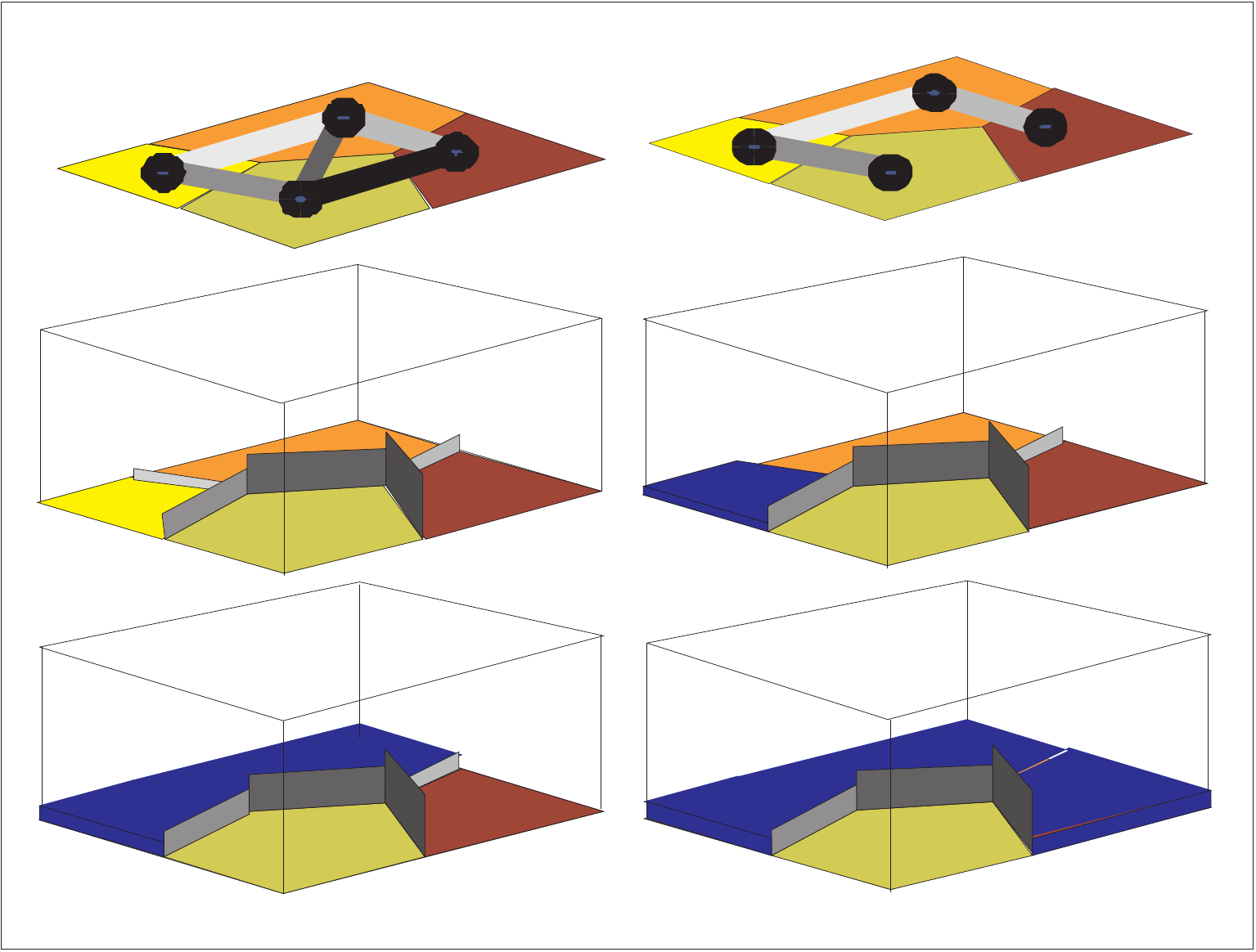}%
\end{center}
\end{figure}
%EndExpansion
%

%TCIMACRO{\TeXButton{EndFrame}{\end{frame}}}%
%BeginExpansion
\end{frame}%
%EndExpansion
%

%TCIMACRO{\TeXButton{BeginFrame}{\begin{frame}}}%
%BeginExpansion
\begin{frame}%
%EndExpansion
%

%TCIMACRO{\QTR{frametitle}{Node weighted graphs}}%
%BeginExpansion
\frametitle{Node weighted graphs}%
%EndExpansion

Any flooding $\tau\geq f$ on a node weighted graph $G_{n}$ also is a flooding
of the derived edge weighted graph $G_{e}$ with edge weights $\delta_{en}e.$
The preceding results apply.\ Any flooding $\tau\geq f$ of a MST of $G_{e}$
also is a flooding of $G_{e}$ and of $G_{n}.$%

%TCIMACRO{\TeXButton{EndFrame}{\end{frame}}}%
%BeginExpansion
\end{frame}%
%EndExpansion
%

%TCIMACRO{\TeXButton{BeginFrame}{\begin{frame}}}%
%BeginExpansion
\begin{frame}%
%EndExpansion
%

%TCIMACRO{\QTR{frametitle}{Algorithmic consequences}}%
%BeginExpansion
\frametitle{Algorithmic consequences}%
%EndExpansion

This result has interesting algorithmic implications. It is possible to
compute highest flooding using a minimum spanning tree of the graph with a
dramatically lower number of edges. However, one has to take in the balance
the time needed for constructing the graph. It may be interesting if one has
to construct several dominated flooding of the same graph.

We will meet later an algorithm which combines the construction of the MST and
of a particular flooding of the graph.\ %

%TCIMACRO{\TeXButton{EndFrame}{\end{frame}}}%
%BeginExpansion
\end{frame}%
%EndExpansion
%

%TCIMACRO{\TeXButton{BeginFrame}{\begin{frame}}}%
%BeginExpansion
\begin{frame}%
%EndExpansion

\begin{center}
{\Large \alert{Flooding with shortest distance algorithms}}

Many shortest distance algorithms exist. Each of the following has specific advantages.\ 

Algorithm of Berge

Algorithm of Dijkstra

Algorithm of Prim

Core expanding algorithm

Path algebra
\end{center}

%

%TCIMACRO{\TeXButton{EndFrame}{\end{frame}}}%
%BeginExpansion
\end{frame}%
%EndExpansion
%

%TCIMACRO{\TeXButton{BeginFrame}{\begin{frame}}}%
%BeginExpansion
\begin{frame}%
%EndExpansion

\begin{center}
{\Large \alert{Algorithms for computing the highest flooding on tank
networks}}
\end{center}

%

%TCIMACRO{\TeXButton{EndFrame}{\end{frame}}}%
%BeginExpansion
\end{frame}%
%EndExpansion
%

%TCIMACRO{\TeXButton{BeginFrame}{\begin{frame}}}%
%BeginExpansion
\begin{frame}%
%EndExpansion

\begin{center}
{\Large \alert{The algorithm of Berge}}
\end{center}

%

%TCIMACRO{\TeXButton{EndFrame}{\end{frame}}}%
%BeginExpansion
\end{frame}%
%EndExpansion
%

%TCIMACRO{\TeXButton{BeginFrame}{\begin{frame}}}%
%BeginExpansion
\begin{frame}%
%EndExpansion
%

%TCIMACRO{\QTR{frametitle}{The algorithm of Berge}}%
%BeginExpansion
\frametitle{The algorithm of Berge}%
%EndExpansion

\textbf{Initialisation}

The algorithm of Berge is initialised with the function $\omega$. For each
node $p:\tau_{p}=\omega_{p}.$

This distribution being not a flooding, the algorithm applies until stability
the relation

Repeat until $\tau_{p}^{(m)}=\tau_{p}^{(m-1)}:\tau_{p}^{(n)}=\omega_{p}\wedge%
%TCIMACRO{\tbigwedge \limits_{q\text{ neighbor of }p}}%
%BeginExpansion
{\textstyle\bigwedge\limits_{q\text{ neighbor of }p}}
%EndExpansion
\left(  \tau_{q}^{(n-1)}\vee e_{pq}\right)  $

\textbf{Convergence: }$\tau_{p}^{(m)}$ decreases at each iteration.\ It has a
lower ceiling, the smallest value of $\omega,$ therefore it converges.%

%TCIMACRO{\TeXButton{EndFrame}{\end{frame}}}%
%BeginExpansion
\end{frame}%
%EndExpansion
%

%TCIMACRO{\TeXButton{BeginFrame}{\begin{frame}}}%
%BeginExpansion
\begin{frame}%
%EndExpansion
%

%TCIMACRO{\QTR{frametitle}{The algorithm of Berge}}%
%BeginExpansion
\frametitle{The algorithm of Berge}%
%EndExpansion

\textbf{Improved version with less memory accesses:}

As $\tau$ can only decrease at each iteration, replacing the ceiling function
$\omega$ by the value taken by $\tau$ at iteration $(n-1)$ produces an
equivalent algorithm with less memory accesses: the value of $\omega$ has only
to be fetched at initialization.\ 

Initialisation: $\tau_{p}^{(0)}=\omega_{p}$

Repeat until $\tau_{p}^{(m)}=\tau_{p}^{(m-1)}:\tau_{p}^{(n)}=\tau_{p}%
^{(n-1)}\wedge%
%TCIMACRO{\tbigwedge \limits_{q\text{ neighbor of }p}}%
%BeginExpansion
{\textstyle\bigwedge\limits_{q\text{ neighbor of }p}}
%EndExpansion
\left(  \tau_{q}^{(n-1)}\vee e_{pq}\right)  $%

%TCIMACRO{\TeXButton{EndFrame}{\end{frame}}}%
%BeginExpansion
\end{frame}%
%EndExpansion
%

%TCIMACRO{\TeXButton{BeginFrame}{\begin{frame}}}%
%BeginExpansion
\begin{frame}%
%EndExpansion
%

%TCIMACRO{\QTR{frametitle}{Software or hardware implementation of the algorithm
%of Berge}}%
%BeginExpansion
\frametitle{Software or hardware implementation of the algorithm of Berge}%
%EndExpansion

Using a local neighborhood, extremely versatile, as the nodes may be processed
in any order, the algorithm of Berge t is well suited for software or hardware
implementation based on a systematic scan of the graph.\ The algorithm is
parallel or recursive:

\begin{itemize}
\item $\tau_{p}^{(n)}=\tau_{p}^{(n-1)}\wedge%
%TCIMACRO{\tbigwedge \limits_{q\text{ neighbor of }p}}%
%BeginExpansion
{\textstyle\bigwedge\limits_{q\text{ neighbor of }p}}
%EndExpansion
\left(  \tau_{q}^{(n-1)}\vee e_{pq}\right)  $ represents a parallel
implementation of the algorithm : the arguments for computing $\tau_{p}^{(n)}$
are all those obtained during the previous scan.

\item The recursive implementation separates the nodes already met during the
current scanning and the nodes in the future:\newline$\tau_{p}^{(n)}=\tau
_{p}^{(n-1)}\wedge%
%TCIMACRO{\tbigwedge \limits_{q\text{ past neighbor of }p}}%
%BeginExpansion
{\textstyle\bigwedge\limits_{q\text{ past neighbor of }p}}
%EndExpansion
\left(  \tau_{q}^{(n)}\vee e_{pq}\right)  \wedge%
%TCIMACRO{\tbigwedge \limits_{q\text{ future neighbor of }p}}%
%BeginExpansion
{\textstyle\bigwedge\limits_{q\text{ future neighbor of }p}}
%EndExpansion
\left(  \tau_{q}^{(n-1)}\vee e_{pq}\right)  \newline$Alternating a forward
scan and a backward scan for the graph permits an accelerated convergence of
the flooding levels.
\end{itemize}

%

%TCIMACRO{\TeXButton{EndFrame}{\end{frame}}}%
%BeginExpansion
\end{frame}%
%EndExpansion
%

%TCIMACRO{\TeXButton{BeginFrame}{\begin{frame}}}%
%BeginExpansion
\begin{frame}%
%EndExpansion

\begin{center}
{\Large \alert{The algorithm of Moore-Dijkstra}}
\end{center}

%

%TCIMACRO{\TeXButton{EndFrame}{\end{frame}}}%
%BeginExpansion
\end{frame}%
%EndExpansion
%

%TCIMACRO{\TeXButton{BeginFrame}{\begin{frame}}}%
%BeginExpansion
\begin{frame}%
%EndExpansion
%

%TCIMACRO{\QTR{frametitle}{The Moore Dijkstra algorithm}}%
%BeginExpansion
\frametitle{The Moore Dijkstra algorithm}%
%EndExpansion

The Dijkstra algorithm is a greedy algorithm.\ A set $S$ contains all nodes
whose distance is known. For the outside neighbors of $S,$ this distance is
estimated: for $p\in S$ and $q\notin S,$ $\tau_{q}\leq\tau_{p}\vee e_{pq.\ }$
And we have $\tau_{q}=\tau_{p}\vee e_{pq}$ if the shortest path to $q$ follows
the edge $(p,q).\ $This is the case for the node in $\overline{S}$ with the
lowest estimation. This node may be introduced into $S $ and the estimation of
the distance of its neighbors still in $\overline{S}$ updated.\ The nodes
introduced in $S$ have increasing values, as the estimation of all nodes in
$\overline{S}$ is higher than the estimation of the nodes in $S.$\bigskip

The edges linking each node with the node through which it has been flooded in
the algorithm form a tree.\ This tree is rooted at $\Omega$ and contains a
never decreasing geodesic path between $\Omega$ and each node.%

%TCIMACRO{\TeXButton{EndFrame}{\end{frame}}}%
%BeginExpansion
\end{frame}%
%EndExpansion
%

%TCIMACRO{\TeXButton{BeginFrame}{\begin{frame}}}%
%BeginExpansion
\begin{frame}%
%EndExpansion
%

%TCIMACRO{\QTR{frametitle}{The Moore Dijkstra algorithm}}%
%BeginExpansion
\frametitle{The Moore Dijkstra algorithm}%
%EndExpansion

\ \textbf{Initialisation:}

$\qquad$ $S=\{\Omega\}$ and $\tau_{\Omega}=-\infty$ ; for each node $p$ in
$N=\overline{S}:\tau_{p}=\omega_{p}$

\textbf{Flooding:}

\qquad While $\overline{S}\neq\varnothing$ repeat:

\qquad\qquad Select $j\in\overline{S}$ for which $\tau_{j}=\min_{i\in
\overline{S}}\left[  \tau_{i}\right]  $

\qquad\qquad$\qquad\overline{S}=\overline{S}\backslash\{j\}$

\qquad\qquad\qquad For any neighbor $i$ of $j$ in $\overline{S}$ do $\tau
_{i}=\min\left[  \tau_{i},\tau_{j}\vee e_{ji}\right]  $

\qquad End While%

%TCIMACRO{\TeXButton{EndFrame}{\end{frame}}}%
%BeginExpansion
\end{frame}%
%EndExpansion
%

%TCIMACRO{\TeXButton{BeginFrame}{\begin{frame}}}%
%BeginExpansion
\begin{frame}%
%EndExpansion
%

%TCIMACRO{\QTR{frametitle}{The Moore Dijkstra algorithm without the dummy node
%and edges}}%
%BeginExpansion
\frametitle{The Moore Dijkstra algorithm without the dummy node and edges}%
%EndExpansion

The dummy node $\Omega$ and the dummy edges linking $\Omega$ with the nodes of
$N$ is useless in practice:

\textbf{Initialization:}

$\qquad S=\varnothing$ ; $\overline{S}=N$ ; for each node $p$ in $N:\tau
_{p}=\omega_{p}$

\textbf{Flooding:}

While $\overline{S}\neq\varnothing$ repeat:

\qquad\qquad Select $j\in\overline{S}$ for which $\tau_{j}=\min_{i\in
\overline{S}}\left[  \tau_{i}\right]  $

\qquad\qquad$\overline{S}=\overline{S}\backslash\{j\}$

\qquad\qquad For any neighbor $i$ of $j$ in $\overline{S}$ do $\tau_{i}%
=\min\left[  \tau_{i},\tau_{j}\vee e_{ji}\right]  $

End While%

%TCIMACRO{\TeXButton{EndFrame}{\end{frame}}}%
%BeginExpansion
\end{frame}%
%EndExpansion
%

%TCIMACRO{\TeXButton{BeginFrame}{\begin{frame}}}%
%BeginExpansion
\begin{frame}%
%EndExpansion
%

%TCIMACRO{\QTR{frametitle}{Simplification of the algorithm of Dijkstra}}%
%BeginExpansion
\frametitle{Simplification of the algorithm of Dijkstra}%
%EndExpansion

When the node $j$ is introduced into $S,$ it has the highest value in $S.$ The instruction%

%TCIMACRO{\TEXTsymbol{<}}%
%BeginExpansion
$<$%
%EndExpansion
For any neighbor $i$ of $j$ in $\overline{S}$ do $\tau_{i}=\min\left[
\tau_{i},\tau_{j}\vee e_{ji}\right]  $%
%TCIMACRO{\TEXTsymbol{>}}%
%BeginExpansion
$>$%
%EndExpansion

may be simplified in:%

%TCIMACRO{\TEXTsymbol{<}}%
%BeginExpansion
$<$%
%EndExpansion
For any neighbor $i$ of $j$ verifying $\tau_{j}\vee e_{ji}<\tau_{i}$ do
$\tau_{i}=\tau_{j}\vee e_{ji}$%
%TCIMACRO{\TEXTsymbol{>}}%
%BeginExpansion
$>$%
%EndExpansion

as a node $i$ verifying $\tau_{i}>\tau_{j}$ cannot belong to $S.\ $Checking
whether $i$ belongs to $\overline{S}$ is not necessary, leading to the
following algorithm.%

%TCIMACRO{\TeXButton{EndFrame}{\end{frame}}}%
%BeginExpansion
\end{frame}%
%EndExpansion
%

%TCIMACRO{\TeXButton{BeginFrame}{\begin{frame}}}%
%BeginExpansion
\begin{frame}%
%EndExpansion
%

%TCIMACRO{\QTR{frametitle}{Simplification of the algorithm of Dijkstra}}%
%BeginExpansion
\frametitle{Simplification of the algorithm of Dijkstra}%
%EndExpansion

Initialization:

$\qquad S=\varnothing$ ; $\overline{S}=N$ ; for each node $p$ in $N:\tau
_{p}=\omega_{p}$

Flooding:

While $\overline{S}\neq\varnothing$ repeat:

\qquad\qquad Select $j\in\overline{S}$ for which $\tau_{j}=\min_{i\in
\overline{S}}\left[  \tau_{i}\right]  $

\qquad\qquad$\overline{S}=\overline{S}\backslash\{j\}$

\qquad\qquad For any neighbor $i$ of $j$ verifying $\tau_{j}\vee e_{ji}%
<\tau_{i}$ do $\tau_{i}=\tau_{j}\vee e_{ji}$

End While%

%TCIMACRO{\TeXButton{EndFrame}{\end{frame}}}%
%BeginExpansion
\end{frame}%
%EndExpansion%
%TCIMACRO{\TeXButton{BeginFrame}{\begin{frame}}}%
%BeginExpansion
\begin{frame}%
%EndExpansion
%

%TCIMACRO{\QTR{frametitle}{The algorithm of Dijkstra with a funnel structure}}%
%BeginExpansion
\frametitle{The algorithm of Dijkstra with a funnel structure}%
%EndExpansion

Fetching the node with the smallest value $\tau_{j}$ within $\overline{S}$ is
made easy by using an adequate data structure such as a "funnel" structure:
nodes with any order of priority may be stored in the funnel ; but only one of
the nodes with the smallest priority is extracted at any time.

Possible implementation: an ordered bucket structure. A node is introduced in
the bucket corresponding to its priority.\ Each extracted node is chosen among
the nodes in the bucket with highest priority.

If the buckets have the structure of a queue, we speak about hierarchical queues.%

%TCIMACRO{\TeXButton{EndFrame}{\end{frame}}}%
%BeginExpansion
\end{frame}%
%EndExpansion
%

%TCIMACRO{\TeXButton{BeginFrame}{\begin{frame}}}%
%BeginExpansion
\begin{frame}%
%EndExpansion
%

%TCIMACRO{\QTR{frametitle}{The algorithm of Dijkstra with a funnel structure}}%
%BeginExpansion
\frametitle{The algorithm of Dijkstra with a funnel structure}%
%EndExpansion

Before being introduced into $S,$ the distance of each node has to be
estimated anew every time one of its neighbor is introduced into $S.\ $If we
use a "singel occupancy funnel", a node occupies only one location and its
priority is updated if needed. In a multiple occupancy funnel (MOF), a node
may occupy more than one location, with distinct priority.\ When a node is
extracted for the first time, its value is correct.\ When it is extracted
another time one has to discard it.\ %

%TCIMACRO{\TeXButton{EndFrame}{\end{frame}}}%
%BeginExpansion
\end{frame}%
%EndExpansion
%

%TCIMACRO{\TeXButton{BeginFrame}{\begin{frame}}}%
%BeginExpansion
\begin{frame}%
%EndExpansion
%

%TCIMACRO{\QTR{frametitle}{The algorithm of Dijkstra with a funnel structure}}%
%BeginExpansion
\frametitle{The algorithm of Dijkstra with a funnel structure}%
%EndExpansion

Consider a flooding of the following graph.\ \ Initially the node $p$ is in
$\Phi$ with a value $3$ and a yellow colour.\ The two other nodes are not in
$\Phi.\ \ $When $p$ is extracted from $\Phi,$ both its neighbors are
introduced in $\Phi$ with estimates equal to $5$ and $9$ (yellow colour).\ The
node $q$ is then extracted from $\Phi$ and its neighbor $r$ is introduced in
$\Phi$ with a new estimate $7.;\ $at the same time, the node $r$ of the graph
gets this same value.\ When $r$ is extracted from $\Phi$ for the first time,
there is an identity between its priority and the flooding level of $r$ in the
graph. The second time it is extracted, its priority is $9,$ higher than the
value in the graph and has to be discarded from further processing.\ %

%TCIMACRO{\FRAME{ftbpFU}{3.5924in}{0.7109in}{0pt}{\Qcb{Propagation of the
%flooding on an edge weighted graph using the Dijkstra algorithm.\ }%
%}{\Qlb{inond5}}{inond5-eps-converted-to.pdf}{\special{ language "Scientific Word";
%type "GRAPHIC";  maintain-aspect-ratio TRUE;  display "USEDEF";
%valid_file "F";  width 3.5924in;  height 0.7109in;  depth 0pt;
%original-width 3.5838in;  original-height 0.6867in;  cropleft "0";
%filename 'inond5-eps-converted-to.pdf';file-properties "XNPEU";}}}%
%BeginExpansion
\begin{figure}
[ptb]
\begin{center}
\includegraphics[
height=0.7109in,
width=3.5924in
]%
{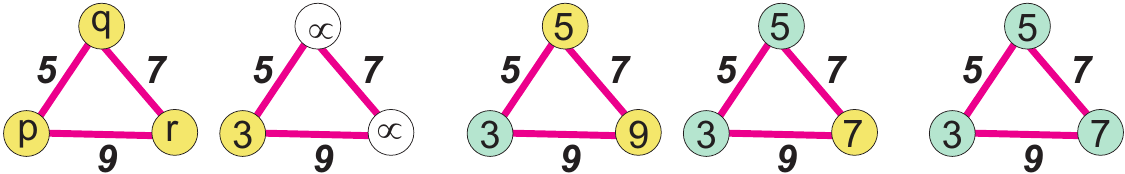}%
\caption{Propagation of the flooding on an edge weighted graph using the
Dijkstra algorithm.\ }%
\label{inond5}%
\end{center}
\end{figure}
%EndExpansion
%

%TCIMACRO{\TeXButton{EndFrame}{\end{frame}}}%
%BeginExpansion
\end{frame}%
%EndExpansion
%

%TCIMACRO{\TeXButton{BeginFrame}{\begin{frame}}}%
%BeginExpansion
\begin{frame}%
%EndExpansion
%

%TCIMACRO{\QTR{frametitle}{The algorithm of Dijkstra with a funnel structure}}%
%BeginExpansion
\frametitle{The algorithm of Dijkstra with a funnel structure}%
%EndExpansion

\textbf{Initialization:}

\qquad create $\Phi$, a multi occupancy funnel ; $\tau=\infty$

\qquad for each node $p$ verifying $\omega_{p}<\infty$

\qquad\qquad introduce $p$ into $\Phi$ with a priority $\omega_{p}$ ;
$\tau_{p}=\omega_{p}$

\textbf{Flooding}

\qquad While $\Phi$ is not empty repeat:

\qquad\qquad Extract from $\Phi$ the node $j$ with the lowest prioriy
$\lambda$

\qquad\qquad If $\tau_{j}=$ $\lambda$

\qquad\qquad\qquad For any neighbor $i$ of $j$ such that $\tau_{j}\vee
e_{ji}<\tau_{i}$

\qquad\qquad\qquad\qquad$\tau_{i}=\tau_{j}\vee e_{ji}$

\qquad\qquad\qquad\qquad introduce $i$ into $\Phi$ with the priority $\tau
_{i}$

End While%

%TCIMACRO{\TeXButton{EndFrame}{\end{frame}}}%
%BeginExpansion
\end{frame}%
%EndExpansion%
%TCIMACRO{\TeXButton{BeginFrame}{\begin{frame}}}%
%BeginExpansion
\begin{frame}%
%EndExpansion
%

%TCIMACRO{\QTR{frametitle}{Marker based segmentation}}%
%BeginExpansion
\frametitle{Marker based segmentation}%
%EndExpansion

In the case of marker based segmentation a number of nodes are markers.\ The
aim is to produce a Voronoi tessellation of the graph: each node is assigned
to the marker which is closest for the ultrametric distance.\ Several
solutions exist for breaking the ties, if a node is at the same distance of
distinct markers.

The preceding algorithm may be used, the reduced set $\widehat{S}$ of ceiling
minima, being the set of nodes belonging to the markers. In the context of
segmentation, a distinct label is assigned to each marker.\ This label will be
assigned to the total region flooded through this marker.%

%TCIMACRO{\TeXButton{EndFrame}{\end{frame}}}%
%BeginExpansion
\end{frame}%
%EndExpansion
%

%TCIMACRO{\TeXButton{BeginFrame}{\begin{frame}}}%
%BeginExpansion
\begin{frame}%
%EndExpansion
%

%TCIMACRO{\QTR{frametitle}{Marker based segmentation}}%
%BeginExpansion
\frametitle{Marker based segmentation}%
%EndExpansion

\textbf{Initialization:}

\qquad create a HQ $\Phi$

\qquad create an image $\zeta$ which will hold the labels

\qquad assign to each marker node $p$ a distinct label $\zeta_{p},$ a flooding
value $\tau_{p}=0$ and introduce $p$ into $\Phi$ with a priority $0$

\qquad for all other nodes $\tau=\infty$

\textbf{Flooding}

\qquad While $\Phi$ is not empty repeat:

\qquad\qquad Extract from $\Phi$ the node $j$ with the lowest prioriy
$\lambda$

\qquad\qquad If $\tau_{i}=$ $\lambda$

\qquad\qquad\qquad For any neighbor $i$ of $j$ such that $\tau_{j}\vee
e_{ji}<\tau_{i}$

\qquad\qquad\qquad\qquad$\tau_{i}=\tau_{j}\vee e_{ji}$

\qquad\qquad\qquad\qquad$\zeta_{i}=\zeta_{j}$

\qquad\qquad\qquad\qquad introduce $i$ into $\Phi$ with the priority $\tau
_{i}$

End While%

%TCIMACRO{\TeXButton{EndFrame}{\end{frame}}}%
%BeginExpansion
\end{frame}%
%EndExpansion
%

%TCIMACRO{\TeXButton{BeginFrame}{\begin{frame}}}%
%BeginExpansion
\begin{frame}%
%EndExpansion
%

%TCIMACRO{\QTR{frametitle}{Advantages of a hierarchical queue}}%
%BeginExpansion
\frametitle{Advantages of a hierarchical queue}%
%EndExpansion

A hierarchical queue are governed by a double hierarchical order. A
hierarchical queue is a series of first first out queues, having each a
priority.\ A node is introduced in the queue corresponding to its
priority.\ The node which is extracted at any time of the HQ is the node which
has been introduced first in the queue with the highest priority.\ The
priorities among the queues organize that the flooding progresses in an order
of increasing altitudes. The FIFO\ structure of the queue ensures that a node
inside a plateau is flooded in an order proportional to its distance to the
lower border of the plateau.%

%TCIMACRO{\TeXButton{EndFrame}{\end{frame}}}%
%BeginExpansion
\end{frame}%
%EndExpansion
%

%TCIMACRO{\TeXButton{BeginFrame}{\begin{frame}}}%
%BeginExpansion
\begin{frame}%
%EndExpansion

\begin{center}
{\Large \alert{The algorithm of Prim}}
\end{center}

%

%TCIMACRO{\TeXButton{EndFrame}{\end{frame}}}%
%BeginExpansion
\end{frame}%
%EndExpansion
%

%TCIMACRO{\TeXButton{BeginFrame}{\begin{frame}}}%
%BeginExpansion
\begin{frame}%
%EndExpansion
%

%TCIMACRO{\QTR{frametitle}{Constructing the MST}}%
%BeginExpansion
\frametitle{Constructing the MST}%
%EndExpansion

On a graph, from a node $p$ to a node $q,$ the flood always follows the paths
of lowest sup-section linking $p$ and $q.$ All such paths belong to the MST of
the graph. Hence, it is possible to combine the flooding with the construction
of the MST. The algorithm of PRIM\ constructs the MST rooted in $\Omega.$

\textbf{Initialisation}

Initially, the tree $T$ spans only the node $\Omega$.

\textbf{Expansion}

As long as the tree does not contain all nodes of the graph:

\qquad Chose the lowest edge $(q,s)$ in the cocycle of $T,$ such that $q\in T
$ and $s\notin T.$ Append the node $s$ to the tree: $T=T\cup\{s\}$%

%TCIMACRO{\TeXButton{EndFrame}{\end{frame}}}%
%BeginExpansion
\end{frame}%
%EndExpansion
%

%TCIMACRO{\TeXButton{BeginFrame}{\begin{frame}}}%
%BeginExpansion
\begin{frame}%
%EndExpansion
%

%TCIMACRO{\QTR{frametitle}{Flooding an edge weighted graph following its MST}}%
%BeginExpansion
\frametitle{Flooding an edge weighted graph following its MST}%
%EndExpansion

Flooding with the algorithm of PRIM

The result of the preceding algorithm is a tree rooted at $\Omega.\ $Each
other node $p$ is linked with $\Omega$ through a unique path.\ The flood
coming from $\Omega$ necessarily follows this path.\ The flooding of the nodes
and the construction of the tree may be done simultaneously.

\textbf{Initialisation}

Initially, the tree $T$ spans only the node $\Omega$ : $T=\{\Omega
\}.\ \tau_{\Omega}=0.$

\textbf{Expansion}

As long the tree does not contain all nodes of the graph:

\qquad Chose the lowest edge $(q,s)$ in the cocycle of $T,$ such that $q\in T
$ and $s\notin T.$

\qquad Append the edge $(q,s)$ and the node $s$ to the tree: $T=T\cup\{s\}$

\qquad$\tau_{s}=\tau_{q}\vee e_{qs}$%

%TCIMACRO{\TeXButton{EndFrame}{\end{frame}}}%
%BeginExpansion
\end{frame}%
%EndExpansion
%

%TCIMACRO{\TeXButton{BeginFrame}{\begin{frame}}}%
%BeginExpansion
\begin{frame}%
%EndExpansion
%

%TCIMACRO{\QTR{frametitle}{Flooding an edge weighted graph following its MST}}%
%BeginExpansion
\frametitle{Flooding an edge weighted graph following its MST}%
%EndExpansion

\textbf{Analysis of the algorithm}

The nodes are introduced with a never decreasing flood level. A\ node with a
flood level $\lambda$ first floods its neighbors appended through an edge
which is lower or equal to the current flooding level: these neighbors get the
current flood level $\lambda$ and are appended to the tree.\ If this is not
possible anymode, the smallest edge in the cocycle of the tree with a weight
$>\lambda$ is followed, introducing the first node with a weight $>\lambda$
into the tree.

The PRIM\ algorithm is a particular avatar of Dijkstra's algorithm.\ Among all
neighboring nodes of $T$ for which the estimated flooding level is the
smallest, the algorithm of PRIM first considers those linked with the tree
through the lowest edge.\ %

%TCIMACRO{\TeXButton{EndFrame}{\end{frame}}}%
%BeginExpansion
\end{frame}%
%EndExpansion
%

%TCIMACRO{\TeXButton{BeginFrame}{\begin{frame}}}%
%BeginExpansion
\begin{frame}%
%EndExpansion
%

%TCIMACRO{\QTR{frametitle}{Scheduling with a HQ}}%
%BeginExpansion
\frametitle{Scheduling with a HQ}%
%EndExpansion

\textbf{Initialization:}

\qquad create $\Phi$, a multi occupancy funnel.\ 

\qquad$\tau=\infty$

\qquad for each node $p$ verifying $\omega_{p}<\infty$, introduce $p$ into
$\Phi$ with a priority $\omega_{p}$

$\qquad\lambda=-\infty$

\textbf{Flooding}

\qquad While $\Phi$ is not empty repeat:

\qquad\qquad Extract from $\Phi$ the node $p$ with the lowest prioriy $\mu$

\qquad\qquad if $\tau_{p}=\infty$

\qquad\qquad\qquad If $\mu>$ $\lambda:\lambda=\mu$

\qquad\qquad\qquad$\tau_{p}=\lambda$

\qquad\qquad\qquad For any neighbor $q$ of $p$ such that $\tau_{q}=\infty$

\qquad\qquad\qquad\qquad introduce $q$ into $\Phi$ with the priority $e_{pq}$

\textbf{Remark: }Replacing the last instruction with
%TCIMACRO{\TEXTsymbol{<}}%
%BeginExpansion
$<$%
%EndExpansion
introduce $i$ into the funnel with the priority $\lambda\vee e_{ji}$%
%TCIMACRO{\TEXTsymbol{>} }%
%BeginExpansion
$>$
%EndExpansion
produces the algorithm of Dijkstra.%

%TCIMACRO{\TeXButton{EndFrame}{\end{frame}}}%
%BeginExpansion
\end{frame}%
%EndExpansion
%

%TCIMACRO{\TeXButton{BeginFrame}{\begin{frame}}}%
%BeginExpansion
\begin{frame}%
%EndExpansion
%

%TCIMACRO{\QTR{frametitle}{Marker based segmentation with the algorithm of
%Prim}}%
%BeginExpansion
\frametitle{Marker based segmentation with the algorithm of Prim}%
%EndExpansion

\textbf{Initialisation}

$T=\{\varnothing\}$

For each marker $p:$

$\qquad\tau_{p}=0$

\qquad$T=T\cup\{p\}$

\qquad assign a new label $\zeta_{p}$

\textbf{Expansion}

As long the tree does not contain all nodes of the graph:

\qquad Chose the lowest edge $(q,s)$ in the cocycle of $T,$ such that $q\in T
$ and $s\notin T.$

\qquad Append the edge $(q,s)$ and the node $s$ to the tree: $T=T\cup\{s\}$

\qquad$\tau_{s}=\tau_{q}\vee e_{qs}$

\qquad$\zeta_{s}=\zeta_{q}$%

%TCIMACRO{\TeXButton{EndFrame}{\end{frame}}}%
%BeginExpansion
\end{frame}%
%EndExpansion
%

%TCIMACRO{\TeXButton{BeginFrame}{\begin{frame}}}%
%BeginExpansion
\begin{frame}%
%EndExpansion
%

%TCIMACRO{\QTR{frametitle}{Marker based segmentation with the algorithm of
%Prim}}%
%BeginExpansion
\frametitle{Marker based segmentation with the algorithm of Prim}%
%EndExpansion

If we are not interested by the flooding level but only by the Voronoi
partition associated to the markers:

\textbf{Initialisation}

For each marker $p,$ assign a new label $\zeta_{p}$

For all other nodes $q:\zeta_{q}=-\infty$

\textbf{Expansion}

As long as there are nodes with a label $\zeta=-\infty$

\qquad Chose the lowest edge $(q,s)$ verifying $\zeta_{q}>-\infty$ and
$\zeta_{s}=-\infty$

\qquad$\zeta_{s}=\zeta_{q}$%

%TCIMACRO{\TeXButton{EndFrame}{\end{frame}}}%
%BeginExpansion
\end{frame}%
%EndExpansion
%

%TCIMACRO{\TeXButton{BeginFrame}{\begin{frame}}}%
%BeginExpansion
\begin{frame}%
%EndExpansion

\begin{center}
{\Large \alert{Shortest path algorithms on node weighted graphs}}
\end{center}

%

%TCIMACRO{\TeXButton{EndFrame}{\end{frame}}}%
%BeginExpansion
\end{frame}%
%EndExpansion
%

%TCIMACRO{\TeXButton{BeginFrame}{\begin{frame}}}%
%BeginExpansion
\begin{frame}%
%EndExpansion
%

%TCIMACRO{\QTR{frametitle}{From edge weighted floodings to node weighted
%floodings}}%
%BeginExpansion
\frametitle{From edge weighted floodings to node weighted floodings}%
%EndExpansion

Any flooding $\tau$ of a node weighted graph $G_{n}$, above the ground level
$f$ also is a flooding on an edge weighted graph $G_{e}$ with edge weights
$e_{pq}=f_{p}\vee f_{q}$: -%
%TCIMACRO{\TEXTsymbol{>} }%
%BeginExpansion
$>$
%EndExpansion
all results and algorithms established for $G_{e}$ are applicable to r $G_{n}$
simply by replacing $e_{pq}$ by $f_{p}\vee f_{q}$ and remembering that
$\tau\geq f,$ they get simpler.%

%TCIMACRO{\TeXButton{EndFrame}{\end{frame}}}%
%BeginExpansion
\end{frame}%
%EndExpansion
%

%TCIMACRO{\TeXButton{BeginFrame}{\begin{frame}}}%
%BeginExpansion
\begin{frame}%
%EndExpansion

\begin{center}
{\Large \alert{The algorithm of Berge}}
\end{center}

%

%TCIMACRO{\TeXButton{EndFrame}{\end{frame}}}%
%BeginExpansion
\end{frame}%
%EndExpansion
%

%TCIMACRO{\TeXButton{BeginFrame}{\begin{frame}}}%
%BeginExpansion
\begin{frame}%
%EndExpansion
%

%TCIMACRO{\QTR{frametitle}{The algorithm of Berge}}%
%BeginExpansion
\frametitle{The algorithm of Berge}%
%EndExpansion

Initialisation: $\tau_{p}^{(0)}=\omega_{p}$

Repeat until $\tau_{p}^{(m)}=\tau_{p}^{(m-1)}:\tau_{p}^{(n)}=\omega_{p}\wedge%
%TCIMACRO{\tbigwedge \limits_{q\text{ neighbor of }p}}%
%BeginExpansion
{\textstyle\bigwedge\limits_{q\text{ neighbor of }p}}
%EndExpansion
\left(  \tau_{q}^{(n-1)}\vee f_{p}\vee f_{q}\right)  =\omega_{p}\wedge%
%TCIMACRO{\tbigwedge \limits_{q\text{ neighbor of }p}}%
%BeginExpansion
{\textstyle\bigwedge\limits_{q\text{ neighbor of }p}}
%EndExpansion
\left(  \tau_{q}^{(n-1)}\vee f_{p}\right)  $

\textbf{A variant of the algorithm}

Replacing the ceiling function $\omega$ by the value taken by $\tau$ at
iteration $(n-1)$ since $\tau$ can only decrease at each iteration. The
algorithm becomes:

Initialisation: $\tau_{p}^{(0)}=\omega_{p}$

Repeat until $\tau_{p}^{(m)}=\tau_{p}^{(m-1)}:\tau_{p}^{(n)}=\tau_{p}%
^{(n-1)}\wedge%
%TCIMACRO{\tbigwedge \limits_{q\text{ neighbor of }p}}%
%BeginExpansion
{\textstyle\bigwedge\limits_{q\text{ neighbor of }p}}
%EndExpansion
\left(  \tau_{q}^{(n-1)}\vee f_{p}\right)  $

We recognize the classical algorithm. Repeat until stability $\tau
=f\vee\varepsilon\tau$%

%TCIMACRO{\TeXButton{EndFrame}{\end{frame}}}%
%BeginExpansion
\end{frame}%
%EndExpansion
%

%TCIMACRO{\TeXButton{BeginFrame}{\begin{frame}}}%
%BeginExpansion
\begin{frame}%
%EndExpansion

\begin{center}
{\Large \alert{The algorithm of Prim}}
\end{center}

%

%TCIMACRO{\TeXButton{EndFrame}{\end{frame}}}%
%BeginExpansion
\end{frame}%
%EndExpansion
%

%TCIMACRO{\TeXButton{BeginFrame}{\begin{frame}}}%
%BeginExpansion
\begin{frame}%
%EndExpansion
%

%TCIMACRO{\QTR{frametitle}{The algorithm of Prim}}%
%BeginExpansion
\frametitle{The algorithm of Prim}%
%EndExpansion

The algorithm of PRIM\ remains exactly the same as for pipe networks.\ By
replacing $e_{qs}$ by its value $f_{q}\vee f_{s},$ and since $\tau_{q}\geq
f_{q}$, we get $\tau_{q}\vee f_{q}\vee f_{s}=\tau_{q}\vee f_{s}$.\ The
flooding of the nodes and the construction of the tree may be done simultaneously.

\textbf{Initialisation}

Initially, the tree $T$ has only the node $\Omega$ and no edge: $T=\{\Omega
,\varnothing\}.\ \tau_{\Omega}=0.$

\textbf{Expansion}

As long the tree does not contain all nodes of the graph:

\qquad Chose the edge $(q,s)$ with the lowest weight $f_{q}\vee f_{s}$ in the
cocycle of $T,$ such that $q\in T$ and $s\notin T.$

\qquad Assign the node $s$ to the tree: $T=T\cup\{s\}$

\qquad$\tau_{s}=\tau_{q}\vee f_{q}\vee f_{s}=\tau_{q}\vee f_{s}$%

%TCIMACRO{\TeXButton{EndFrame}{\end{frame}}}%
%BeginExpansion
\end{frame}%
%EndExpansion
%

%TCIMACRO{\TeXButton{BeginFrame}{\begin{frame}}}%
%BeginExpansion
\begin{frame}%
%EndExpansion

\begin{center}
{\Large \alert{The algorithm of Dijkstra}}
\end{center}

%

%TCIMACRO{\TeXButton{EndFrame}{\end{frame}}}%
%BeginExpansion
\end{frame}%
%EndExpansion
%

%TCIMACRO{\TeXButton{BeginFrame}{\begin{frame}}}%
%BeginExpansion
\begin{frame}%
%EndExpansion
%

%TCIMACRO{\QTR{frametitle}{The algorithm of Dijkstra}}%
%BeginExpansion
\frametitle{The algorithm of Dijkstra}%
%EndExpansion

Initialization:

$\qquad S=\varnothing$ ; for each node $p$ in $N:\tau_{p}=\omega_{p}$

Flooding:

While $S\neq N$ repeat:

\qquad\qquad Select $j\in\overline{S}$ for which $\tau_{j}=\min_{i\in
\overline{S}}\left[  \tau_{i}\right]  $

\qquad\qquad$S=S\cup\{j\}$

\qquad\qquad For any neighbor $i$ of $j$ verifying $\tau_{i}>\tau_{j}\vee
f_{i}$ do $\tau_{i}=\tau_{j}\vee f_{i}$

End While%

%TCIMACRO{\TeXButton{EndFrame}{\end{frame}}}%
%BeginExpansion
\end{frame}%
%EndExpansion
%

%TCIMACRO{\TeXButton{BeginFrame}{\begin{frame}}}%
%BeginExpansion
\begin{frame}%
%EndExpansion
%

%TCIMACRO{\QTR{frametitle}{Avoiding unnecessary work}}%
%BeginExpansion
\frametitle{Avoiding unnecessary work}%
%EndExpansion

$j$ is the smallest neighbor of $i.\ $If $i$ is flooded through one of its
neighbors, this neighbor can only be $j$ and as soon the value $\tau_{i}$ is
computed once, this value is correct and final : $\tau_{i}=\tau_{j}\vee f_{i}%
$. For this reason, we may use an image of binary flags $\zeta$, in order to
flag all nodes for which the flooding value is known

Initialization:

$\qquad S=\varnothing$ ; for each node $p$ in $N:\tau_{p}=\omega_{p}$ and
$\zeta_{p}=0$

Flooding:

While $S\neq N$ repeat:

\qquad\qquad Select $j\in\overline{S}$ for which $\tau_{j}=\min_{i\in
\overline{S}}\left[  \tau_{i}\right]  $

\qquad\qquad$S=S\cup\{j\}$

\qquad\qquad For any neighbor $i$ of $j$ such that $\zeta_{i}=0$ and $\tau
_{i}>\tau_{j}\vee f_{i}$

\qquad\qquad\qquad$\zeta_{i}=1$

\qquad\qquad\qquad$\tau_{i}=\tau_{j}\vee f_{i}$

End While%

%TCIMACRO{\TeXButton{EndFrame}{\end{frame}}}%
%BeginExpansion
\end{frame}%
%EndExpansion
%

%TCIMACRO{\TeXButton{BeginFrame}{\begin{frame}}}%
%BeginExpansion
\begin{frame}%
%EndExpansion

\begin{center}
{\Large \alert{The regional minima of the ceiling function are sufficient,
and any overset of them...}}
\end{center}

%

%TCIMACRO{\TeXButton{EndFrame}{\end{frame}}}%
%BeginExpansion
\end{frame}%
%EndExpansion%
%TCIMACRO{\TeXButton{BeginFrame}{\begin{frame}}}%
%BeginExpansion
\begin{frame}%
%EndExpansion
%

%TCIMACRO{\QTR{frametitle}{Finding an overset of the regional minima nodes of
%the ceiling function}}%
%BeginExpansion
\frametitle{Finding an overset of the regional minima nodes of the ceiling
function}%
%EndExpansion

Each regional minimum of a flooding contains a regional minimum of the ceiling
function $\omega$.\ If it were not constrained at this level, the level of the
flooding would be higher in this regional minimum.\ Outside the regional
minima, the level of the flooding not constrained by $\omega.$

Ideally, the algorithm of Dijkstra should be initialized using one node and
only one node in each regional minimum of $\omega$.\ As the regional minima
are costly to compute, a cheap overset of these nodes offers a better compromise.%

%TCIMACRO{\TeXButton{EndFrame}{\end{frame}}}%
%BeginExpansion
\end{frame}%
%EndExpansion
%

%TCIMACRO{\TeXButton{BeginFrame}{\begin{frame}}}%
%BeginExpansion
\begin{frame}%
%EndExpansion
%

%TCIMACRO{\QTR{frametitle}{A first overset of the ceiling minima}}%
%BeginExpansion
\frametitle{A first overset of the ceiling minima}%
%EndExpansion

The regional minima of $\omega$ are plateaus of uniform altitude, without
lower neighbors.\ During a forward raster scan of the image, the pixels are
detected which have only higher neighbors in the past and no lower neighbor in
the future.\ The set $X$ is obained in a forward scan through the image (for a
2D image, from left to rights and from top to bottom).

$X=\left\{  p\mid\tau_{p}<%
%TCIMACRO{\tbigwedge \limits_{q\in\operatorname*{past}(p)}}%
%BeginExpansion
{\textstyle\bigwedge\limits_{q\in\operatorname*{past}(p)}}
%EndExpansion
\tau_{q}\right\}  \wedge\left\{  p\mid\tau_{p}\leq%
%TCIMACRO{\tbigwedge \limits_{q\in\operatorname*{future}(p)}}%
%BeginExpansion
{\textstyle\bigwedge\limits_{q\in\operatorname*{future}(p)}}
%EndExpansion
\tau_{q}\right\}  $

This algorithm finds the entry points in the regional minima and in a certain
number of plateaus.\ We may reduce the plateaus as follows.%

%TCIMACRO{\TeXButton{EndFrame}{\end{frame}}}%
%BeginExpansion
\end{frame}%
%EndExpansion
%

%TCIMACRO{\TeXButton{BeginFrame}{\begin{frame}}}%
%BeginExpansion
\begin{frame}%
%EndExpansion
%

%TCIMACRO{\QTR{frametitle}{Reducing the number of candidates}}%
%BeginExpansion
\frametitle{Reducing the number of candidates}%
%EndExpansion

A classical algorithm for constructing regional minima.\ The flooding of
$\omega$ with $\omega+1$ as ceiling function produces a new function
$\widehat{\omega}.\ $The regional minima of $\omega$ are all nodes verifying
$\widehat{\omega}.$ For the sake of economy only a partial flooding is done,
suppressing a number of plateaus which are not regional minima:

\begin{itemize}
\item using geodesic erosions defined as:\newline$\varepsilon_{\omega
}(g)=\varepsilon g\vee\omega$\newline$\varepsilon_{\omega}^{(n)}%
(g)=\varepsilon_{\omega}(\varepsilon_{\omega}^{(n)}(g))$\newline The set of
nodes $Y$ verifying $\varepsilon_{\omega}^{(n)}(g)>\omega$ is an overset of
the regional minima, decreasing with the number of iterations $n.$

\item Using one pass of a recursive geodesic erosion of $\omega$ above $g$
with a backward scanning order\newline$\overleftarrow{\varepsilon}_{\omega
}(g)(p)=\omega_{p}$ $\vee%
%TCIMACRO{\tbigwedge \limits_{q\in\operatorname*{past}(p)}}%
%BeginExpansion
{\textstyle\bigwedge\limits_{q\in\operatorname*{past}(p)}}
%EndExpansion
g_{q},$ \newline The set $Z$ verifying $\overleftarrow{\varepsilon}_{\omega
}(g)>\omega$ is an overset of the regional minima
\end{itemize}

CEILING\_MINIMA is one of the following sets : $X,$ $X\wedge Y$ or $X\wedge
Z.$%

%TCIMACRO{\TeXButton{EndFrame}{\end{frame}}}%
%BeginExpansion
\end{frame}%
%EndExpansion
%

%TCIMACRO{\TeXButton{BeginFrame}{\begin{frame}}}%
%BeginExpansion
\begin{frame}%
%EndExpansion
%

%TCIMACRO{\QTR{frametitle}{The Dijkstra algorithm with a reduced initialisation
%set of ceiling minima.}}%
%BeginExpansion
\frametitle{The Dijkstra algorithm with a reduced initialisation set of
ceiling minima.}%
%EndExpansion

A binary tag $\zeta:$ $\zeta$ $=0$ for nodes with an unknown flooding level;
$\zeta=1$ after the first time the flooding level is computed.

\textbf{Initialization:}

\qquad create a HQ $\Phi$ ; $\tau=\infty$ ; $\zeta=0$

\qquad for each node $p$ belonging to $\widehat{S}=$"ceiling minima",

\qquad\qquad introduce $p$ into $\Phi$ with a priority $\omega_{p}$

\qquad\qquad$\tau_{p}=\omega_{p}$

\textbf{Flooding}

\qquad While $\Phi$ is not empty repeat:

\qquad\qquad Extract from $\Phi$ the node $j$ with the lowest prioriy
$\lambda$

\qquad\qquad If $\tau_{i}=$ $\lambda$

\qquad\qquad\qquad For any neighbor $i$ of $j$ such that $\zeta_{i}=0$ and
$\tau_{j}\vee e_{ji}<\tau_{i}$

\qquad\qquad\qquad\qquad$\tau_{i}=\tau_{j}\vee e_{ji}$ ; $\zeta_{i}=1$ ;
introduce $i$ into $\Phi$ with the priority $\tau_{i}$

End While%

%TCIMACRO{\TeXButton{EndFrame}{\end{frame}}}%
%BeginExpansion
\end{frame}%
%EndExpansion
%

%TCIMACRO{\TeXButton{BeginFrame}{\begin{frame}}}%
%BeginExpansion
\begin{frame}%
%EndExpansion
%

%TCIMACRO{\QTR{frametitle}{An equivalent algorithm with a 3 state tag}}%
%BeginExpansion
\frametitle{An equivalent algorithm with a 3 state tag}%
%EndExpansion

The following algorithm is equivalent and uses a 3 state flag: ("unknown",
"final", "in S") = ("u" , "f", "s") = $(0,1,2)$ with the following meanings:
a) "unknown"="u"$=0$ are the nodes whose flooding value has not been computed
yet; b) "final"="f"=1 computed nodes, not yet in $S$ ; c) "in S"= "s"$=2$ for
pixels introduced into $S$%

%TCIMACRO{\TeXButton{EndFrame}{\end{frame}}}%
%BeginExpansion
\end{frame}%
%EndExpansion
%

%TCIMACRO{\TeXButton{BeginFrame}{\begin{frame}}}%
%BeginExpansion
\begin{frame}%
%EndExpansion

\textbf{Flooding algorithm}

\textbf{Initialization:}

\qquad create a HQ $\Phi$ ; $\tau=\infty$ ; $\zeta=0$

\qquad for each node $p$ belonging to $\widehat{S}=$"ceiling minima",

\qquad\qquad introduce $p$ into $\Phi$ with a priority $\omega_{p}$

\qquad\qquad$\tau_{p}=\omega_{p}$

\textbf{Flooding}

\qquad While $\Phi$ is not empty repeat:

\qquad\qquad Extract from $\Phi$ the node $j$ with the lowest prioriy

\qquad\qquad if $\zeta_{j}<2$

\qquad\qquad\qquad$\zeta_{j}=2$

\qquad\qquad\qquad For any neighbor $i$ of $j$ such that $\zeta_{i}=0$

\qquad\qquad\qquad\qquad if $\tau_{j}\vee e_{ji}<\tau_{i}$ then $\tau_{i}%
=\tau_{j}\vee e_{ji}$

\qquad\qquad\qquad\qquad$\zeta_{i}=1$ ; introduce $i$ into $\Phi$ with the
priority $\tau_{i}$

End While%

%TCIMACRO{\TeXButton{EndFrame}{\end{frame}}}%
%BeginExpansion
\end{frame}%
%EndExpansion
%

%TCIMACRO{\TeXButton{BeginFrame}{\begin{frame}}}%
%BeginExpansion
\begin{frame}%
%EndExpansion
%

%TCIMACRO{\QTR{frametitle}{Illustration}}%
%BeginExpansion
\frametitle{Illustration}%
%EndExpansion

The figure presents in red a topographic surface and in green a ceiling
function.\ The successive lines below the figure present the evolution of the
algorithm. The algorithm is illustrated step by step. In each group of 3
lines\newline- the first represent the status of the node, "u"=unknow, "f" =
flooded, "s" in S\newline- the second represent the current estimated flooding
value of the nodes\newline- the third represent the nodes in $\Phi$%

%TCIMACRO{\TeXButton{EndFrame}{\end{frame}}}%
%BeginExpansion
\end{frame}%
%EndExpansion
%

%TCIMACRO{\TeXButton{BeginFrame}{\begin{frame}}}%
%BeginExpansion
\begin{frame}%
%EndExpansion
%

%TCIMACRO{\FRAME{ftbpFU}{2.4267in}{2.787in}{0pt}{\Qcb{}}{\Qlb{inonde1}%
%}{inonde1-eps-converted-to.pdf}{\special{ language "Scientific Word";  type "GRAPHIC";
%width 2.4267in;  height 2.787in;  depth 0pt;  original-width 5.1033in;
%original-height 5.8705in;  cropleft "0";  croptop "1";  cropright "1";
%cropbottom "0";  filename 'inonde1-eps-converted-to.pdf';file-properties "XNPEU";}}}%
%BeginExpansion
\begin{figure}
[ptb]
\begin{center}
\includegraphics[
height=2.787in,
width=2.4267in
]%
{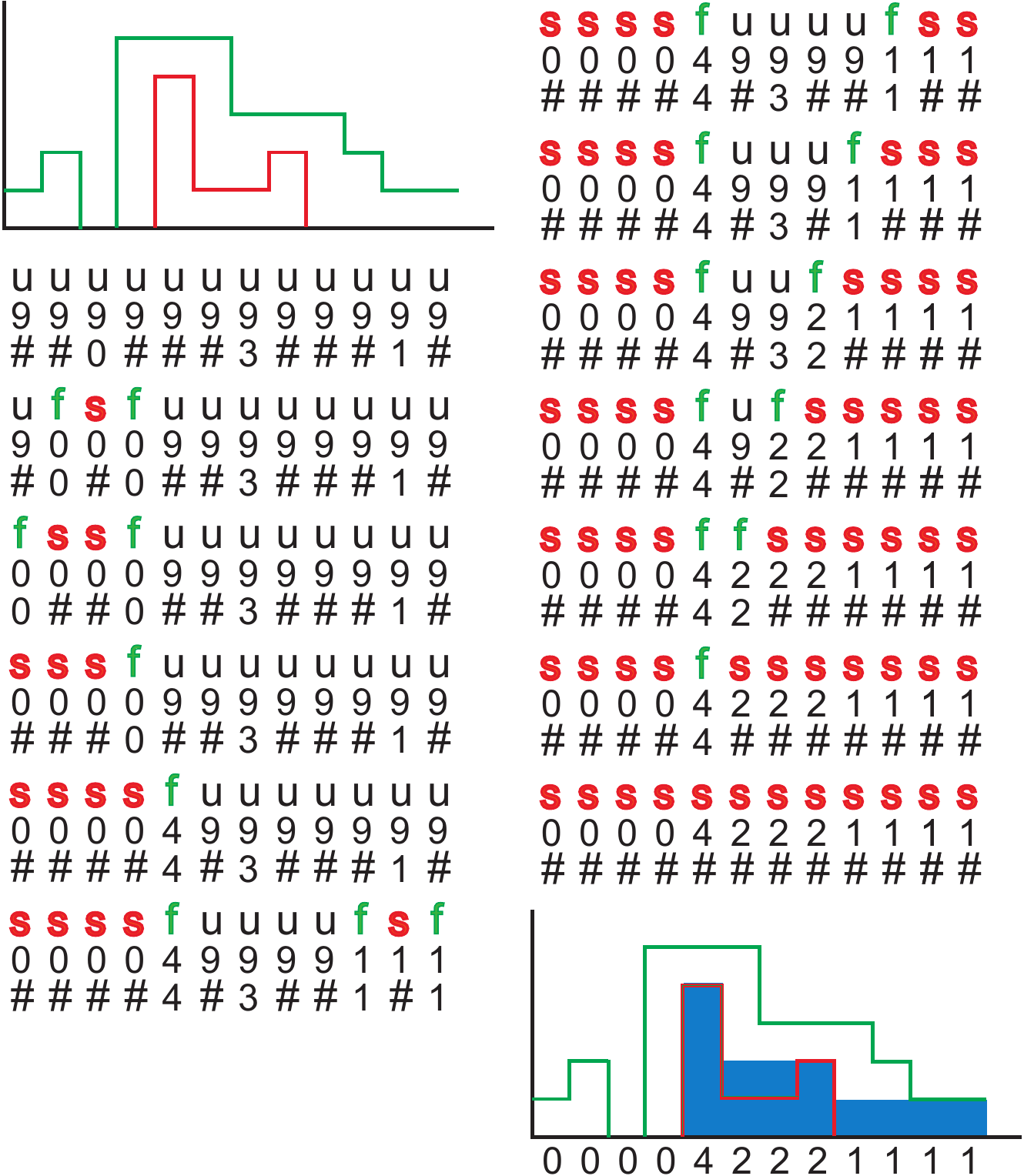}%
\label{inonde1}%
\end{center}
\end{figure}
%EndExpansion
%

%TCIMACRO{\TeXButton{EndFrame}{\end{frame}}}%
%BeginExpansion
\end{frame}%
%EndExpansion%
%TCIMACRO{\TeXButton{BeginFrame}{\begin{frame}}}%
%BeginExpansion
\begin{frame}%
%EndExpansion
%

%TCIMACRO{\QTR{frametitle}{Speeding up the flooding}}%
%BeginExpansion
\frametitle{Speeding up the flooding}%
%EndExpansion

In the Dijkstra algorithm, $S$ contains all nodes with a known flooding ; at
each iteration, the node with the smallest estimated flooding value is
introduced into $S.$ One node is introduced at each iteration of the
algorithm.\ Faster floodings are possible if one remarks:

\begin{enumerate}
\item if $f_{p}=f_{q},$ then $\tau_{p}=\tau_{q},$ as the flooding is a
connected operator.

\item If $f_{p}=\omega_{p},$ then ground and ceiling levels are identical,
hence $\tau_{p}=f_{p}=\omega_{p}$

\item if $f_{p}\vee f_{q}<\tau_{p}$ then $\tau_{q}=f_{q}.$ The proof is the
following.\ $\{f_{p}\vee f_{q}<\tau_{p}\}\Leftrightarrow\{f_{p}<\tau_{p}$ and
$f_{q}<\tau_{p}\}.\ $On the other hand the criterion for floodings $\{\tau
_{p}>\tau_{q}\}\Rightarrow\{f_{p}=\tau_{p}\}$ is equivalent with $\{f_{p}%
<\tau_{p}\}\Rightarrow\{\tau_{p}\leq\tau_{q}\}$.\ On the other hand $\tau
_{q}\leq\tau_{p}\vee f_{q}=\tau_{p}\vee f_{p}\vee f_{q}=\tau_{p}$ which shows
that $\tau_{p}=\tau_{q}$

\item if $f_{q}<\tau_{p}$ then $\tau_{q}\leq f_{q}\vee\tau_{p}=\tau_{p.\ }.\ $
\end{enumerate}

%

%TCIMACRO{\TeXButton{EndFrame}{\end{frame}}}%
%BeginExpansion
\end{frame}%
%EndExpansion
%

%TCIMACRO{\TeXButton{BeginFrame}{\begin{frame}}}%
%BeginExpansion
\begin{frame}%
%EndExpansion
%

%TCIMACRO{\QTR{frametitle}{Speeding up the algorithm of Dijkstra}}%
%BeginExpansion
\frametitle{Speeding up the algorithm of Dijkstra}%
%EndExpansion

The two following rules permit to speed up the algorithm of Dijkstra.

\begin{enumerate}
\item if $f_{q}\geq\tau_{p}$ then $\tau_{q}=f_{q}.$ The proof is the
following.\ For any neighboring nodes $p$ and $q$ the flooding levels verify
$\tau_{q}\leq\tau_{p}\vee f_{q}=f_{q}.\ $But as $\tau_{q}\geq f_{q}$ we get
$\tau_{q}=f_{q}.$

\item suppose that $(p,q)$ are neighbors and $p$ is the node of $\partial
^{-}S$ for which the flooding level $\tau_{p}$ is the lowest.\ If $f_{q}%
<\tau_{p},$ and if $q$ is to be flooded by a node in $S,$ this node
necessarily is $p$ and the estimated flooding level of $q$ is $\tau_{q}%
=\tau_{p}\vee f_{q}=\tau_{p}.$
\end{enumerate}

%

%TCIMACRO{\TeXButton{EndFrame}{\end{frame}}}%
%BeginExpansion
\end{frame}%
%EndExpansion
%

%TCIMACRO{\TeXButton{BeginFrame}{\begin{frame}}}%
%BeginExpansion
\begin{frame}%
%EndExpansion
%

%TCIMACRO{\QTR{frametitle}{The core expanding algorithm}}%
%BeginExpansion
\frametitle{The core expanding algorithm}%
%EndExpansion

We may now derive a fast algorithm from these remarks. Suppose that during the
flooding, the set $S$ represents all flooded nodes and $p$ is the node of
$\partial^{-}S$ for which the flooding level $\tau_{p}$ is the lowest.\ If
$f_{q}\geq\tau_{p}$ then $\tau_{q}=f_{q}.\ $If on the contrary $f_{q}<\tau
_{p},$ we apply Dijkstra's algorithm and introduce into $S$ the node with the
smallest estimate.\ If there exists in $\overline{S}$ a node $j $ with a
smaller ceiling value $\omega_{j}$ as $\tau_{p}:S=S\cup\{j\}.$ If not,
$\tau_{q}=\tau_{p}$ is the flooding estimation of a node in $\overline{S}$
which is the lowest and $S=S\cup\{q\}.$

This shows that in the case where $\omega_{j}\geq\tau_{p},$ all neighbors of
$p$ may be introduced at once into the set $S,$ yielding the following algorithm:%

%TCIMACRO{\TeXButton{EndFrame}{\end{frame}}}%
%BeginExpansion
\end{frame}%
%EndExpansion
%

%TCIMACRO{\TeXButton{BeginFrame}{\begin{frame}}}%
%BeginExpansion
\begin{frame}%
%EndExpansion
%

%TCIMACRO{\QTR{frametitle}{The core expanding algorithm}}%
%BeginExpansion
\frametitle{The core expanding algorithm}%
%EndExpansion

\textbf{Initialization:}

$\qquad S=\varnothing$ ;

\textbf{Flooding:}

While $S\neq N$ repeat:

\qquad Set $\lambda=\min_{i\in\overline{S}}\left[  \omega_{i}\right]  $ ; if
$j$ does not exist $\lambda=\infty$

\qquad Set $\mu=\tau_{p}$ for $p\in\partial^{-}S$ for which $\tau_{p}%
=\min_{i\in\partial^{-}S}\left[  \tau_{i}\right]  $

\qquad\qquad if $\lambda<\mu$ : $S=S\cup\{j\}$ and $\tau_{j}=\omega_{j}$

\qquad\qquad else

\qquad\qquad\qquad For each neighbor $q$ of $p$ in $\overline{S}$ do:

\qquad\qquad\qquad\qquad$\tau_{q}=\tau_{p}\vee f_{q}$

\qquad\qquad\qquad\qquad$S=S\cup\{q\}$

End While

At each iteration, the algorithm has to fetch the node $p$ with the smallest
flooding value in $\partial^{-}S$ which is easily done with a HQ ; the node
$j$ with the smallest ceiling value $\omega_{j.\ }$ can be easily done if the
nodes are ordered with increasing values in a FIFO.%

%TCIMACRO{\TeXButton{EndFrame}{\end{frame}}}%
%BeginExpansion
\end{frame}%
%EndExpansion
%

%TCIMACRO{\TeXButton{BeginFrame}{\begin{frame}}}%
%BeginExpansion
\begin{frame}%
%EndExpansion
%

%TCIMACRO{\QTR{frametitle}{Illustration of the core expanding algorithm}}%
%BeginExpansion
\frametitle{Illustration of the core expanding algorithm}%
%EndExpansion

The topographic surface is in red and the ceiling function in green.\ At
initialisation, the nodes with a label "i" are the nodes of the ceiling
minima.\ The smallest of them is introduced into $S$ and immediately expanded,
introducing both its neighbors into $S$.\ In line 5, another ceiling minima is
introduced into $S$ and immediately expanded.\ %

%TCIMACRO{\TeXButton{EndFrame}{\end{frame}}}%
%BeginExpansion
\end{frame}%
%EndExpansion
%

%TCIMACRO{\TeXButton{BeginFrame}{\begin{frame}}}%
%BeginExpansion
\begin{frame}%
%EndExpansion
%

%TCIMACRO{\FRAME{ftbpFU}{1.1391in}{3.1158in}{0pt}{\Qcb{expanding algorithm}}%
%{}{inond6-eps-converted-to.pdf}{\special{ language "Scientific Word";  type "GRAPHIC";
%width 1.1391in;  height 3.1158in;  depth 0pt;  original-width 2.4666in;
%original-height 6.8601in;  cropleft "0";  croptop "1";  cropright "1";
%cropbottom "0";  filename 'inond6-eps-converted-to.pdf';file-properties "XNPEU";}}}%
%BeginExpansion
\begin{figure}
[ptb]
\begin{center}
\includegraphics[
height=3.1158in,
width=1.1391in
]%
{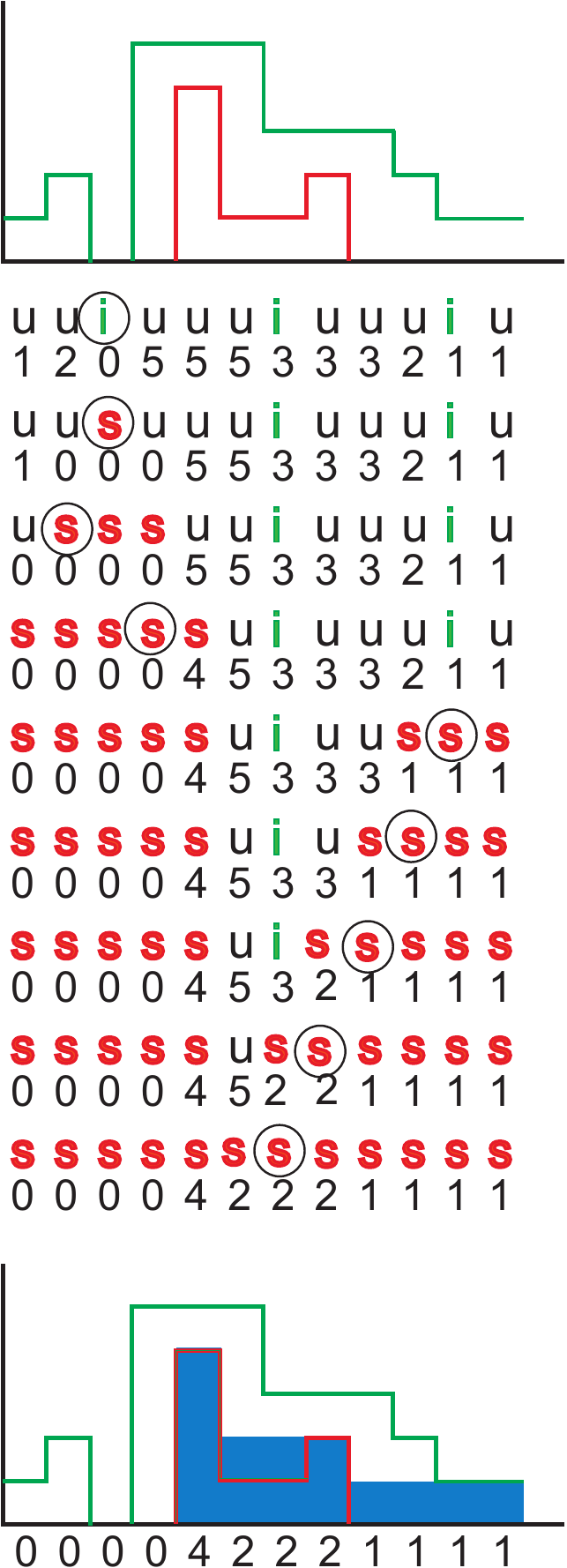}%
\caption{expanding algorithm}%
\end{center}
\end{figure}
%EndExpansion
%

%TCIMACRO{\TeXButton{EndFrame}{\end{frame}}}%
%BeginExpansion
\end{frame}%
%EndExpansion
%

%TCIMACRO{\TeXButton{BeginFrame}{\begin{frame}}}%
%BeginExpansion
\begin{frame}%
%EndExpansion

\begin{center}
{\Large \alert{The lakes of an ultrametric distance function form a
dendrogram}}
\end{center}

%

%TCIMACRO{\TeXButton{EndFrame}{\end{frame}}}%
%BeginExpansion
\end{frame}%
%EndExpansion
%

%TCIMACRO{\TeXButton{BeginFrame}{\begin{frame}}}%
%BeginExpansion
\begin{frame}%
%EndExpansion
%

%TCIMACRO{\QTR{frametitle}{Distance on a graph based on the maximal edge weight
%along the chain}}%
%BeginExpansion
\frametitle{Distance on a graph based on the maximal edge weight along the
chain}%
%EndExpansion

The weights are assigned to the edges, and represent their altitudes.

\textbf{Altitude of a chain: }The altitude of a chain is equal to the highest
weight of the edges along the chain.

\textbf{Flooding distance between two nodes: }The flooding distance
$\operatorname*{fldist}(x,y)$ between nodes $x$ and $y$ is equal to the
minimal altitude of all chains between $x$ and $y$. During a flooding process,
in which a source is placed at location $x,$ the flood would proceed along
this chain of minimal highest altitude to reach the pixel $y$. If there is no
chain between them, the level distance is equal to $\infty$.

\textbf{Triangular inequality :} For $(x,y,z):d(x,z)\leq d(x,y)\vee d(y,z)$ :
ultrametric inequality%

%TCIMACRO{\TeXButton{EndFrame}{\end{frame}}}%
%BeginExpansion
\end{frame}%
%EndExpansion
%

%TCIMACRO{\TeXButton{BeginFrame}{\begin{frame}}}%
%BeginExpansion
\begin{frame}%
%EndExpansion
%

%TCIMACRO{\QTR{frametitle}{The flooding distance is an ultrametric distance}}%
%BeginExpansion
\frametitle{The flooding distance is an ultrametric distance}%
%EndExpansion

An ultrametric distance verifies

* reflexivity : $d(x,x)=0$

* symmetry: $d(x,y)=d(y,x)$

* ultrametric inequality: for all $x,y,z:d(x,y)\leq max\{d(x,z),d(z,y)$\} :
the lowest lake containing both $x$ and $y$ is lower or equal than the lowest
lake containing $x,$ $y$ and $z.$%

%TCIMACRO{\TeXButton{EndFrame}{\end{frame}}}%
%BeginExpansion
\end{frame}%
%EndExpansion
%

%TCIMACRO{\TeXButton{BeginFrame}{\begin{frame}}}%
%BeginExpansion
\begin{frame}%
%EndExpansion
%

%TCIMACRO{\QTR{frametitle}{The balls of an ultrametric distance}}%
%BeginExpansion
\frametitle{The balls of an ultrametric distance}%
%EndExpansion

For $p\in E$ the closed ball of centre $p$ and radius $\rho$ is defined by
$\operatorname*{Ball}(p,\rho)=\left\{  q\in E\mid d(p,q)\leq\rho\right\}
.\;$The open ball of centre $p$ and radius $\rho$ is defined by $\overset
{\circ}{\operatorname*{Ball}}(p,\rho)=\left\{  q\in E\mid d(p,q)<\rho\right\}
.$

Every triangle is isosceles. Let us consider three distinct points $p,q,r$ and
suppose that the largest edge of this triangle is $pq.\ $Then $d(p,q)\leq
d(p,r)\vee d(r,q),$ showing that the two larges edges of the triangle have the
same length.\ 

\begin{lemma}
Each element of a closed ball $\operatorname*{Ball}(p,\rho)$ is centre of this ball
\end{lemma}

\textbf{Proof: }Suppose that $q$ is an element of $\operatorname*{Ball}%
(p,\rho)$.\ Let us show that then $q$ also is centre of this ball.\ If
$r\in\operatorname*{Ball}(p,\rho):$ $d(q,r)\leq\max\left\{
d(q,p),d(p,r)\right\}  =\rho,$ hence $r\in\operatorname*{Ball}(q,\rho),$
showing that $\operatorname*{Ball}(p,\rho)\subset\operatorname*{Ball}%
(q,\rho).\ $Exchanging the roles of $p$ and $q$ shows that
$\operatorname*{Ball}(p,\rho)=\operatorname*{Ball}(q,\rho)$%

%TCIMACRO{\TeXButton{EndFrame}{\end{frame}}}%
%BeginExpansion
\end{frame}%
%EndExpansion
%

%TCIMACRO{\TeXButton{BeginFrame}{\begin{frame}}}%
%BeginExpansion
\begin{frame}%
%EndExpansion
%

%TCIMACRO{\QTR{frametitle}{The balls of an ultrametric distance}}%
%BeginExpansion
\frametitle{The balls of an ultrametric distance}%
%EndExpansion

\begin{lemma}
Two closed balls $\operatorname*{Ball}(p,\rho)$ and $\operatorname*{Ball}%
(q,\rho)$ with the same radius are either disjoint or identical.\ 
\end{lemma}

\textbf{Proof: }If $\operatorname*{Ball}(p,\rho)$ and $\operatorname*{Ball}%
(q,\rho)$ are not disjoint, then they contain at least one common point
$r.\ $According to the preceding lemma, $r$ is then centre of both balls
$\operatorname*{Ball}(p,\rho)$ and $\operatorname*{Ball}(q,\rho)$, showing
that they are identical.\ 

\begin{lemma}
The radius of a ball is equal to its diameter.
\end{lemma}

\textbf{Proof: }Let $\operatorname*{Ball}(p,\rho)$ be a ball of radius
$\rho.\ $Let $q$ and $r$ be two nodes with the largest distance in
$\operatorname*{Ball}(p,\rho).\ $This distance $\lambda$ is called diameter of
the ball and verifies : $\lambda=d(q,r)\leq d(q,p)\vee d(p,r)\leq\rho.\;$Hence
$\lambda\leq\rho.$ If there exists two nodes in $\operatorname*{Ball}(p,\rho)$
with a distance equal to $\rho$, then $\lambda\geq\rho$.\ In this case
$\lambda=\rho.$%

%TCIMACRO{\TeXButton{EndFrame}{\end{frame}}}%
%BeginExpansion
\end{frame}%
%EndExpansion%
%TCIMACRO{\TeXButton{BeginFrame}{\begin{frame}}}%
%BeginExpansion
\begin{frame}%
%EndExpansion

\begin{center}
{\Large \alert{Reminders on dendrograms}}
\end{center}

%

%TCIMACRO{\TeXButton{EndFrame}{\end{frame}}}%
%BeginExpansion
\end{frame}%
%EndExpansion
%

%TCIMACRO{\TeXButton{BeginFrame}{\begin{frame}}}%
%BeginExpansion
\begin{frame}%
%EndExpansion
%

%TCIMACRO{\QTR{frametitle}{The structure associated to an order relation}}%
%BeginExpansion
\frametitle{The structure associated to an order relation}%
%EndExpansion

$E$ : a domain with a finite number of elements called points

$\mathcal{X}$ : a subset of $\mathcal{P}(E)$, with the order relation
$\subset$

$\operatorname*{supp}(\mathcal{X}):$ union of all sets belonging to
$\mathcal{X}$ is called support of $\mathcal{X}$ : $\operatorname*{supp}%
(\mathcal{X}).$

The subsets of $\mathcal{X}$ may be structured into:%

%TCIMACRO{\TeXButton{EndFrame}{\end{frame}}}%
%BeginExpansion
\end{frame}%
%EndExpansion
%

%TCIMACRO{\TeXButton{BeginFrame}{\begin{frame}}}%
%BeginExpansion
\begin{frame}%
%EndExpansion

\begin{itemize}
\item the summits : $\operatorname*{Sum}(\mathcal{X})=\{A\in\mathcal{X}%
\mid\forall B\in\mathcal{X}:$ $A\subset B$ $\Rightarrow A=B\}$

\item the leaves : $\operatorname*{Leav}(\mathcal{X})=\{A\in\mathcal{X}%
\mid\forall B\in\mathcal{X}:$ $B\subset A$ $\Rightarrow A=B\}$

\item the predecessors : $\operatorname*{Pred}(A)=\{B\in\mathcal{X}\mid$
$A\subset B\}$

\item the immediate predecessors : $\operatorname*{ImPred}(A)=\{B\in
\mathcal{X}\mid\left\{  U\mid U\in\mathcal{X},A\subset U\text{ and }U\subset
B\right\}  =(A,B)\}$

\item the successors : $\operatorname*{Succ}(A)=\{B\in\mathcal{X}\mid B\subset
A\} $

\item the immediate successors : $\operatorname*{ImSucc}(A)=\{B\in
\mathcal{X}\mid\left\{  U\mid U\in\mathcal{X},B\subset U\text{ and }U\subset
A\right\}  =(A,B)\}$

\item the uncles : $\operatorname*{uncle}(A)=\{B\in\mathcal{X}\mid
\operatorname*{ImPred}(B)\in\operatorname*{Pred}(A),B\notin
\operatorname*{Pred}(A),\operatorname*{ImPred}(B)\neq\operatorname*{ImPred}%
(A)\}$

\item the brothers : $\operatorname*{brother}(A)=\{B\in\mathcal{X}%
\mid\operatorname*{ImPred}(B)\in\operatorname*{Pred}(A),B\notin
\operatorname*{Pred}(A),\operatorname*{ImPred}(B)=\operatorname*{ImPred}(A)\}$
\end{itemize}

%

%TCIMACRO{\TeXButton{EndFrame}{\end{frame}}}%
%BeginExpansion
\end{frame}%
%EndExpansion
%

%TCIMACRO{\TeXButton{BeginFrame}{\begin{frame}}}%
%BeginExpansion
\begin{frame}%
%EndExpansion
%

%TCIMACRO{\QTR{frametitle}{Dendrograms}}%
%BeginExpansion
\frametitle{Dendrograms}%
%EndExpansion

\begin{definition}
$\mathcal{X}$ is a dendrogram if and only if the set $\operatorname*{Pred}(A)$
of the predecessors of $A,$ with the order relation induced by $\subset$ is a
total order.\ 
\end{definition}

The maximal element of this family is a summit, which is the unique summit
containing $A.$ The smallest element is $\operatorname*{ImPred}(A)$, the
father of $A,$ which is unique.\ 

The arcs point from each element of the dendrogram to its immediate successor.%

%TCIMACRO{\FRAME{ftbpFU}{1.4911in}{0.89in}{0pt}{\Qcb{A dendrogram}%
%}{\Qlb{dendro}}{dendro-eps-converted-to.pdf}{\special{ language "Scientific Word";
%type "GRAPHIC";  maintain-aspect-ratio TRUE;  display "USEDEF";
%valid_file "F";  width 1.4911in;  height 0.89in;  depth 0pt;
%original-width 2.8684in;  original-height 1.6895in;  cropleft "0";
%filename 'dendro-eps-converted-to.pdf';file-properties "XNPEU";}}}%
%BeginExpansion
\begin{figure}
[ptb]
\begin{center}
\includegraphics[
height=0.89in,
width=1.4911in
]%
{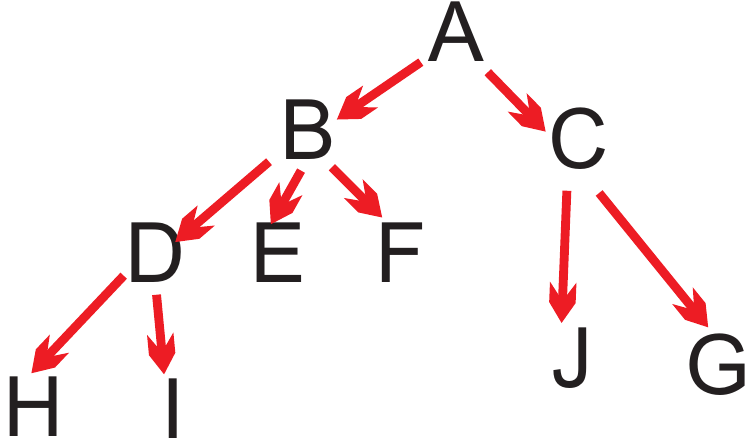}%
\caption{A dendrogram}%
\label{dendro}%
\end{center}
\end{figure}
%EndExpansion
%

%TCIMACRO{\TeXButton{EndFrame}{\end{frame}}}%
%BeginExpansion
\end{frame}%
%EndExpansion
%

%TCIMACRO{\TeXButton{BeginFrame}{\begin{frame}}}%
%BeginExpansion
\begin{frame}%
%EndExpansion
%

%TCIMACRO{\QTR{frametitle}{Characterizations of a dendrogram}}%
%BeginExpansion
\frametitle{Characterizations of a dendrogram}%
%EndExpansion

The following properties are equivalent:\newline1)$\mathcal{X}$ is a
dendrogram \newline2) $U,V,A\in\mathcal{X\ }:A\subset U$ and $A\subset V$
$\Rightarrow U\subset V$ or $V\subset U$\newline3) $U,V\in\mathcal{X}%
:U\nsubseteqq V$ and $V\nsubseteqq U\Rightarrow U\cap V=\varnothing$%

%TCIMACRO{\TeXButton{EndFrame}{\end{frame}}}%
%BeginExpansion
\end{frame}%
%EndExpansion
%

%TCIMACRO{\TeXButton{BeginFrame}{\begin{frame}}}%
%BeginExpansion
\begin{frame}%
%EndExpansion
%

%TCIMACRO{\QTR{frametitle}{A dendrogram of lakes}}%
%BeginExpansion
\frametitle{A dendrogram of lakes}%
%EndExpansion

Due to their particular properties, the closed balls of an ultrametric
distance function form a dendrogram. Consider a particular closed ball
$A=\operatorname*{Ball}(p,\rho).\ $We have to show that $\operatorname*{Pred}%
(A)$ is completely ordered for $\subset.\ $Consider two predecessors of $A,$ a
ball $B=\operatorname*{Ball}(q,\lambda)$ and a ball $C=\operatorname*{Ball}%
(s,\mu).\ $As the node $p$ belongs to both balls $B$ and $C,$ it is also
center of these balls.\ Thus $B$ and $C$ are two balls with the same center
$p$, and $\operatorname*{Ball}(p,\lambda\wedge\mu)\subset\operatorname*{Ball}%
(p,\lambda\vee\mu).$%

%TCIMACRO{\TeXButton{EndFrame}{\end{frame}}}%
%BeginExpansion
\end{frame}%
%EndExpansion
%

%TCIMACRO{\TeXButton{BeginFrame}{\begin{frame}}}%
%BeginExpansion
\begin{frame}%
%EndExpansion

\begin{center}
{\Large \alert{Creation of a dendrogram of lakes}}
\end{center}

%

%TCIMACRO{\TeXButton{EndFrame}{\end{frame}}}%
%BeginExpansion
\end{frame}%
%EndExpansion
%

%TCIMACRO{\TeXButton{BeginFrame}{\begin{frame}}}%
%BeginExpansion
\begin{frame}%
%EndExpansion
%

%TCIMACRO{\QTR{frametitle}{The creation of lakes}}%
%BeginExpansion
\frametitle{The creation of lakes}%
%EndExpansion

If the shortest path between $\Omega$ and $p$ passes through $q$ we have
$\tau_{p}=\tau_{q}\vee d(p,q)$ if not $\tau_{p}<\tau_{q}\vee d(p,q).$

\begin{lemma}
Any two nodes of an edge weighted graph verify $\tau_{p}\leq\tau_{q}\vee
d(p,q).$
\end{lemma}

Suppose that if $d(p,q)<\tau_{p}.\ $Then $d(p,q)<\tau_{p}\leq\tau_{q}\vee
d(p,q)$ implies $\tau_{p}\leq\tau_{q}.$ So $d(p,q)<\tau_{p}\leq\tau_{q}$ which
similarly implies $\tau_{q}\leq\tau_{p}.$ Hence $\tau_{p}=\tau_{q}$ which is
compatible with the laws of hydrostatics.\ 

\begin{lemma}
If two nodes $p$ and $q$ of an edge weighted graph verify $d(p,q)<\tau_{p}$ or
$d(p,q)<\tau_{q},$ then $\tau_{p}=\tau_{q}.$
\end{lemma}

%

%TCIMACRO{\TeXButton{EndFrame}{\end{frame}}}%
%BeginExpansion
\end{frame}%
%EndExpansion
%

%TCIMACRO{\TeXButton{BeginFrame}{\begin{frame}}}%
%BeginExpansion
\begin{frame}%
%EndExpansion
%

%TCIMACRO{\QTR{frametitle}{The creation of lakes}}%
%BeginExpansion
\frametitle{The creation of lakes}%
%EndExpansion

\begin{lemma}
If an open ball $\overset{\circ}{\operatorname*{Ball}}(p,\lambda)$ has one
node with a flooding level $\mu\geq\lambda,$ then its flooding level is
uniform and equal to $\mu.$
\end{lemma}

\textbf{Proof: }Suppose that the node $s$ in $\overset{\circ}%
{\operatorname*{Ball}}(p,\lambda)$ verifies $\tau_{s}\geq\lambda.\ $The node
$s$ as any node of an open ball is center of this ball. If $q\in\overset
{\circ}{\operatorname*{Ball}}(s,\lambda),$ we have $d(s,q)<\lambda\leq\tau
_{s}.\ $Applying the preceding lemma yields $\tau_{s}=\tau_{q}.$ As this is
true for each node of $\overset{\circ}{\operatorname*{Ball}}(p,\lambda),$ we
have shown that the flooding level is constant and equal to $\tau_{s}$ on the
entire ball $\overset{\circ}{\operatorname*{Ball}}(p,\lambda).$

In particular if $\tau_{p}=\lambda,$ then the flooding level in $\overset
{\circ}{\operatorname*{Ball}}(p,\lambda)$ is equal to $\lambda.$%

%TCIMACRO{\TeXButton{EndFrame}{\end{frame}}}%
%BeginExpansion
\end{frame}%
%EndExpansion
%

%TCIMACRO{\TeXButton{BeginFrame}{\begin{frame}}}%
%BeginExpansion
\begin{frame}%
%EndExpansion
%

%TCIMACRO{\QTR{frametitle}{The extension of the lakes}}%
%BeginExpansion
\frametitle{The extension of the lakes}%
%EndExpansion

Define $\varepsilon_{e}(X):$ the lowest edge of the cocycle of $X$.\ 

For $\tau_{p}=\lambda,$ the flooding level on $\overset{\circ}%
{\operatorname*{Ball}}(p,\lambda)$ is constant and equal to $\lambda.\ $As
long as $\tau_{p}<\varepsilon_{e}(\overset{\circ}{\operatorname*{Ball}%
}(p,\lambda)),$ the extension of $\overset{\circ}{\operatorname*{Ball}}%
(p,\tau_{p})$ remains the same, the flooding is uniform and equal to $\tau
_{p}$ ; $\overset{\circ}{\operatorname*{Ball}}(p,\tau_{p})$ is a regional
minimum, as all its nodes have the same weight, its the distance between its
nodes is smaller than $\tau_{p},$ and its cocycle edges are higher than
$\tau_{p.}.$\bigskip

For $\tau_{p}=\varepsilon_{e}(\overset{\circ}{\operatorname*{Ball}}%
(p,\lambda))=\mu$ there exists an edge in the cocycle of $\overset{\circ
}{\operatorname*{Ball}}(p,\lambda)$ with a weight $\mu.$ The closed ball
$\operatorname*{Ball}(p,\mu)$ strictly contains $\overset{\circ}%
{\operatorname*{Ball}}(p,\lambda).\ $As $\tau_{p}=\mu,$ then for any other
node $s$ in $\operatorname*{Ball}(p,\mu)$ we have $\tau_{s}\leq\tau_{p}\vee
d(p,s)\leq\mu.$ However on $\overset{\circ}{\operatorname*{Ball}}%
(p,\mu)\subset$ $\operatorname*{Ball}(p,\mu)$ the flood is constant and equal
to $\tau_{p}=\mu$.%

%TCIMACRO{\TeXButton{EndFrame}{\end{frame}}}%
%BeginExpansion
\end{frame}%
%EndExpansion
%

%TCIMACRO{\TeXButton{BeginFrame}{\begin{frame}}}%
%BeginExpansion
\begin{frame}%
%EndExpansion
%

%TCIMACRO{\QTR{frametitle}{The extension of the lakes}}%
%BeginExpansion
\frametitle{The extension of the lakes}%
%EndExpansion

For any value $\sigma>\mu,$ we have $\operatorname*{Ball}(p,\mu)\subset$
$\overset{\circ}{\operatorname*{Ball}}(p,\sigma)\ ;$ if $\tau_{p}=\sigma$ the
flood is constant on $\overset{\circ}{\operatorname*{Ball}}(p,\sigma)$ and as
$\operatorname*{Ball}(p,\mu)\subseteq$ $\overset{\circ}{\operatorname*{Ball}%
}(p,\sigma)$ the flood also is constant on $\operatorname*{Ball}(p,\mu).$

In fact as long as $\sigma<\varepsilon_{e}(\operatorname*{Ball}(p,\mu)),$
$\operatorname*{Ball}(p,\mu)=$ $\overset{\circ}{\operatorname*{Ball}}%
(p,\sigma).\ $

\begin{lemma}
If there is at least one node with a weight $\mu$ in a closed ball
$\operatorname*{Ball}(p,\mu)$ of level $\mu,$ all other nodes in this ball
have a flooding level $\leq\mu.$
\end{lemma}

The diameter of $Y$ is $\lambda.$ Such a closed ball is called lake zone of
level $\lambda,$ as the level of flooding inside is $\leq\lambda$.%

%TCIMACRO{\TeXButton{EndFrame}{\end{frame}}}%
%BeginExpansion
\end{frame}%
%EndExpansion
%

%TCIMACRO{\TeXButton{BeginFrame}{\begin{frame}}}%
%BeginExpansion
\begin{frame}%
%EndExpansion
%

%TCIMACRO{\QTR{frametitle}{The growing extension of a lake containing a
%particular node}}%
%BeginExpansion
\frametitle{The growing extension of a lake containing a particular node}%
%EndExpansion

The extension of the lakes containing a node $p$ as its flooding level $\eta$
increases is:

- for $\eta<\varepsilon_{ne}p,$ the lake $X_{0}=\{p\}$ is a regional minimum lake.

- for $\eta=\varepsilon_{ne}p,$ the lake $X_{1}=\operatorname*{Ball}%
(p,\varepsilon_{ne}p)$ is a lake zone. The flood level is equal to $\eta$ on
$X_{0}$ and $\leq\eta$ everywhere else on $X_{1}.$ We have
$\operatorname*{diam}(X_{1})=\varepsilon_{ne}p=\varepsilon_{e}X_{0}.$

- for $\operatorname*{diam}(X_{1})<\eta<\varepsilon_{e}X_{1},$ the lake
$X_{2}=\operatorname*{Ball}(p,\eta)$ is a regional minimum lake with the
extension $X_{1}.$

- for $\eta=\varepsilon_{e}X_{1},$ the lake $X_{2}=\operatorname*{Ball}%
(p,\varepsilon_{e}X_{1})$ is a lake zone. The flood level is equal to $\eta$
on $X_{1}$ and $\leq\eta$ everywhere else on $X_{2}.$ We have
$\operatorname*{diam}(X_{2})=\varepsilon_{e}X_{1}$

- ...

- the alternating series of regional minima lakes and lake zones goes on until
all nodes of $N$ are flooded.\ %

%TCIMACRO{\TeXButton{EndFrame}{\end{frame}}}%
%BeginExpansion
\end{frame}%
%EndExpansion
%

%TCIMACRO{\TeXButton{BeginFrame}{\begin{frame}}}%
%BeginExpansion
\begin{frame}%
%EndExpansion
%

%TCIMACRO{\QTR{frametitle}{Dominated flooding on a dendrogram}}%
%BeginExpansion
\frametitle{Dominated flooding on a dendrogram}%
%EndExpansion

We now define the level of the dominated flooding under $\omega$ on the
dendrogram of the lakes.\ Define $\omega(X)$ the smallest value taken by the
ceiling function $\omega$ on $X$. The lakes containing the node $p$ form an
increasing series of nested sets $\kappa^{(n)}\{p\},$ the smallest being
$\{p\},$ the largest being the root $\kappa^{(m)}\{p\}$ of the dendrogram.

The operator $\omega(X)$ is decreasing and the operator $\operatorname*{diam}%
(X)$ increasing with $X.\ $As the series $\kappa^{(n)}\{p\}$ is increasing
with $n,$ we get a series of decreasing values $\omega(\kappa^{(n)}\{p\})$ and
a series of increasing values $\operatorname*{diam}(\kappa^{(n)}\{p\}):$

a) as the set $\{p\}$ has no inside edge, we have $\operatorname*{diam}%
(\kappa^{(0)}\{p\})=\operatorname*{diam}\{p\}=-\infty.\ $Hence $\omega
\{p\}>\operatorname*{diam}\{p\}=-\infty$\bigskip

b) if $\kappa^{(m)}\{p\}$ is the root of the dendrogram and at the root we
still have $\omega(\kappa^{(m)}\{p\})>\operatorname*{diam}(\kappa
^{(m)}\{p\}),$ i.e. the ceiling of $p$ is higher than the root of the
dendrogram, then $\tau_{p}=\omega(\kappa^{(m)}\{p\})$%

%TCIMACRO{\TeXButton{EndFrame}{\end{frame}}}%
%BeginExpansion
\end{frame}%
%EndExpansion
%

%TCIMACRO{\TeXButton{BeginFrame}{\begin{frame}}}%
%BeginExpansion
\begin{frame}%
%EndExpansion
%

%TCIMACRO{\FRAME{ftbpFU}{3.8489in}{2.0266in}{0pt}{\Qcb{Left: Dendrogram
%associated to a MST. All nodes have a ceiling function equal to $\infty$
%excepting the nodes $c$ and $h,$ with values $6$ and $1.$\newline Right: Each
%node of the dendrogram is assigned a ceiling value equal to the minimum
%ceiling value of all leaves it contains}}{\Qlb{mstfl16}}{mstfl16-eps-converted-to.pdf}%
%width 3.8489in;  height 2.0266in;  depth 0pt;  original-width 8.4906in;
%original-height 4.4417in;  cropleft "0";  croptop "1";  cropright "1";
%cropbottom "0";  filename 'mstfl16-eps-converted-to.pdf';file-properties "XNPEU";}}}%
%BeginExpansion
\begin{figure}
[ptb]
\begin{center}
\includegraphics[
height=2.0266in,
width=3.8489in
]%
{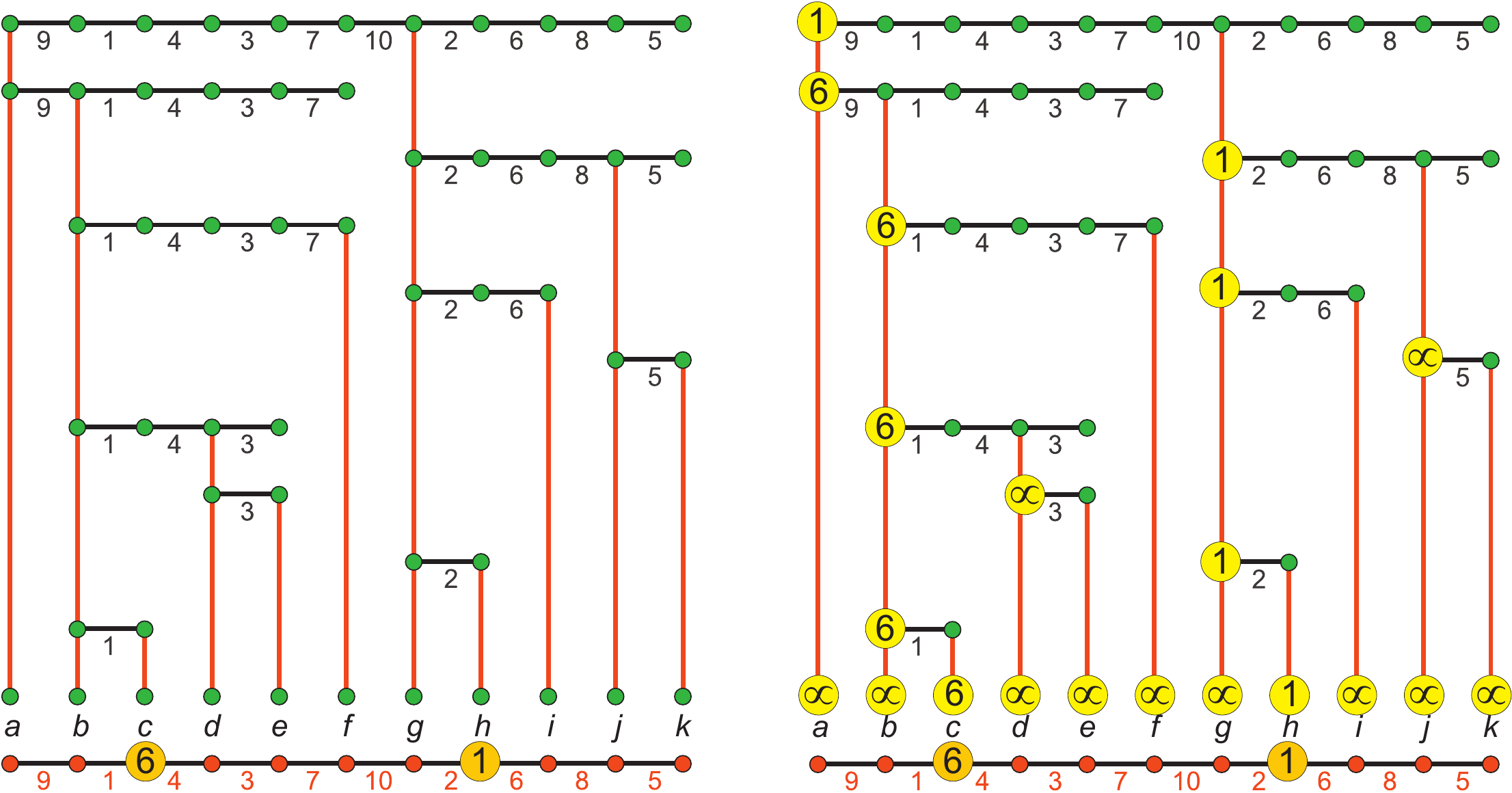}%
\caption{Left: Dendrogram associated to a MST. All nodes have a ceiling
function equal to $\infty$ excepting the nodes $c$ and $h,$ with values $6$
and $1.$\newline Right: Each node of the dendrogram is assigned a ceiling
value equal to the minimum ceiling value of all leaves it contains}%
\label{mstfl16}%
\end{center}
\end{figure}
%EndExpansion
%

%TCIMACRO{\TeXButton{EndFrame}{\end{frame}}}%
%BeginExpansion
\end{frame}%
%EndExpansion%
%TCIMACRO{\TeXButton{BeginFrame}{\begin{frame}}}%
%BeginExpansion
\begin{frame}%
%EndExpansion
%

%TCIMACRO{\QTR{frametitle}{Dominated flooding on a dendrogram}}%
%BeginExpansion
\frametitle{Dominated flooding on a dendrogram}%
%EndExpansion

c) In the other cases, there exists an index $k<m$ such that:

$\omega(\kappa^{(m)}\{p\})\leq\operatorname*{diam}(\kappa^{(m)}\{p\}),$ let
$k\leq m$ be the smallest index for which $\omega(\kappa^{(k)}\{p\})\leq
\operatorname*{diam}(\kappa^{(k)}\{p\})$ (rel.\ 1)\newline%
$\operatorname*{diam}(\kappa^{(k-1)}\{p\})<\omega(\kappa^{(k-1)}\{p\})\leq$
$\operatorname*{diam}(\kappa^{(k)}\{p\})=\varepsilon_{e}(\kappa^{(k-1)}\{p\})$

The previous relation implies that $\tau_{\kappa^{(k-1)}\{p\}}=\omega
(\kappa^{(k-1)}\{p\}$ and on $\kappa^{(k)}\{p\}$ the maximal flooding level is
$\operatorname*{diam}(\kappa^{(k)}\{p\}.$

In particular if $Y$ is a brother of $\kappa^{(k-1)}\{p\}$ then $Y$ is the
root of a sub-dendrogram which may be processed independently, with
$\omega(Y)=\operatorname*{diam}(\kappa^{(k)}\{p\}\wedge\omega(Y)$%

%TCIMACRO{\TeXButton{EndFrame}{\end{frame}}}%
%BeginExpansion
\end{frame}%
%EndExpansion
%

%TCIMACRO{\TeXButton{BeginFrame}{\begin{frame}}}%
%BeginExpansion
\begin{frame}%
%EndExpansion
%

%TCIMACRO{\QTR{frametitle}{Dominated flooding on a dendrogram}}%
%BeginExpansion
\frametitle{Dominated flooding on a dendrogram}%
%EndExpansion

Each uncle $Y_{i}$ of $\kappa^{(k)}\{p\}$ with a father $\kappa^{(l)}\{p\}, $
$l>k$ becomes the root of sub-dendrogram which may be processed independently,
and with a ceiling level $\omega(Y_{i})=\operatorname*{diam}(\kappa
^{(l)}\{p\}\wedge\omega(Y_{i}).$

This process cuts the upstream of $\kappa^{(k)}\{p\}$ in a number of
sub-dendrograms which may then be processed independently one from another.%

%TCIMACRO{\TeXButton{EndFrame}{\end{frame}}}%
%BeginExpansion
\end{frame}%
%EndExpansion
%

%TCIMACRO{\TeXButton{BeginFrame}{\begin{frame}}}%
%BeginExpansion
\begin{frame}%
%EndExpansion
%

%TCIMACRO{\QTR{frametitle}{Illustration}}%
%BeginExpansion
\frametitle{Illustration}%
%EndExpansion
%

%TCIMACRO{\FRAME{ftbpF}{4.1644in}{2.1146in}{0pt}{}{}{dendrab-eps-converted-to.pdf}%
%width 4.1644in;  height 2.1146in;  depth 0pt;  original-width 8.8136in;
%original-height 4.7347in;  cropleft "0";  croptop "1";  cropright "1";
%cropbottom "0";  filename 'dendrAB-eps-converted-to.pdf';file-properties "XNPEU";}}}%
%BeginExpansion
\begin{figure}
[ptb]
\begin{center}
\includegraphics[
height=2.1146in,
width=4.1644in
]%
{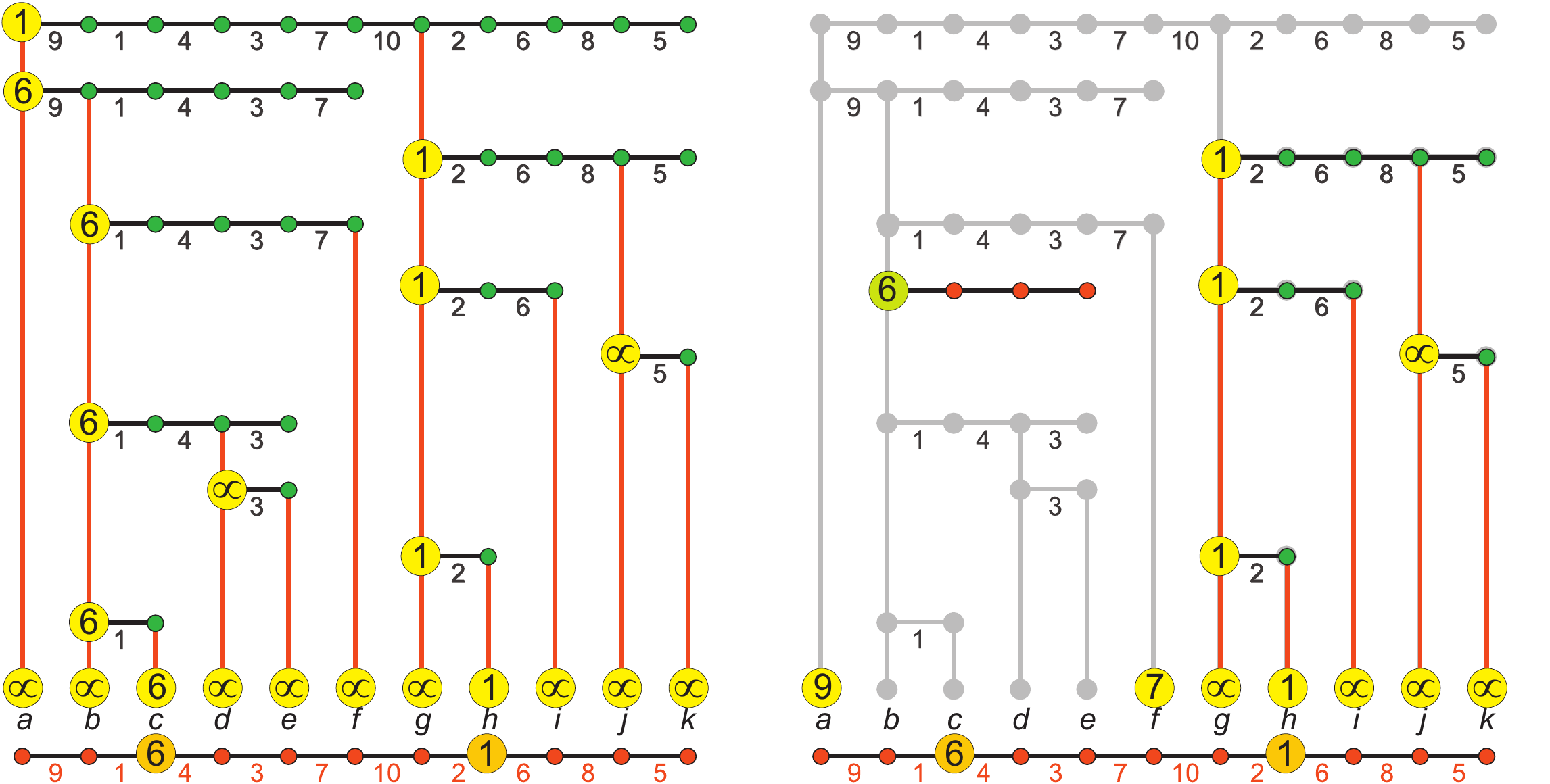}%
\end{center}
\end{figure}
%EndExpansion
%

%TCIMACRO{\TeXButton{EndFrame}{\end{frame}}}%
%BeginExpansion
\end{frame}%
%EndExpansion
%

%TCIMACRO{\TeXButton{BeginFrame}{\begin{frame}}}%
%BeginExpansion
\begin{frame}%
%EndExpansion

What about the lake containing the node $c\ ?$The smallest index for which
$\omega(\kappa^{(k)}\{c\})\leq\operatorname*{diam}(\kappa^{(k)}\{c\})$, is
$k=3, $ with $\kappa^{(3)}\{c\}=[b,c,d,e,f]$ having a diameter $7,$ whereas
$\omega(\kappa^{(3)}\{c\})=6.$ For $k=2,$ we get $\kappa^{(2)}\{c\}=[b,c,d,e]$
having a diameter $4,$ whereas $\omega(\kappa^{(2)}\{c\})=6.$ Hence:
$\kappa^{(2)}\{c\}=[b,c,d,e]$ is $\tau_{c}=\tau_{\kappa^{(2)}\{c\}}%
=\omega(\kappa^{(2)}\{c\})=6$%

%TCIMACRO{\TeXButton{EndFrame}{\end{frame}}}%
%BeginExpansion
\end{frame}%
%EndExpansion
%

%TCIMACRO{\TeXButton{BeginFrame}{\begin{frame}}}%
%BeginExpansion
\begin{frame}%
%EndExpansion
%

%TCIMACRO{\FRAME{ftbpF}{4.1644in}{2.0772in}{0pt}{}{}{dendrcd-eps-converted-to.pdf}%
%width 4.1644in;  height 2.0772in;  depth 0pt;  original-width 8.8011in;
%original-height 4.3603in;  cropleft "0";  croptop "1";  cropright "1";
%cropbottom "0";  filename 'dendrCD-eps-converted-to.pdf';file-properties "XNPEU";}}}%
%BeginExpansion
\begin{figure}
[ptb]
\begin{center}
\includegraphics[
height=2.0772in,
width=4.1644in
]%
{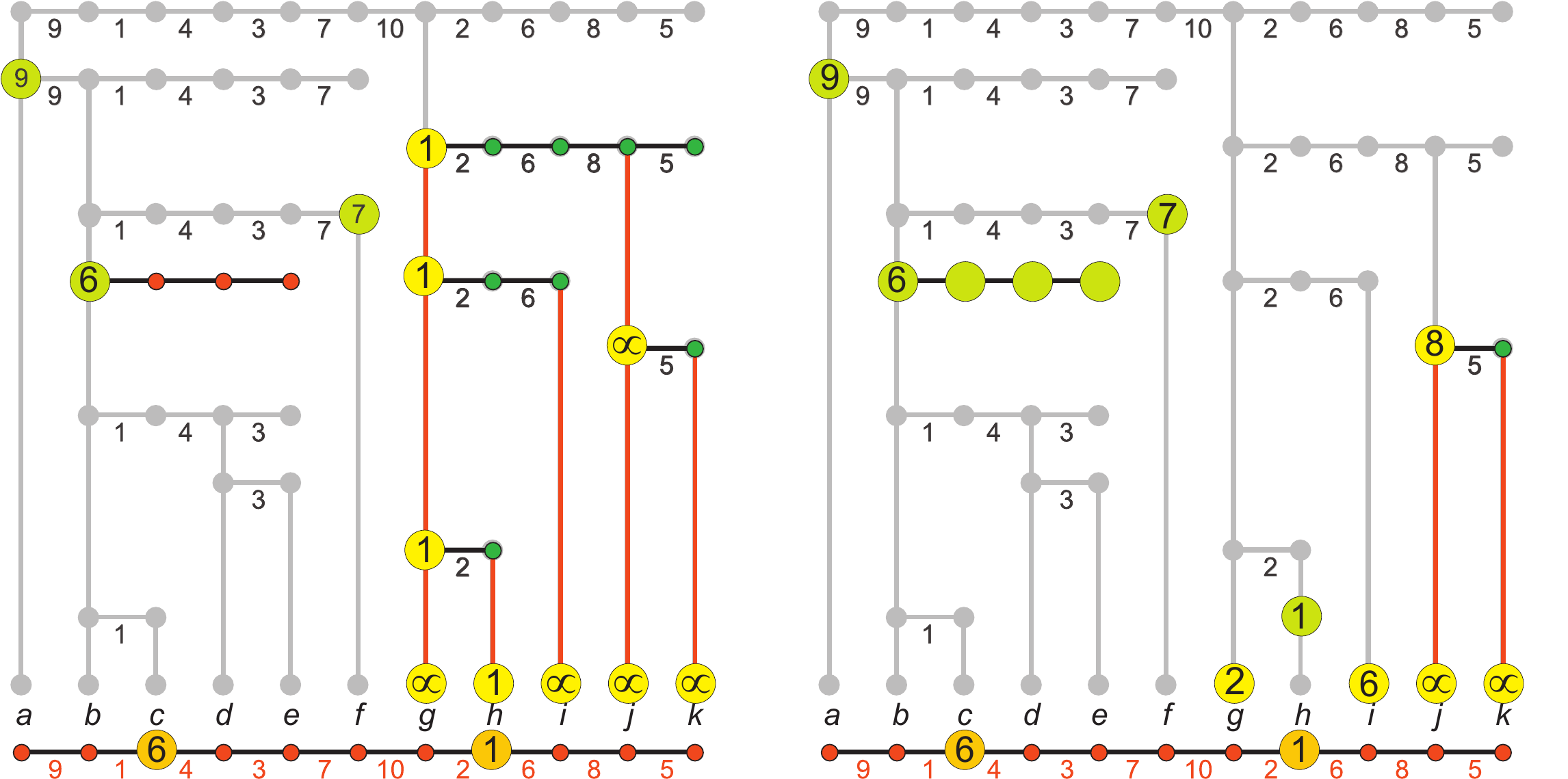}%
\end{center}
\end{figure}
%EndExpansion
%

%TCIMACRO{\TeXButton{EndFrame}{\end{frame}}}%
%BeginExpansion
\end{frame}%
%EndExpansion
%

%TCIMACRO{\TeXButton{BeginFrame}{\begin{frame}}}%
%BeginExpansion
\begin{frame}%
%EndExpansion

All ancestors of $\kappa^{(2)}\{c\}$ may be pruned.\ For $k>2,$ $\kappa
^{(k)}\{c\}$ is an ancestor of $c,$ the flooding level of all its immediate
successors which are not ancestors of $c,$ that is, brothers of $\kappa
^{(k-1)}\{c\}$ is lower or equal than $\operatorname*{diam}(\kappa
^{(k)}\{c\})$.\ The edge linking each brother $Y\ $of $\kappa^{(k-1)}\{c\}$
with its father $\kappa^{(k)}\{c\}$ is cut ; like that $Y$ becomes the root of
a sub-dendrogram ; as its flooding level is lower or equal than
$\operatorname*{diam}(\kappa^{(k)}\{c\}),$ one sets $\omega(Y)=\omega
(Y)\wedge\operatorname*{diam}(\kappa^{(k)}\{c\}).\ $On the same time all
ancestors of $\kappa^{(2)}\{c\} $ and the edges linking them are suppressed.\ %

%TCIMACRO{\TeXButton{EndFrame}{\end{frame}}}%
%BeginExpansion
\end{frame}%
%EndExpansion
%

%TCIMACRO{\TeXButton{BeginFrame}{\begin{frame}}}%
%BeginExpansion
\begin{frame}%
%EndExpansion

The set $\kappa^{(2)}\{c\}=[b,c,d,e]$ got its flooding level $6$ and its
upstream is pruned:

- $\kappa^{(3)}\{c\}=[b,c,d,e,f]$ is suppressed and the node $\{f\}$ becomes
the root of sub-dendrogram, with a ceiling value $\omega(\{f\})=\omega
(\{f\})\wedge\operatorname*{diam}(\kappa^{(3)}\{c\})=7.\ \ $As the
sub-dendrogram is reduced to a node, its ceiling value is its flooding value,
$7.$

- $\kappa^{(4)}\{c\}=[a,b,c,d,e,f]$ is suppressed and the node $\{a\}$ becomes
the root of sub-dendrogram, with a ceiling value $\omega(\{a\})=\omega
(\{a\})\wedge\operatorname*{diam}(\kappa^{(4)}\{c\})=9.\ \ $As the
sub-dendrogram is reduced to a node, its ceiling value is its flooding value,
$9.$

- $\kappa^{(5)}\{c\}=N$ is suppressed and the node $[g,h,i,j,k]$ becomes the
root of sub-dendrogram, with a ceiling value $\omega([g,h,i,j,k])=\omega
([g,h,i,j,k])\wedge\operatorname*{diam}(\kappa^{(5)}\{c\})=1.\ $The ceiling
value $\omega([g,h,i,j,k])$ of the root being known, flooding this
subdendrogram becomes completely independent from the rest of the processing.%

%TCIMACRO{\TeXButton{EndFrame}{\end{frame}}}%
%BeginExpansion
\end{frame}%
%EndExpansion
%

%TCIMACRO{\TeXButton{BeginFrame}{\begin{frame}}}%
%BeginExpansion
\begin{frame}%
%EndExpansion
%

%TCIMACRO{\FRAME{ftbpF}{4.0515in}{2.0838in}{0pt}{}{}{dendref-eps-converted-to.pdf}%
%width 4.0515in;  height 2.0838in;  depth 0pt;  original-width 8.5612in;
%original-height 4.3744in;  cropleft "0";  croptop "1";  cropright "1";
%cropbottom "0";  filename 'dendrEF-eps-converted-to.pdf';file-properties "XNPEU";}}}%
%BeginExpansion
\begin{figure}
[ptb]
\begin{center}
\includegraphics[
height=2.0838in,
width=4.0515in
]%
{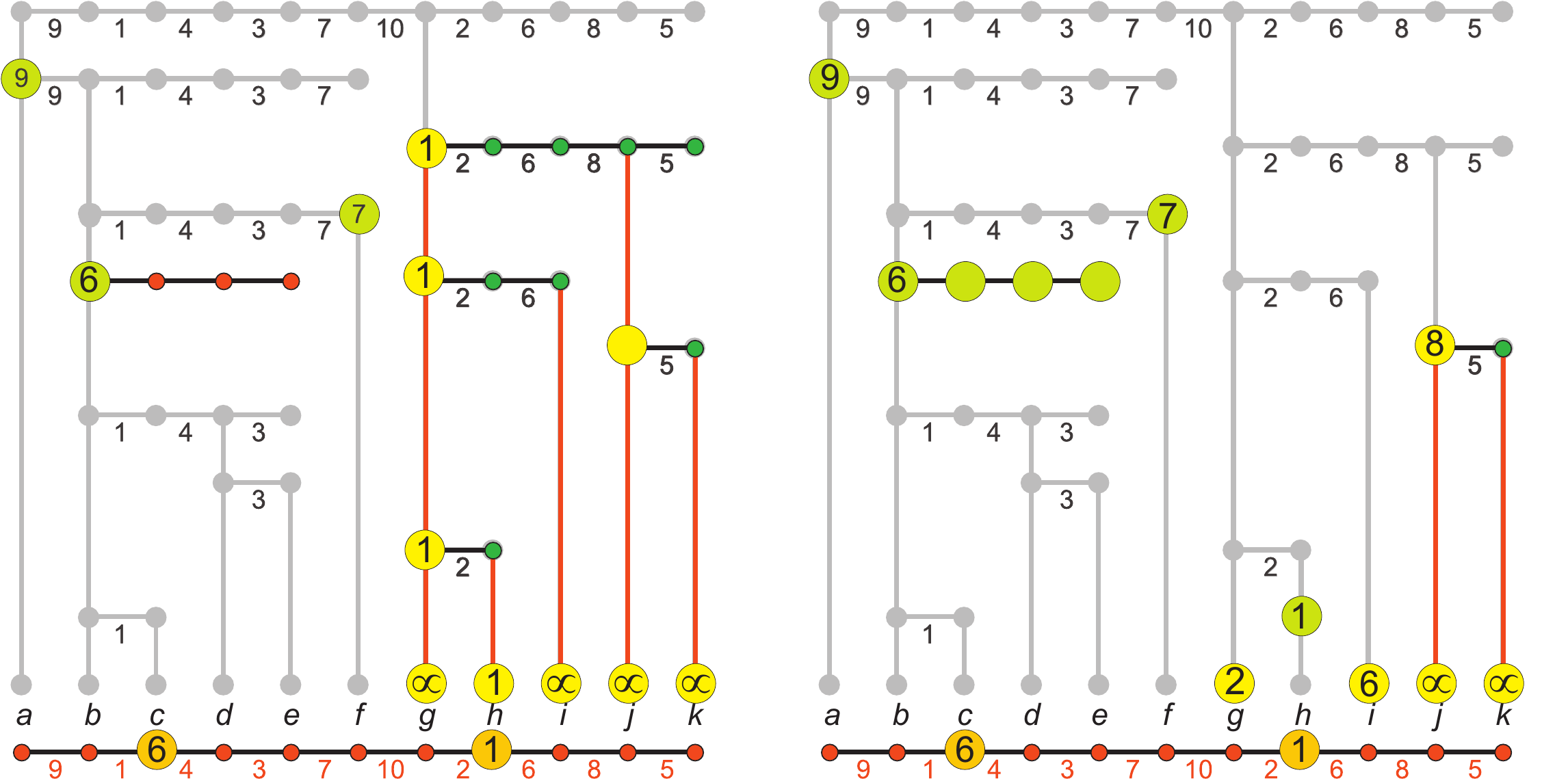}%
\end{center}
\end{figure}
%EndExpansion
%

%TCIMACRO{\TeXButton{EndFrame}{\end{frame}}}%
%BeginExpansion
\end{frame}%
%EndExpansion
%

%TCIMACRO{\TeXButton{BeginFrame}{\begin{frame}}}%
%BeginExpansion
\begin{frame}%
%EndExpansion

The smallest index for which $\omega(\kappa^{(k)}\{h\})\leq
\operatorname*{diam}(\kappa^{(k)}\{h\})$, is $k=1,$ with $\kappa
^{(1)}\{h\}=[g,h].\ $The flooding level of $\kappa^{(0)}\{h\}=[h]$ is
$\tau_{c}=\tau_{\kappa^{(0)}\{c\}}=\omega(\kappa^{(0)}\{h\})=1$ and the
upstream of $h$ can be pruned.\ 

- $\kappa^{(1)}\{h\}=[g,h]$ is suppressed and the node $\{g\}$ becomes the
root of sub-dendrogram, with a ceiling value $\omega(\{g\})=\omega
(\{g\})\wedge\operatorname*{diam}(\kappa^{(1)}\{h\})=1.\ \ $As the
sub-dendrogram is reduced to a node, its ceiling value is its flooding value:
$1.$

- $\kappa^{(2)}\{h\}=[g,h,i]$ is suppressed and the node $\{i\}$ becomes the
root of sub-dendrogram, with a ceiling value $\omega(\{i\})=\omega
(\{i\})\wedge\operatorname*{diam}(\kappa^{(2)}\{h\})=6.\ \ $As the
sub-dendrogram is reduced to a node, its ceiling value is its flooding value,
$6.$

- $\kappa^{(3)}\{h\}=[g,h,i,j,k]$ is suppressed and the node $[j,k]$ becomes
the root of sub-dendrogram, with a ceiling value $\omega([j,k])=\omega
([j,k])\wedge\operatorname*{diam}(\kappa^{(3)}\{h\})=8.\ $The node $[j,k]$
being the root of a dendrogram with a ceiling value higher than its diameter
gets flooded at the level of the ceiling value : $8.\ $This achieves the
process since there are no more sub-dendrograms to process.\ %

%TCIMACRO{\TeXButton{EndFrame}{\end{frame}}}%
%BeginExpansion
\end{frame}%
%EndExpansion
%

%TCIMACRO{\TeXButton{BeginFrame}{\begin{frame}}}%
%BeginExpansion
\begin{frame}%
%EndExpansion

\begin{center}
{\Large \alert{Contraction/expansion of flat zones and  dendrogram flooding}}
\end{center}

%

%TCIMACRO{\TeXButton{EndFrame}{\end{frame}}}%
%BeginExpansion
\end{frame}%
%EndExpansion
%

%TCIMACRO{\TeXButton{BeginFrame}{\begin{frame}}}%
%BeginExpansion
\begin{frame}%
%EndExpansion
%

%TCIMACRO{\QTR{frametitle}{Contracting inside edges of flat zones of node
%weighted graphs.}}%
%BeginExpansion
\frametitle{Contracting inside edges of flat zones of node weighted graphs.}%
%EndExpansion

$G_{n}:$ a node weighted graph with a ground level $f$ and a ceiling function
$\omega.\ $

Consider an edge $(p,q)$ such that $f_{p}=f_{q}.$ Contracting this edge =

- suppressing the edge $(p,q)$

- merging both nodes into a new node $s,$ with a ground value $f_{s}%
=f_{p}=f_{q}$ and a ceiling value $\omega_{s}=\omega_{p}\wedge\omega_{q}$

- suppress the edge linking a node $t$ with $p$ or $q$ and replace it with an
edge $(t,s).$

After contraction we get a new graph $G_{n}^{\prime}$ with a ground level
$f^{\prime}$ and a ceiling function $\omega^{\prime}$%

%TCIMACRO{\TeXButton{EndFrame}{\end{frame}}}%
%BeginExpansion
\end{frame}%
%EndExpansion
%

%TCIMACRO{\TeXButton{BeginFrame}{\begin{frame}}}%
%BeginExpansion
\begin{frame}%
%EndExpansion
%

%TCIMACRO{\QTR{frametitle}{Contracting inside edges of flat zones does not
%modify the dominated flooding }}%
%BeginExpansion
\frametitle{Contracting inside edges of flat zones does not modify the
dominated flooding }%
%EndExpansion

$\tau:$ the highest flooding of $G_{n}$ under $\omega$

$\tau^{\prime}:$ the highest flooding of $G_{n}$ under $\omega^{\prime}$

We show that on all common nodes $\tau=\tau^{\prime}$ and that $\tau_{p}%
=\tau_{q}=\tau_{s}$

We first remark that replacing $\omega_{p}$ and $\omega_{q}$ by $\omega
_{p}\wedge\omega_{q}$ does not change $\tau.$

The shortest path linking $\ s$ in $G_{n}^{\prime}$ with $\Omega$ is the same
as the shortest path linking $\Omega$ with $p$ or $q$ in $G_{n}^{\prime}.$

\begin{corollary}
All inside edges of flat zones may be contracted and produce each a unique
node without changing the highest flooding of the graph.\ 
\end{corollary}

%

%TCIMACRO{\TeXButton{EndFrame}{\end{frame}}}%
%BeginExpansion
\end{frame}%
%EndExpansion
%

%TCIMACRO{\TeXButton{BeginFrame}{\begin{frame}}}%
%BeginExpansion
\begin{frame}%
%EndExpansion
%

%TCIMACRO{\QTR{frametitle}{Combining contractions and flooding on a
%dendrogram}}%
%BeginExpansion
\frametitle{Combining contractions and flooding on a dendrogram}%
%EndExpansion

In the following figure A we want to construct the highest flooding of the red
function under the green one.\ The contraction of the flat zones produces in
fig.B a graph $G^{\prime}$ with 5 nodes $(a,b,c,d,e)$ with ground levels
$(0,4,1,2,0)$ and ceiling levels $(0,5,3,3,1).$ The edges are then weighted
with $\delta_{en}n,$ yielding the weights indicated in blue in fig.B.\ Fig.C
presents the associated dendrogram.\ We first flood the node $e.$
$\kappa^{(0)}(e)=\{e\}.\ $As $\operatorname*{diam}(\kappa^{(0)}\{e\})<\omega
(\kappa^{(0)}\{e\})\leq$ $\operatorname*{diam}(\kappa^{(1)}\{e\}),$ the
flooding level of $e$ is $\omega(\kappa^{(0)}\{e\}=1.$

$e$ has two brothers, the nodes $c$ and $d,$ roots of subdendrograms reduced
to 1 node.\ Having the same ceiling value, we have $\tau_{d}=\tau
_{c}=\operatorname*{diam}(\kappa^{(1)}\{e\}\wedge\omega(c)=2\wedge3=2.$

$e$ has two uncles, the nodes $a$ and $b,$ roots of subdendrograms reduced to
1 node.\ Their flooding value is $\tau_{a}=\operatorname*{diam}(\kappa
^{(1)}\{a\}\wedge\omega(a)=4\wedge0=0$ and $\tau_{b}=\operatorname*{diam}%
(\kappa^{(1)}\{b\}\wedge\omega(b)=4\wedge5=4$

During the expansion, each node of the graph $G^{\prime}$ is replaced by the
flat zone of the graph $G$ it represents, with identical flooding values, as
illustrated in fig.F.%

%TCIMACRO{\TeXButton{EndFrame}{\end{frame}}}%
%BeginExpansion
\end{frame}%
%EndExpansion
%

%TCIMACRO{\TeXButton{BeginFrame}{\begin{frame}}}%
%BeginExpansion
\begin{frame}%
%EndExpansion
%

%TCIMACRO{\QTR{frametitle}{Illustration}}%
%BeginExpansion
\frametitle{Illustration}%
%EndExpansion
%

%TCIMACRO{\FRAME{ftbpFU}{4.8335in}{1.9701in}{0pt}{\Qcb{Creation of flat zones
%and flooding of the edge associated graph}}{\Qlb{contractflod1}}%
%{contractflod1-eps-converted-to.pdf}{\special{ language "Scientific Word";  type "GRAPHIC";
%width 4.8335in;  height 1.9701in;  depth 0pt;  original-width 7.6264in;
%original-height 3.0826in;  cropleft "0";  croptop "1";  cropright "1";
%cropbottom "0";  filename 'contractflod1-eps-converted-to.pdf';file-properties "XNPEU";}}}%
%BeginExpansion
\begin{figure}
[ptb]
\begin{center}
\includegraphics[
height=1.9701in,
width=4.8335in
]%
{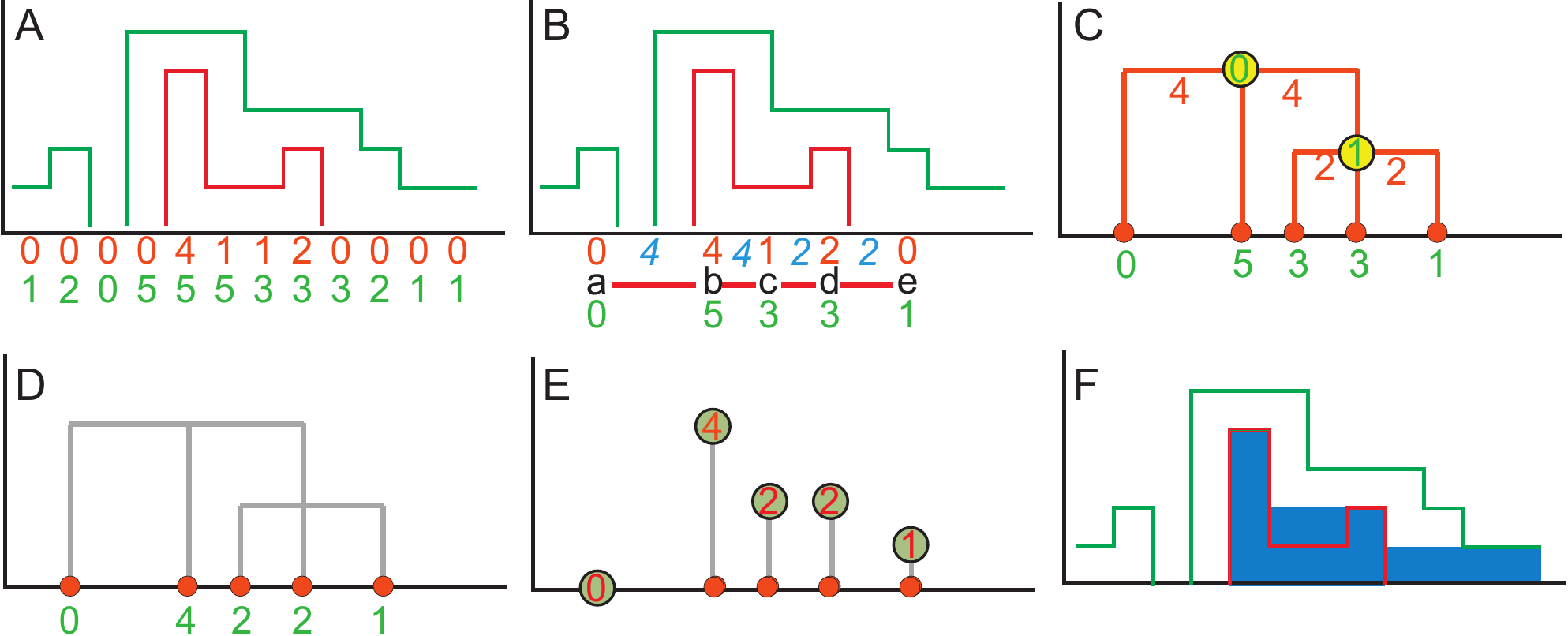}%
\caption{Creation of flat zones and flooding of the edge associated graph}%
\label{contractflod1}%
\end{center}
\end{figure}
%EndExpansion
%

%TCIMACRO{\TeXButton{EndFrame}{\end{frame}}}%
%BeginExpansion
\end{frame}%
%EndExpansion
%

%TCIMACRO{\TeXButton{BeginFrame}{\begin{frame}}}%
%BeginExpansion
\begin{frame}%
%EndExpansion
%

%TCIMACRO{\QTR{frametitle}{Construction of the MST while contracting the edges
%in the flat zones.\ }}%
%BeginExpansion
\frametitle{Construction of the MST while contracting the edges in the flat
zones.\ }%
%EndExpansion

Given a node weighted graph $G_{n}$ we assign to the edges weights equal to
$\delta_{en}n.$ It is beneficial to combine the construction of the MST of
this and simultaneously contract the edges.\ 

\textbf{Initialisation}

Create a tree with one node $p$ of the graph.\ 

\textbf{Expansion}

As long as the tree does not contain all nodes of the graph:

\qquad Chose the lowest edge $(q,s)$ in the cocycle of $T,$ such that $q\in T
$ and $s\notin T.$

\qquad If $f_{q}=f_{s}:$ contract the edge $(q,s)$ on $q$ and link $q$ with
the neighbors of $s$ not yet in $T$

\qquad Else : append the node $s$ to the tree: $T=T\cup\{s\}$%

%TCIMACRO{\TeXButton{EndFrame}{\end{frame}}}%
%BeginExpansion
\end{frame}%
%EndExpansion
%

%TCIMACRO{\TeXButton{BeginFrame}{\begin{frame}}}%
%BeginExpansion
\begin{frame}%
%EndExpansion

\begin{center}
{\Large \alert{Relations between floodings on edge and node weighted graphs.}}
\end{center}

%

%TCIMACRO{\TeXButton{EndFrame}{\end{frame}}}%
%BeginExpansion
\end{frame}%
%EndExpansion
%

%TCIMACRO{\TeXButton{BeginFrame}{\begin{frame}}}%
%BeginExpansion
\begin{frame}%
%EndExpansion
%

%TCIMACRO{\QTR{frametitle}{Flooding a node weighted graph = flooding an edge
%weighted graph}}%
%BeginExpansion
\frametitle{Flooding a node weighted graph = flooding an edge weighted graph}%
%EndExpansion

$G_{e}$ = edge weighted graph, $G^{n}$ = node weighted graph, $G_{e}^{n}$ =
node and edge weighted graph

For $G_{e}:\eta_{p}=\left(  \varepsilon_{ne}e\right)  _{p},$ i.e. the weight
of the lowest edge adjacent to the node $p.$

\begin{theorem}
Consider $G_{n},$ a node weighted graph, and $G_{e},$ the derived edge
weighted graph with edge weights $e=\delta_{en}n.\ $We then have the following
equivalences: $\{\tau\geq n$ e-flooding of $G_{e}\}\Leftrightarrow\{\tau$
n-flooding of $G_{n}\}$(eq-1)
\end{theorem}

This theorem has important algorithmic consequences.\ For constructing the
highest flooding on the node weighted graph under a ceiling function $\omega$
we may construct the highest flooding of the edge weighted graph $G_{e}$ under
$\omega.$%

%TCIMACRO{\TeXButton{EndFrame}{\end{frame}}}%
%BeginExpansion
\end{frame}%
%EndExpansion
%

%TCIMACRO{\TeXButton{BeginFrame}{\begin{frame}}}%
%BeginExpansion
\begin{frame}%
%EndExpansion
%

%TCIMACRO{\QTR{frametitle}{The waterfall flooding of an edge weighted graph}}%
%BeginExpansion
\frametitle{The waterfall flooding of an edge weighted graph}%
%EndExpansion

Consider an edge weighted graph $G_{e}.$ The waterfall flooding consists in
assigning to each node of the graph a flooding level equal to the lowest
adjacent edge: $\eta=\varepsilon_{ne}e.\ $The function $\eta$ is a particular
flooding of $G_{e}.$

\begin{lemma}
In an edge weighted graph $G_{e},$ a function $\tau$ on the nodes is a valid
flooding if and only if $\tau\vee\eta$ is a valid flooding: $\{\tau$
e-flooding of $G_{e}\}\Leftrightarrow\{\tau\vee\eta$ e-flooding of $G_{n}\}$
\end{lemma}

This equivalence has the following consequence:

\begin{itemize}
\item by replacing $\tau_{p}<\eta_{p}$ by $\eta_{p},$ $\tau$ remains an
e-flooding of $G_{e}$

\item by replacing $\tau_{p}=\eta_{p}$ by $\tau_{p}^{\prime}<\eta_{p},$
$\tau^{\prime}$ remains an e-flooding of $G_{e}$
\end{itemize}

%

%TCIMACRO{\TeXButton{EndFrame}{\end{frame}}}%
%BeginExpansion
\end{frame}%
%EndExpansion
%

%TCIMACRO{\TeXButton{BeginFrame}{\begin{frame}}}%
%BeginExpansion
\begin{frame}%
%EndExpansion
%

%TCIMACRO{\QTR{frametitle}{Edge weighted graph invariant by $\gamma_{e}$.\ }}%
%BeginExpansion
\frametitle{Edge weighted graph invariant by $\gamma_{e}$.\ }%
%EndExpansion

Given $G_{e}$ : an edge weighted graph verifying $e=\gamma_{e}e=\delta
_{en}\varepsilon_{ne}e.\ \ $We create a node weighted graph $G^{\eta}$ with
node weights $\eta=\varepsilon_{ne}e$.\ 

As the edge weights verify $e=\delta_{en}\varepsilon_{ne}e=\delta_{en}\eta$ we
have the equivalence 1:

$\{\tau\vee\eta\geq\eta$ e-flooding of $G_{e}\}\Leftrightarrow\{\tau\vee\eta$
n-flooding of $G^{\eta}\}.$

Equivalence eq-2 : $\{\tau$ e-flooding of $G_{e}\}\Leftrightarrow\{\tau
\vee\eta$ e-flooding of $G_{e}\}.\ $Thus:

\begin{theorem}
Consider $G_{e},$ an edge weighted graph where each edge is the lowest edge of
one of its extremities (invariant by $\gamma_{e})$, and $G^{\eta},$ the
derived node weighted graph with node weights $\eta=\varepsilon_{ne}e,$ we
then have the following equivalences: $\{\tau$ e-flooding of $G_{e}%
\}\Leftrightarrow\{\tau\vee\eta$ e-flooding of $G_{e}\}\Leftrightarrow
\{\tau\vee\eta$ n-flooding of $G^{\eta}\}.$
\end{theorem}

%

%TCIMACRO{\TeXButton{EndFrame}{\end{frame}}}%
%BeginExpansion
\end{frame}%
%EndExpansion
%

%TCIMACRO{\TeXButton{BeginFrame}{\begin{frame}}}%
%BeginExpansion
\begin{frame}%
%EndExpansion
%

%TCIMACRO{\QTR{frametitle}{Edge weighted graph invariant by $\gamma_{e}$.\ }}%
%BeginExpansion
\frametitle{Edge weighted graph invariant by $\gamma_{e}$.\ }%
%EndExpansion

Consider now $\tau^{\prime}\geq\eta$ an e-flooding of $G_{e}$ which is also a
n-flooding of $G_{\eta}.\ $Consider a subset of nodes $A$ of $N$.\ We define a
new node distribution as follows:

\begin{itemize}
\item on $A:\tau\leq\eta$

\item on $N/A:\tau=\tau^{\prime}$
\end{itemize}

This distribution verifies $\tau^{\prime}=\tau\vee\eta.\ $Hence, as stated in
the preceding theorem, $\tau$ and $\tau^{\prime}$ are both e-floodings of
$G_{e}.$%

%TCIMACRO{\TeXButton{EndFrame}{\end{frame}}}%
%BeginExpansion
\end{frame}%
%EndExpansion
%

%TCIMACRO{\TeXButton{BeginFrame}{\begin{frame}}}%
%BeginExpansion
\begin{frame}%
%EndExpansion
%

%TCIMACRO{\QTR{frametitle}{Case of a node weighted graph}}%
%BeginExpansion
\frametitle{Case of a node weighted graph}%
%EndExpansion

Given a node weighted graph $G^{n},$ we define $\varphi_{n}n=\varepsilon
_{ne}\delta_{en}n=\eta.$ On the other hand we assign to the edges the weights
$e=\delta_{en}n=\delta_{en}\varepsilon_{ne}\delta_{en}n=\delta_{en}\varphi
_{n}n=\delta_{en}\eta.$

The graph $G^{\eta}$ is a flooding graph as $\eta=\varepsilon_{ne}e$ and
$e=\delta_{en}\eta.$ The preceding results apply to the graphs $G_{e},G^{\eta
}$ and $G^{e}.$

$\{\tau$ n-flooding of $G^{n}\}\Leftrightarrow\{\tau\geq n$ e-flooding of
$G_{e}\}\Leftrightarrow\{\tau\vee\eta$ e-flooding of $G_{e}\}\Leftrightarrow
\{\tau\vee\eta$ n-flooding of $G^{\eta}\}.$%

%TCIMACRO{\TeXButton{EndFrame}{\end{frame}}}%
%BeginExpansion
\end{frame}%
%EndExpansion
%

%TCIMACRO{\TeXButton{BeginFrame}{\begin{frame}}}%
%BeginExpansion
\begin{frame}%
%EndExpansion
%

%TCIMACRO{\QTR{frametitle}{Case of dominated floodings}}%
%BeginExpansion
\frametitle{Case of dominated floodings}%
%EndExpansion

\begin{theorem}
A flooding $\tau$ is the highest flooding of an edge weighted graph $G_{e}$
under a ceiling function $\omega$ if and only if $\tau\vee\eta$ is the largest
flooding of $G_{e}$ under the ceiling function $\omega\vee\eta.$
\end{theorem}

\begin{theorem}
If $\chi$ is the largest flooding of $G_{e}$ under the ceiling function
$\omega\vee\eta,$ then $\chi\wedge\omega$ is the largest flooding of $G_{e}$
under the ceiling function $\omega.$
\end{theorem}

%

%TCIMACRO{\TeXButton{EndFrame}{\end{frame}}}%
%BeginExpansion
\end{frame}%
%EndExpansion
%

%TCIMACRO{\TeXButton{BeginFrame}{\begin{frame}}}%
%BeginExpansion
\begin{frame}%
%EndExpansion

\begin{center}
{\Large \alert{Combining contraction.expansion of flat zones and closing of
the isolated regional minima}}
\end{center}

%

%TCIMACRO{\TeXButton{EndFrame}{\end{frame}}}%
%BeginExpansion
\end{frame}%
%EndExpansion
%

%TCIMACRO{\TeXButton{BeginFrame}{\begin{frame}}}%
%BeginExpansion
\begin{frame}%
%EndExpansion
%

%TCIMACRO{\QTR{frametitle}{Case of a node weighted graph}}%
%BeginExpansion
\frametitle{Case of a node weighted graph}%
%EndExpansion

$G^{n}:$ a node weighted graph with ground level $f$ and ceiling level
$\omega$.\ 

Contracting its flat zones produces a graph $G^{\prime}=KG$ with ground level
$f$ and ceiling level $\omega_{1}.$ Each regional minimum of $f$ becomes an
isolated regional minimum.

The closing $\varphi_{n}$ in the graph $G^{\prime}$ fills each regional
minimum to the level of its lowest neighbor.\ 

$\{\tau$ n-flooding of $G^{\prime n}$ dominated by $\omega_{1}%
\}\Leftrightarrow\{\tau_{1}=\tau\vee\eta$ n-flooding of $G^{\prime\eta}$
dominated by $\omega_{1}\vee\eta\}.$

If we expand the nodes of $G^{\prime},$ replacing each node by the flat zone
it represents, with the flood distribution $\tau_{1}\wedge\omega,$ we get the
flooding of $G$ dominated by $\omega.$%

%TCIMACRO{\TeXButton{EndFrame}{\end{frame}}}%
%BeginExpansion
\end{frame}%
%EndExpansion
%

%TCIMACRO{\TeXButton{BeginFrame}{\begin{frame}}}%
%BeginExpansion
\begin{frame}%
%EndExpansion
%

%TCIMACRO{\QTR{frametitle}{Contraction and closing of the isolated regional
%minima}}%
%BeginExpansion
\frametitle{Contraction and closing of the isolated regional minima}%
%EndExpansion

A commutative diagram illustrates the process.\ 

$%
\begin{array}
[c]{ccccc}%
G,f,\omega & \rightarrow & KG,f,\omega_{1} & \rightarrow & KG,\varphi
_{n}f,\omega_{1}\vee\varphi_{n}f\\
\downarrow &  &  &  & \downarrow\\
\Lambda &  &  &  & \Lambda\\
\downarrow &  &  &  & \downarrow\\
EKG & \longleftarrow & \left[  KG,\tau_{1}\wedge\omega_{1}\right]  &
\longleftarrow & \tau_{1}=\Lambda\left(  KG,\varphi_{n}f,\omega_{1}\vee
\varphi_{n}f\right)
\end{array}
$

with
%TCIMACRO{\TeXButton{EndFrame}{\end{frame}}}%
%BeginExpansion
\end{frame}%
%EndExpansion
%

%TCIMACRO{\TeXButton{BeginFrame}{\begin{frame}}}%
%BeginExpansion
\begin{frame}%
%EndExpansion
{}%

%TCIMACRO{\QTR{frametitle}{Contraction and closing of the isolated regional
%minima : illustration}}%
%BeginExpansion
\frametitle{Contraction and closing of the isolated regional minima :
illustration}%
%EndExpansion
%

%TCIMACRO{\FRAME{ftbpF}{3.0195in}{1.7883in}{0pt}{}{}{frfl2-eps-converted-to.pdf}%
%width 3.0195in;  height 1.7883in;  depth 0pt;  original-width 4.7463in;
%original-height 2.7937in;  cropleft "0";  croptop "1";  cropright "1";
%cropbottom "0";  filename 'frfl2-eps-converted-to.pdf';file-properties "XNPEU";}}}%
%BeginExpansion
\begin{figure}
[ptb]
\begin{center}
\includegraphics[
height=1.7883in,
width=3.0195in
]%
{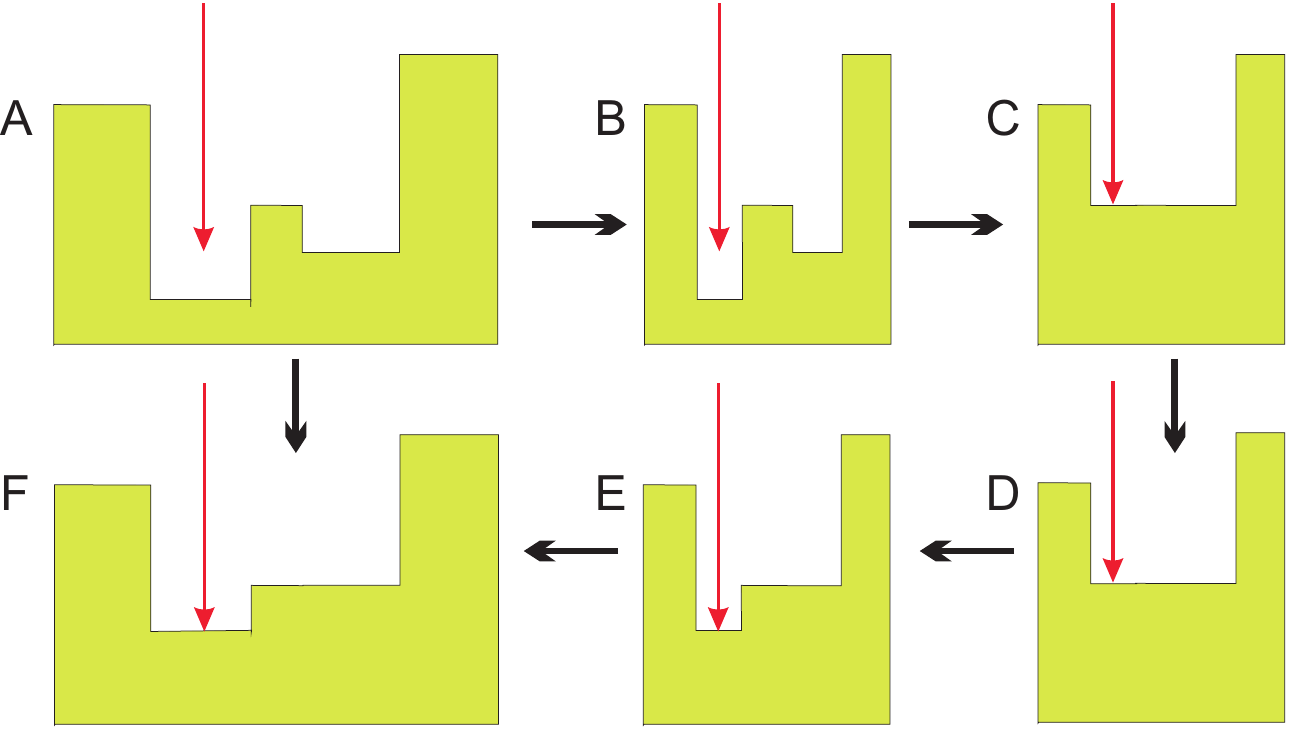}%
\end{center}
\end{figure}
%EndExpansion
%

%TCIMACRO{\TeXButton{EndFrame}{\end{frame}}}%
%BeginExpansion
\end{frame}%
%EndExpansion
%

%TCIMACRO{\TeXButton{BeginFrame}{\begin{frame}}}%
%BeginExpansion
\begin{frame}%
%EndExpansion
%

%TCIMACRO{\QTR{frametitle}{Contraction and closing of the isolated regional
%minima : illustration}}%
%BeginExpansion
\frametitle{Contraction and closing of the isolated regional minima :
illustration}%
%EndExpansion

Fig.A: presents a topographic surface $f$ and a ceiling function equal to
$\infty$ everywhere except at the position of the red arrow.

Fig.B:\ is obtained by contracting the flat zones with the associated ceiling
function $\omega_{1},$ giving a graph $KG$

Fig.C: is obtained by the closing $\varphi_{n}$, which closes the isolated
regional minima with a new ceiling function $\varphi_{n}f\vee\omega$

Fig.D: is the flooding $\tau$ of $\varphi_{n}f$ under $\varphi_{n}f\vee\omega$

Fig.E: the minimum $\tau_{1}\wedge\omega_{1}$ (which is also the flooding of
the function in fig.B under the ceiling $\omega_{1}$)

Fig.F: the expansion of fig.E yields the flooding of $f$ under $\omega.$%

%TCIMACRO{\TeXButton{EndFrame}{\end{frame}}}%
%BeginExpansion
\end{frame}%
%EndExpansion
%

%TCIMACRO{\TeXButton{BeginFrame}{\begin{frame}}}%
%BeginExpansion
\begin{frame}%
%EndExpansion
%

%TCIMACRO{\QTR{frametitle}{Cascading the preceding sequence}}%
%BeginExpansion
\frametitle{Cascading the preceding sequence}%
%EndExpansion

The preceding sequence of transformations constructs a dominated flooding of a
complex function thanks to the flooding of a simpler function. The same
sequence may be applied for flooding this simpler function.\ And so on,
producing a sequence of simpler and simpler functions to flood.%

%TCIMACRO{\TeXButton{EndFrame}{\end{frame}}}%
%BeginExpansion
\end{frame}%
%EndExpansion
%

%TCIMACRO{\TeXButton{BeginFrame}{\begin{frame}}}%
%BeginExpansion
\begin{frame}%
%EndExpansion

\begin{center}
{\Large \alert{Constructing a local flooding on a topographic graph}}
\end{center}

%

%TCIMACRO{\TeXButton{EndFrame}{\end{frame}}}%
%BeginExpansion
\end{frame}%
%EndExpansion
%

%TCIMACRO{\TeXButton{BeginFrame}{\begin{frame}}}%
%BeginExpansion
\begin{frame}%
%EndExpansion
%

%TCIMACRO{\QTR{frametitle}{Fast and local flooding of a topographic surface.}}%
%BeginExpansion
\frametitle{Fast and local flooding of a topographic surface.}%
%EndExpansion

Using all results established above we propose an algorithm for local floodings.\ 

$G_{n}$ a topographic graph, $f$ the ground level.$\ $We want to know the
flooding at a given node $p.$for the dominated flooding of $f$ under $\omega.$
Assigning to the edges the weights $\delta_{en}f$ produces a flooding graph:
the lowest adjacent edge of each node has the same level than this node.\ 

If $p$ belongs to an isolated regional minimum and $\omega_{p}\leq\left(
\varphi_{n}f\right)  _{p},$ then $\tau_{p}=\omega_{p}$

If not, we flood the function $\varphi_{n}f$ under the ceiling function
$\omega\vee\varphi_{n}f$ and get a flooding $\tau^{\prime}.\ $The desired
flooding is $\tau=\tau^{\prime}\wedge\omega.$%

%TCIMACRO{\TeXButton{EndFrame}{\end{frame}}}%
%BeginExpansion
\end{frame}%
%EndExpansion
%

%TCIMACRO{\TeXButton{BeginFrame}{\begin{frame}}}%
%BeginExpansion
\begin{frame}%
%EndExpansion
%

%TCIMACRO{\QTR{frametitle}{Constructing the lake containing p}}%
%BeginExpansion
\frametitle{Constructing the lake containing p}%
%EndExpansion

For $X=\operatorname*{Ball}(p,f_{p}):$ if $\omega(X)\leq f_{p},$ then $p$ is
in the upstream of a lake and is dry: $\tau_{p}=f_{p}$

If $\omega(X)>f_{p}$, we search the lake containing $p$ :

\qquad Until $\operatorname*{diam}(X)<\omega(X)\leq\varepsilon_{e}(X)$ do
$X=\kappa(X)=\operatorname*{Ball}(p,f_{p})$\bigskip

The lake containing $p$ is $X$ at a level $\omega(X).$%

%TCIMACRO{\TeXButton{EndFrame}{\end{frame}}}%
%BeginExpansion
\end{frame}%
%EndExpansion
%

%TCIMACRO{\TeXButton{BeginFrame}{\begin{frame}}}%
%BeginExpansion
\begin{frame}%
%EndExpansion
%

%TCIMACRO{\QTR{frametitle}{Procedure "up\_hill"}}%
%BeginExpansion
\frametitle{Procedure "up\_hill"}%
%EndExpansion

If $X$ is lake of the flooding, its uphill up to a level $\mu$ is constructed
with the procedure "up\_hill$(X,\mu)"$\bigskip

While $\varepsilon_{e}(X)\leq\mu$

\qquad$\lambda=\varepsilon_{e}(X)$

\qquad$Y=\operatorname*{Ball}(X,\varepsilon_{e}(X))$

\qquad For each connected component $Z_{i}$ of $Y/X$ for which $\lambda
>f_{Z_{i}}:$

\qquad\qquad if $\omega(Z_{i})\geq\lambda:\tau_{Z_{i}}=\lambda$

\qquad\qquad else if $\omega$ is minimal at $q$ in $Z_{i}:$ apply
"up-hill($\operatorname*{Ball}(q,\omega_{q}),\lambda)"$

\qquad For each $p\in Y/X$ verifying $f_{p}=\lambda:\tau_{p}=\lambda$

\qquad$X=Y$

End While%

%TCIMACRO{\TeXButton{EndFrame}{\end{frame}}}%
%BeginExpansion
\end{frame}%
%EndExpansion
%

%TCIMACRO{\TeXButton{BeginFrame}{\begin{frame}}}%
%BeginExpansion
\begin{frame}%
%EndExpansion
%

%TCIMACRO{\QTR{frametitle}{Conclusion}}%
%BeginExpansion
\frametitle{Conclusion}%
%EndExpansion

The comparison between floodings on edge weighted graphs and on node weighted
graphs has given a better insight in both of them.

New algorithms have been derived allowing to chose the one is best suited for
each application (type or processor, hardware, parallel processing, etc.)%

%TCIMACRO{\TeXButton{EndFrame}{\end{frame}}}%
%BeginExpansion
\end{frame}%
%EndExpansion

\end{document}